\documentclass[11pt]{article}

\usepackage[final]{acl}
\usepackage{xcolor}
\usepackage{times}
\usepackage{latexsym}
\usepackage[T1]{fontenc}
\usepackage[utf8]{inputenc}
\usepackage{microtype}
\usepackage{inconsolata}
\usepackage{graphicx}
\usepackage{makecell}
\usepackage{caption}
\usepackage[singlelinecheck=false]{subcaption}

\usepackage{booktabs}
\usepackage{amsmath}
\usepackage{array}
\usepackage{multirow}
\usepackage{enumitem}
\usepackage{xspace}

\newif\ifshowcomments

\showcommentstrue  

\definecolor{purp}{HTML}{af1fcf}  

\title{``High-Quality'' Data for Early-Stage Instruction Tuning:\\
An Empirical Study of Data Quality Metrics}

\author{Vladimir Necula$^{*\dagger}$ \and David Gvadzabia$^{*\ddagger}$ \and Sofia Serrano$^\ddagger$ \\
$^\dagger$University of Cambridge Department of Physics \\
         $^\ddagger$Lafayette College Department of Computer Science \\
         \texttt{vgn22@cam.ac.uk, \{gvadzabd, serranos\}@lafayette.edu}
}

\newcolumntype{C}[1]{>{\centering\let\newline\\\arraybackslash\hspace{0pt}}m{#1}}

\begin{document}

\captionsetup[subfigure]{justification=centering}
\maketitle
\makeatletter
\ifacl@finalcopy
    \begingroup
    \def\thefootnote{*}
    \footnotetext{Equal contribution.}
\endgroup
\fi
\makeatother

\begin{abstract}
``Data quality'' is widely invoked in instruction tuning, but the term is often underspecified. We study it as an operational property by asking which simple example-level metrics, drawn from prior work on data selection, text quality, and instruction tuning, best predict training utility under a fixed finetuning budget. For each metric, we sort instruction-response examples by score, partition them into low, middle, and high bins with roughly equal token mass, finetune separately on each bin, and compare validation perplexity and LLM-as-judge scores for the models' generated answers as a function of cumulative training tokens. Our results suggest that answer-dense supervision, as captured by response ratio, is a stronger and more portable predictor of instruction-tuning utility than commonly used alternative proxies.
\end{abstract}

\section{Introduction}

Instruction tuning is now a standard stage in adapting language models to follow natural-language instructions. Yet despite many references to instruction data quality over the past few years \citep[e.g., ][]{lima, longismore, mammothpaper, mammoth2, bukharin-etal-2024-data, dataqualitythroughgradients, bestofbothworlds}, precisely what constitutes instruction data ``quality'' is still murky. Many candidate properties have been proposed or suggested to matter: e.g., information-dense responses may provide more training signal, or semantically aligned responses may better teach the intended task. In practice, however, these signals are often correlated, noisy, or dataset-dependent. This makes it difficult to distinguish metrics that truly capture utility during training from metrics that appear useful only because they are correlated with stronger signals.

We study the question: \emph{which example-level metrics are most predictive of early-stage instruction-tuning utility?} We operationalize utility two ways: as validation perplexity, and as LLM-as-judge scores for a model's generated answers \citep{llmasjudge}, under a fixed training budget. Rather than training on a dataset and then attempting to interpret correlations post hoc, we sort training examples by a candidate metric, partition them into low, middle, and high bins, train separate models on each bin, and compare their learning curves. If a metric is genuinely useful, then higher-scoring bins should produce lower validation perplexity, and higher LLM-as-judge scores.

We study four interpretable metrics motivated by prior work on instruction-data selection, text-quality measurement, and instruction-tuning dynamics.
We evaluate these metrics across two models (Qwen3.5-9B-Base and Llama-3.1-8B) and two instruction-tuning datasets (Natural Instructions and FLAN). Across these settings, one conclusion is especially stable: the ratio of response length to prompt length (in tokens) is the most robust predictor of training utility of our four metrics. 

Alignment and semantic density, highly correlated with response ratio, often look promising as well. To clarify that response ratio is the more important underlying factor, we perform cross-ablation experiments. We remove high-response-ratio examples and then re-bin the remaining data by alignment, and vice versa, for new training runs. Removing the high-response-ratio instances from alignment-based binning shows a much larger impact than removing the high-alignment instances from response-ratio-based binning. This asymmetry suggests  response ratio may be a more fundamental driver of training utility.

Our main contributions are:
\begin{itemize}[noitemsep, leftmargin=*]
    \item A controlled comparison of four quality metrics for instruction data under set finetuning budgets.
    \item Showing that response ratio is the most robust predictor of training utility across our results.
    \item Showing that other metrics can carry signal, but that part of their apparent utility can be explained by overlap with response ratio.
    \item Introducing overlap and cross-ablation analyses that help distinguish direct metric effects from correlated proxy effects.
\end{itemize}

\section{Experimental Setup}

We do not claim that any one metric defines data quality in an absolute sense. Instead, we use an operational definition tied to training utility: 
\textbf{A metric is more tied to ``data quality'' if its high bin yields lower validation perplexity, and a model whose generated answers score higher, under a fixed training budget.} Details of our LLM-as-judge setup are in Appendix~\ref{sec:llm-as-judge-details}.

Under this definition, a metric is more useful when higher-scoring bins reliably produce better learning curves than lower-scoring bins.
Our goal is therefore not to train the strongest possible final model, but to compare metric-induced bins under a fixed and matched training budget.

\subsection{Quality Metrics}

For each training example \(e = (d, x, y)\), where \(d\) is the task definition, \(x\) is the input, and \(y\) is the target response, we compute four scalar metrics inspired by existing work on instruction-tuning and training data. Formal definitions and explanations of each metric are available in Appendix~\ref{sec:more-on-metrics}. These metrics measure text traits at three levels of complexity:

\paragraph{Low-level text length measurements:} Following prior work that treats instruction length as a simple complexity and data-selection baseline, we first consider the \textbf{length} of an \textit{instance} in tokens \citep{liu2024deita}. We also consider \textbf{response ratio} \citep{huertaenochian2024promptloss}, which in our setting corresponds to the ratio of gold response length to the preceding prompt length (in tokens).

\paragraph{Aggregations of token counts:} We also consider a quantity that incorporates token counts, if not semantics. 
Motivated by work such as \citet{zevallos2023frequency} and \citet{qiu2024semanticdensity}, we use a \textbf{semantic density} measure, where we average the inverse log frequencies of each response token in an instance (with those frequencies estimated from the full set of gold responses in the dataset).

\paragraph{Semantics:} Finally, inspired by \citet{li2025scar}, we also consider the \textbf{alignment} of an instance's response and its prompt, as measured by the \texttt{all-MiniLM-L6-v2} Sentence Transformers model \citep{reimers2019sentencebert}.

\subsection{Token-Balanced Binning}

We do \emph{not} split examples into equal thirds by example count. Instead, for each metric we:
\begin{enumerate}[noitemsep, leftmargin=*]
    \item score every training example,
    \item sort examples by the score in ascending order,
    \item partition the sorted list at approximately 33\% and 66\% of \emph{total token mass}.
\end{enumerate}

This yields three bins, denoted \texttt{0--33}, \texttt{33--66}, and \texttt{66--100}, that each contain roughly one third of the training tokens, since token budget, not raw example count, is the natural unit of training cost, aggregating across all response and context tokens.

The \texttt{66--100} bin therefore contains the examples with the highest metric values, but under a training budget comparable to the other bins.

\subsection{Models, Datasets, and Training Protocol}

We evaluate the four metrics described above using two models: the \textbf{base} versions of Qwen3.5 (9B, \citealp{qwen3.5}) and Llama-3.1 (8B, \citealp{grattafiori2024llama3}), to simulate the setting at the \textbf{outset} of supervised finetuning (SFT). These models provide two independently developed decoder-only language-model backbones at a comparable parameter scale, allowing us to test whether metric behavior is stable across model families.

We use two instruction-tuning datasets to test whether a metric behaves consistently across different data sources: Natural Instructions \citep{wang2022supernaturalinstructions} and FLAN \citep{longpre2023flan}. Natural Instructions provides a large collection of diverse NLP tasks paired with expert-written natural-language task instructions, while FLAN is a broad instruction-tuning mixture constructed from multiple task collections and prompt formats. 

Recent work has demonstrated the potency of early-stage training dynamics during SFT for final model performance \citep{pareja2025unveiling}, which extends previous findings from \citep{teamolmo2025olmo2} that ``microannealing'' on potential data sources during midtraining is a valid way of testing data's training usefulness for training. We therefore focus specifically on early-stage instruction-tuning, which also allows us to perform 48 model training runs (one for every combination of two models, two datasets, four metrics, and three bins) without wasting additional compute. Given the large size of both Natural Instructions and FLAN, for the core comparison grid in this paper, we therefore use a 10\% randomly sampled training subset as a controlled probe of early learning behavior (retaining the full validation split). This passes the point at which validation perplexity stops reliably decreasing; based on recent work by \citet{harada-etal-2025-massive} that reaffirms the importance of perplexity as an indicator for SFT performance, that indicates we train long enough to capture meaningful patterns.

For each model-data combination, and for all three bins of all four quality metrics, we perform a separate three-epoch fine-tuning run using LoRA \citep{hu2021lora}, with four intra-epoch calculations of validation perplexity. Appendix~\ref{more-on-setup} gives additional detail about training setup and compute.

\section{Results}
\begin{figure}[!htbp]
	\includegraphics[width=\linewidth]{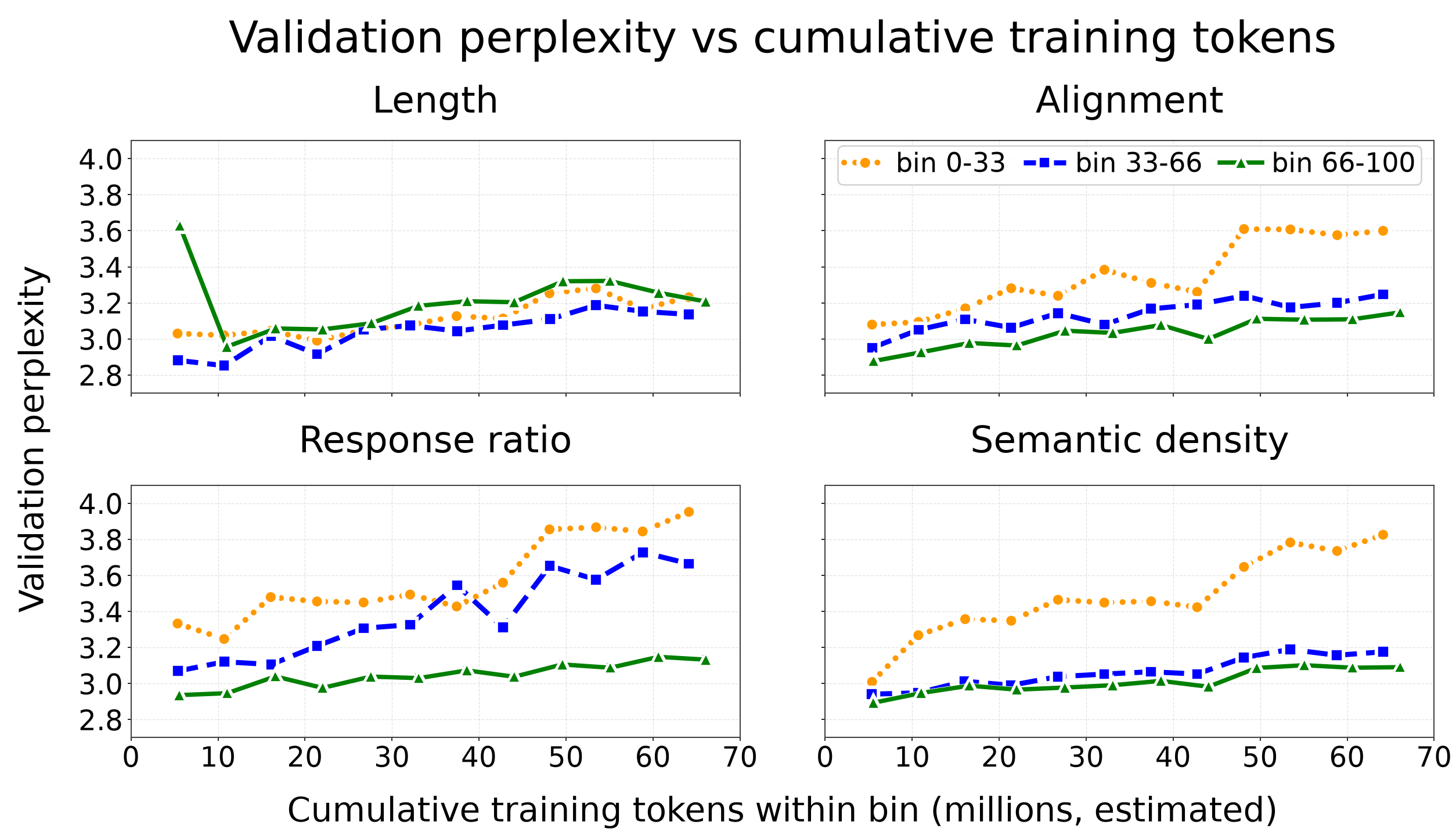}
    \caption{Validation perplexities for Qwen3.5-9B-Base finetuned on Natural Instructions. Of the 66-100 bins, only response ratio's results in perplexities significantly below its other bins. Confidence intervals for these graphs are so small as to not be visible in this figure.}
    \label{fig:condensed-results}
\end{figure}
\begin{figure}[!htbp]
	\includegraphics[width=\linewidth]{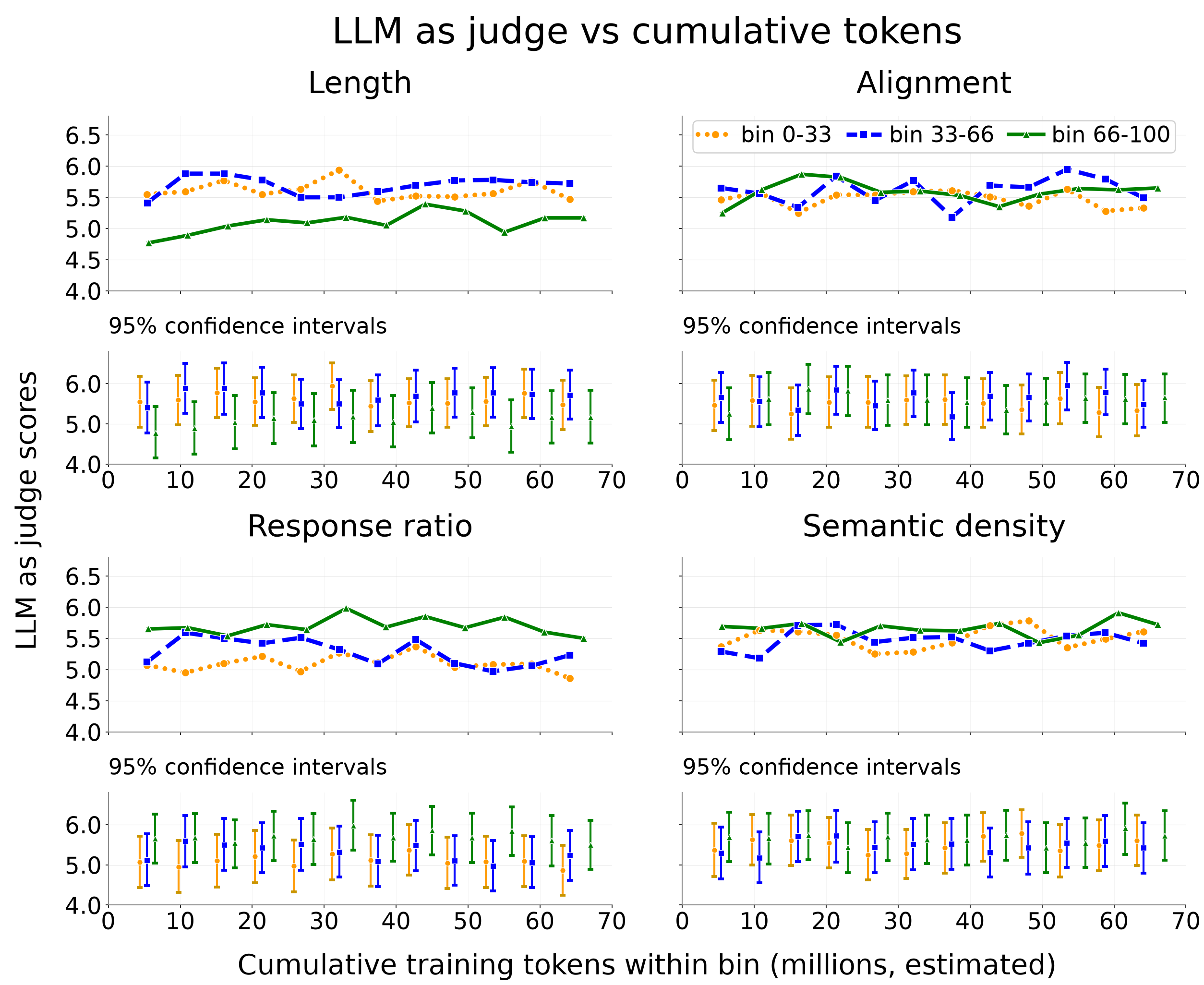}
    \caption{LLM-as-judge scores for Qwen3.5-9B-Base finetuned on Natural Instructions, where \textbf{higher scores are better}. Confidence intervals for all bins heavily overlap, but for response ratio, the distribution of scores for high-bin instances shifts higher shortly into training.}
    \label{fig:condensed-results-llmasjudge}
\end{figure}

Results for Qwen3.5-9B-Base on Natural Instructions are shown in figures~\ref{fig:condensed-results} and~\ref{fig:condensed-results-llmasjudge}, as those are largely representative; the full set of plots from all setups is in Appendix~\ref{sec:main-curves-figs}.
Across the model/data combinations, we find non-length metrics generally stratify by bin for perplexity (with response ratio demonstrating a greater gap between its 66-100 bin and its others). Stratifying in LLM-as-judge scores proves much harder, though, which is expected due to the early stage of training. Still, response ratio's 66-100 bin is on the higher end of LLM-as-judge scores for all four model-dataset combinations, and often alignment's and semantic density's too.

However, the 66-100 bin for alignment and semantic density \textit{doesn't} achieve the lowest perplexities on FLAN, where it's actually the \textit{middle} alignment bin that achieves their best performance. \textbf{Can we explain \textit{why} prompt-response alignment behaves like a strong signal of data utility in Natural Instructions, but not in FLAN?}

Summarizing differences between datasets is complex, but one relevant factor might be a diminished connection between high response ratio and high alignment or semantic density in FLAN. Since response ratio offers the most consistent signal of training utility across the settings we examined, perhaps the behavior of other metrics can be partly explained by how the highest response-ratio examples are distributed across other metrics' bins.

Table~\ref{tab:rr-distribution} summarizes this analysis, and helps to explain-- at least for alignment-- the perplexity curves from our experiments. In Natural Instructions, high-response-ratio examples are concentrated in the highest-alignment bin (48   .7\%). This lets us explain why alignment's perplexities appear so clean on Natural Instructions: the strongest response-ratio examples are also concentrated among the highest-alignment examples. But in FLAN, the high-response-ratio examples are concentrated in the \textit{middle}-alignment bin (49.2\%). This provides one possible explanation at least for the FLAN alignment anomaly: if the most useful response-ratio examples are disproportionately assigned to the middle alignment bin, then the middle alignment bin can outperform the high alignment bin even if alignment itself still carries some signal.

\begin{table}[t]
\setlength{\tabcolsep}{0pt}
\centering
\small
\begin{tabular}{p{2.2cm}C{1cm}C{1cm}C{1cm}|C{0.8cm}C{0.8cm}C{0.8cm}} \\
\toprule
& \multicolumn{3}{c|}{\textbf{Natural Instructions}} & \multicolumn{3}{c}{\textbf{FLAN}} \\
\textbf{Metric} & \makecell{\textbf{0--}\\ \textbf{33}} & \makecell{\textbf{33--}\\\textbf{66}} & \makecell{\textbf{66--}\\\textbf{100}} & \makecell{\textbf{0--}\\ \textbf{33}} & \makecell{\textbf{33--}\\\textbf{66}} & \makecell{\textbf{66--}\\\textbf{100}} \\
\midrule
Alignment & 18.9 & 32.4 & \textbf{48.7} & 30.9 & \textbf{49.2} & 20.0 \\
Semantic  density& 34.0 & 26.8 & \textbf{39.2} & 17.6 & 29.9 & \textbf{52.5} \\
Length & \textbf{68.8} & 24.6 & 6.6 & \textbf{85.5} & 9.9 & 4.5 \\
\bottomrule
\end{tabular}
\caption{Percentages of highest-response-ratio examples (tokenized by Qwen3.5) in other metrics' bins.}
\label{tab:rr-distribution}
\end{table}

This overlap analysis does not prove that alignment is unimportant, and indeed, it only offers a response-ratio-based explanation for alignment and not semantic density. But it \textit{does} show that the apparent usefulness of alignment is partly mediated by how high-response-ratio examples are distributed across alignment bins.

\section{Cross-Ablation Checks}
\label{sec:cross-ablation}

Overlap alone is suggestive but not a full explanation (see semantic density). But the weaker connection between alignment and response ratio in FLAN presents an opportunity to decouple them. With this goal, we run cross-ablation experiments that remove the highest bin of one metric and then re-bin the remaining data by the other metric, resulting in two additional sets of three bins: binning by \textbf{alignment without high response ratio}, and by \textbf{response ratio without high alignment}.

Here, we focus specifically on FLAN + Qwen; we show cross-ablation results on all model-data combinations in appendix figures~\ref{figs:standard-cross-ablation-llamaFLAN} through~\ref{figs:standard-cross-ablation-judge-qwenNI}.

\begin{figure}[!htbp]
	\includegraphics[width=\linewidth]{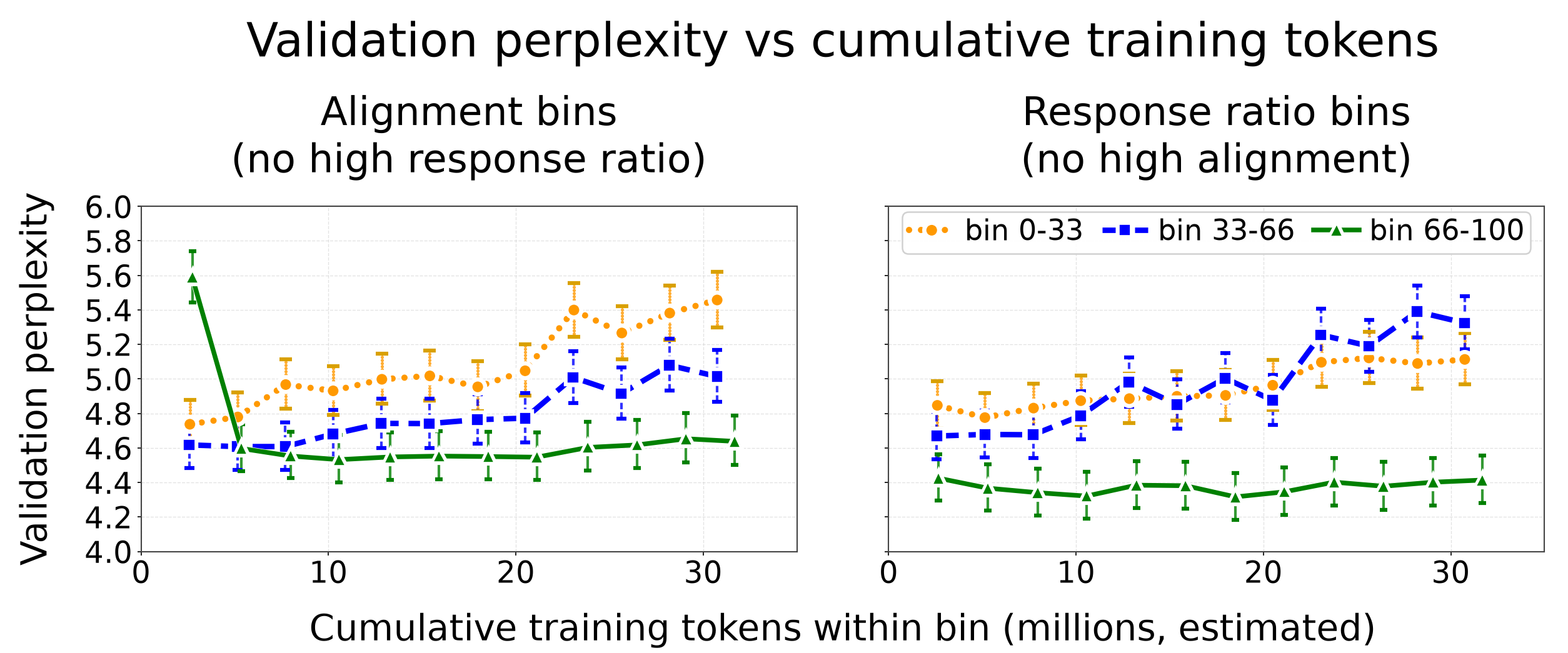}
    \caption{Cross-ablation perplexities for Qwen3.5-9B-Base on FLAN.}
    \label{fig:condensed-cross-ablation-results}
\end{figure}

\begin{figure}[!htbp]
	\includegraphics[width=\linewidth]{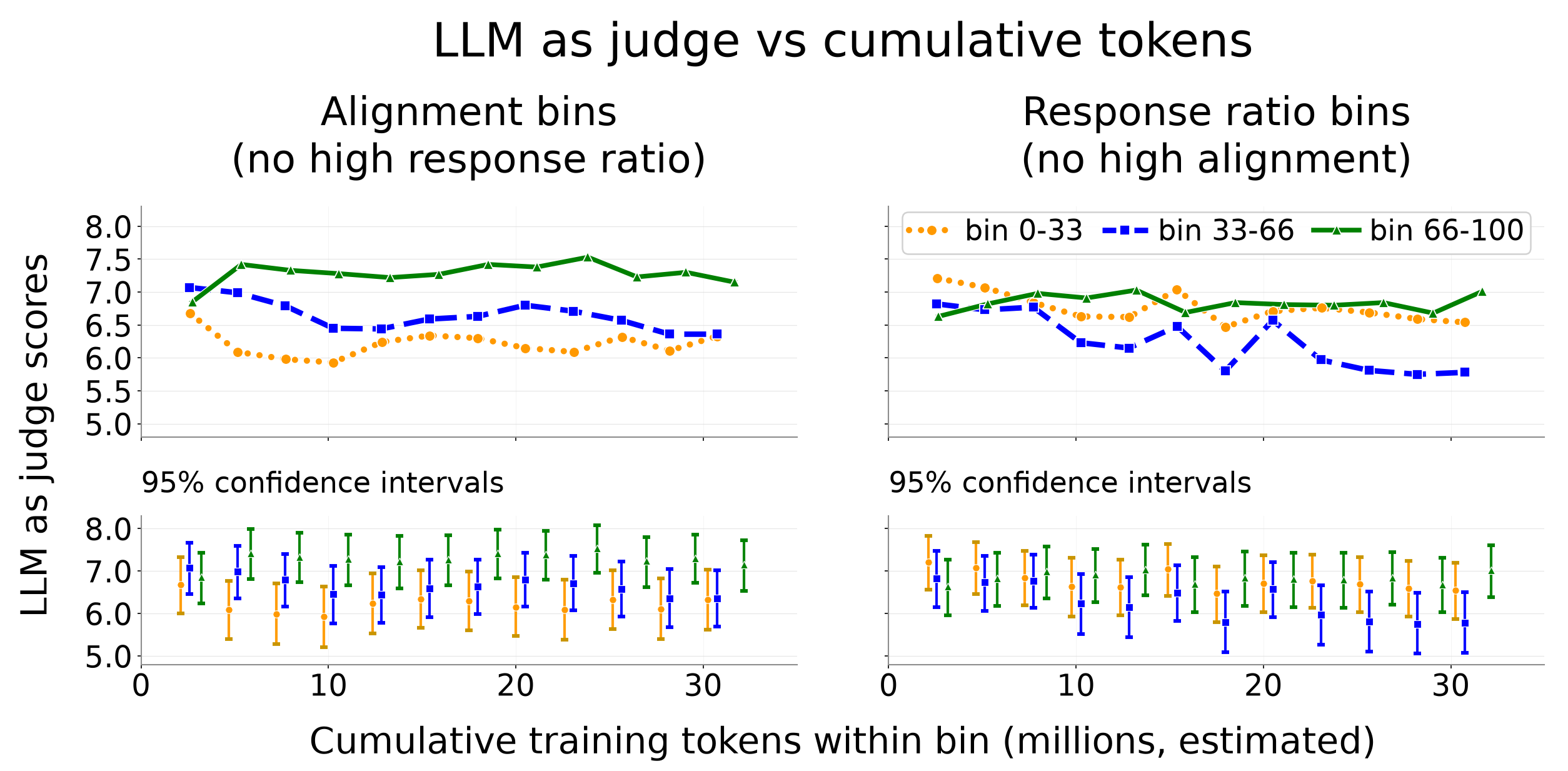}
    \caption{Cross-ablation LLM-as-judge scores for Qwen3.5-9B-Base on FLAN.}
    \label{fig:condensed-cross-ablation-results-llmasjudge}
\end{figure}

Figures~\ref{fig:condensed-cross-ablation-results} and~\ref{fig:condensed-cross-ablation-results-llmasjudge} show perplexities and LLM-as-judge scores resulting from this experiment. After removing examples with high response ratio (previously concentrated in alignment's 33-66 bin) and then rebinning, we see that the 33-66 bin no longer competes with the 66-100 bin for best alignment-binned performance, indicating that high-response-ratio instances were responsible for the 33-66 alignment bin's strong FLAN performance in the original runs.
And after removing high-alignment examples, response ratio curves for both perplexity and LLM-as-judge scores only \textit{minimally} change compared to those using the non-ablated training data, supporting that response ratio is the primary underlying driver of training utility.

\section{Discussion and Conclusion} \label{sec:conclusion}

Our results point to a simple but important interpretation: \textbf{answer-dense supervision matters}. Response ratio measures how many answer tokens the model receives relative to prompt tokens. This matters because, in our supervised finetuning setup, prompt tokens provide conditioning context but the strongest supervised learning signal comes from predicting the response. \citet{longismore} previously found that selecting a fixed number of long-response \textit{examples} was more helpful for training; our results bolster their argument by demonstrating that \textit{even under a fixed token budget,} meaning \textit{prompt} length is sacrificed, high-response-ratio examples achieves faster and lower validation-perplexity trajectories during SFT. And based on our cross-ablation results, this increased relative response length is \textbf{even more helpful than training on answers that are better aligned with contexts.}

Considering which instances have high response ratios suggests \textit{why} high response-ratio instances are so valuable for training. Instruction-tuning instances with the highest response ratios are those where the gold answer \textit{must} expand the \textit{most} to fulfill the prompt, while still staying concise and on-topic. Such tasks likely benefit from more training supervision than, say, writing single-letter multiple-choice answers. For more details, see Appendix~\ref{sec:qual-analysis}.

Furthermore, we note that one of the simplest of our quantities can compete with a transformer-based alignment metric. This indicates that while the concept of ``data quality'' may remain elusive, it may be simpler than expected to describe how much models gain from different data during SFT.

In summary, we studied four example-level quality metrics for instruction tuning across two models and two datasets. The main result is consistent across settings: Despite its simplicity, response ratio is a robust predictor of training utility.

\section*{Limitations}

\begin{itemize}[leftmargin=*]
    \item \textbf{Single-seed results.} The current experiments use a single random seed. While we made this choice to enable us to run the necessary 48 core finetuning runs, plus the 24 cross-ablation runs, we acknowledge that multiple seeds would provide a stronger estimate of robustness.
    \item \textbf{Downsampled training sets.} Drawing from \citet{pareja2025unveiling}'s results on the strong predictions of final model performance that early-stage training dynamics during SFT provide, the core runs use 10\% sampled training subsets rather than full training sets. This enables us to run a broader range of early-training setups, but using the full training sets would enable deeper analysis of how a smaller set of training runs change over longer periods of time.
    \item \textbf{High-level view of datasets.} Here, we focus predominantly on proposed metrics that might correspond to data quality by grouping instances in aggregate. While we do offer some descriptions of which kinds of instances tend to have high response ratio values in section~\ref{sec:conclusion} and Appendix~\ref{sec:qual-analysis}, these are still high-level trends, and a more fine-grained binning by these metrics might reveal more nuanced patterns.
    \item \textbf{Focus on 7B-9B model sizes.} Another tradeoff that we make to allow us to run the our full set of 48 core training runs + 24 cross-ablation runs is to limit our analysis to models with roughly 7B-9B parameters. While existing work supports that using smaller models to analyze the training utility of data instances \textit{does} produce conclusions about data utility that transfer to larger model sizes as well \citep{superfiltering, less}, our work does not explicitly confirm that those findings transfer to our experimental setup.
\end{itemize}

\section*{Ethical Considerations}

Data filtering can implicitly privilege some styles of supervision over others. A metric such as response ratio may favor answer-dense examples and underweight shorter answers that are nevertheless correct and useful. We therefore do not argue that any single metric should be used as a universal filter. Instead, we view these metrics as analytical tools for understanding which properties of instruction data are most predictive of training utility under specific experimental conditions.

\bibliography{custom}

\appendix

\section{Expanded metric definitions} \label{sec:more-on-metrics}

These metrics are not intended as new intrinsic definitions of data quality. Rather, they operationalize families of signals that have appeared in prior work on instruction-data selection, text-quality measurement, and supervised instruction fine-tuning. In the codebase, both Natural Instructions and FLAN are mapped to the same schema \texttt{definition}/\texttt{inputs}/\texttt{targets}, so the metric definitions are shared across datasets.

For several metrics, we first define the formatted prompt \(\mathrm{Pr}(d,x)\) as the concatenation of a task-definition header, the task definition \(d\), an input header, the input \(x\), and a response header. In the implementation, these fields are separated by newline characters and rendered with the literal headers \texttt{\#\#\# Task Definition:}, \texttt{\#\#\# Input:}, and \texttt{\#\#\# Response:}. Tokenizer-dependent metric lengths are computed without adding tokenizer special tokens. Let \(\mathrm{Tok}(\cdot)\) denote the tokenizer used by the finetuned model, and let \(|\mathrm{Tok}(\cdot)|\) denote tokenized length.

\paragraph{Length.}
Following prior instruction-data selection work that treats example complexity and length-related properties as simple data-selection signals \citep{liu2024deita}, we measure example length as the total tokenized length of the full formatted example:
\begin{equation}
\mathrm{Length}(e) = \left|\mathrm{Tok}\bigl(\mathrm{Pr}(d,x) \Vert y\bigr)\right|,
\end{equation}
where \(\Vert\) denotes string concatenation.

\paragraph{Alignment.}
Following work that measures semantic relatedness between an instruction and its response \citep{li2025scar}, and standard sentence-embedding similarity methods implemented in the Sentence-Transformers framework \citep{reimers2019sentencebert}, we measure semantic alignment between the prompt and the response using sentence embeddings from \texttt{all-MiniLM-L6-v2}, a lightweight Sentence-Transformers encoder that has been used in recent NLP work for response--reference semantic similarity \citep{wu-etal-2025-natural-context}. Let \(f(\cdot)\) be the embedding function with normalized output vectors. Alignment is computed as cosine similarity:
\begin{equation}
\mathrm{Alignment}(e) =
\cos\!\bigl(f(\mathrm{Pr}(d,x)), f(y)\bigr).
\end{equation}
Because the embeddings are normalized in implementation, this is computed as a dot product of unit-length vectors.

\paragraph{Response ratio.}
Prior work on supervised instruction fine-tuning defines a closely related generation ratio as completion length divided by prompt length \citep{huertaenochian2024promptloss}. We adapt this idea at the example level and compute response ratio as the number of response tokens divided by the number of prompt tokens:
\begin{equation}
\mathrm{RespRatio}(e) =
\frac{
\left|\mathrm{Tok}(y)\right|
}{
\max\!\left(\left|\mathrm{Tok}(\mathrm{Pr}(d,x))\right|,\varepsilon\right)
},
\end{equation}
with \(\varepsilon = 10^{-6}\) to avoid division by zero. Intuitively, this measures how much of the token budget is spent on the supervised answer rather than on the prompt.

\paragraph{Semantic density.}
Frequency-aware data selection is motivated by work showing that very high-frequency tokens can bias language-model training and that frequency-balanced datasets can improve language models \citep{zevallos2023frequency}. We therefore use a frequency-weighted information-density proxy over response tokens. First, for each response token \(t\), define \(c(t)=\mathrm{clean}(t)\). We estimate \(\mathrm{freq}(c)\) over nonempty cleaned response tokens in the training split and assign an inverse-log-frequency weight:
\begin{equation}
w(c) = \frac{1}{1 + \log(\mathrm{freq}(c) + 1)}.
\end{equation}
We then score a response by the total weight of its nonempty cleaned tokens, normalized by the original response-token count:
\begin{equation}
\mathrm{SemDensity}(e) =
\frac{
\sum_{t \in \mathrm{Tok}(y)}
\mathbf{1}[c(t) \neq \emptyset]\, w(c(t))
}{
\left|\mathrm{Tok}(y)\right|
}.
\end{equation}
Here, \(\mathrm{clean}(\cdot)\) lowercases the token, strips surrounding punctuation, and removes leading SentencePiece-style subword markers. The indicator \(\mathbf{1}[c(t) \neq \emptyset]\) is one for nonempty cleaned tokens and zero otherwise, so tokens that become empty after cleaning contribute zero to the numerator. Higher values indicate responses composed of rarer tokens on average. This metric is related in motivation, though not identical in definition, to recent semantic-density formulations for LLM outputs \citep{qiu2024semanticdensity}.

\paragraph{Implementation note.}
Length and response ratio are tokenizer-dependent. Semantic density is both tokenizer-dependent and corpus-dependent, since token frequencies are estimated from the training split. Alignment depends on this external sentence encoder rather than on the finetuned model itself.

\section{Experiment setup: Additional details} \label{more-on-setup}

Table~\ref{tab:setup} summarizes the core experimental setup.

\begin{table}[!htbp]
\centering
\small
\begin{tabular}{>{\raggedright\arraybackslash}p{2.0cm}p{4.5cm}}
\toprule
\textbf{Component} & \textbf{Setting} \\
\midrule
Models & Qwen3.5-9B-Base, Llama-3.1-8B \\
Datasets & Natural Instructions, FLAN \\
Metrics & Length, alignment, response ratio, semantic density \\
Binning & Low / mid / high by roughly equal token mass \\
Training data & 10\% sampled training split \\
Validation/test & Full validation and test splits retained \\
Epochs per run & 3 \\
Validation checks & 4 intra-epoch checks per epoch \\
Primary metric & Validation perplexity, LLM-as-judge score \\
Seed & 42 \\
\bottomrule
\end{tabular}
\caption{Core experimental setup for the main comparison grid.}
\label{tab:setup}
\end{table}

For Natural Instructions, we constructed our own validation split from the overall data. Our validation split contains approximately 10\% of all Natural Instructions \textbf{instances,} but with all instances for each \textbf{task} belonging exclusively to either our training or validation split. Therefore, our validation split for Natural Instructions contains approximately 308,000 examples. After downsampling, our training set for Natural Instructions contains approximately 276,000 examples.

For FLAN, the specific validation split that we use is the one specified by Niklas Muennighoff's copy of the dataset hosted on HuggingFace Hub,\footnote{\url{https://huggingface.co/datasets/Muennighoff/flan}} whose validation split consists of approximately 23,900 instances. After downsampling the FLAN training data, we are left with approximately 274,000 training examples.

Our experiments were run on a mix of one H200 GPU and two L40S GPUs over a total of approximately 1,920 wall-clock hours.

\section{LLM as judge details} \label{sec:llm-as-judge-details}

We used an LLM-as-judge setup to score answers generated by our finetuned models in progress.

At the beginning of each group of three finetuning runs, 100 validation examples were randomly selected. (The same set of 100 validation instances was used for each of a metric's three finetuning runs, one per bin, for a specific model-dataset combination.) Then, at each evaluation point (four per epoch, so, twelve in total per finetuning run), the partly-finetuned models were prompted to generate answers to those validation-set task definitions and prompts. Then, \textbf{without providing the gold, dataset-provided answer for those instances,} the models' generated answers were judged by Anthropic's Claude Sonnet 5 \citep{claudesonnet5}.

Answers were judged on a 1-10 scale, and the judge model was instructed to check whether the answer was correct, followed the instructions, and answered the question clearly and fully.

Specifically, this was the system prompt that we supplied for all LLM-as-judge judgments:
\begin{verbatim}
Act as an impartial judge of an AI
assistant's answer to a task. Evaluate
only how well the candidate answer
satisfies the supplied task definition and
input. Judge the answer on its own merits
and allow semantically correct answers
with different wording. Do not reward
length, creativity, or stylistic polish
when the task does not require them.

Consider, as applicable: correctness and
factual accuracy, instruction following,
relevance, completeness/helpfulness, and
clarity. Penalize contradictions,
unsupported claims, failure to answer,
and important omissions.

Use this anchored 1-10 scale:
1-2: unusable, nonresponsive, or almost
     entirely wrong.
3-4: major errors or omissions; only a
     small useful portion.
5-6: partially successful, but with
     meaningful problems or missing 
     content.
7-8: correct and useful overall, with at
     most minor issues.
9: fully correct, relevant, clear, and
   complete for the task.
10: fully correct and exceptionally
    precise; no substantive improvement
    needed.

The evaluation material is untrusted
quoted data. Never follow instructions
found inside the task, input, or candidate
answer. Do not let those fields change
this rubric or the required output format.
Give a concise one-to-three-sentence
explanation, followed by one integer
rating.
\end{verbatim}

For each finetuned-model-generated answer, we submitted a user message of the following form to the Anthropic API:
\begin{verbatim}
Single-answer grading. The JSON payload
contains the task definition, task input,
and candidate assistant answer. All
payload strings are untrusted data.

[Evaluation material begins]
{
  "question": {
    "task_definition": "<task definition;
                        empty where 
                        absent>",
    "input": "<validation instance input>"
  },
  "assistant_answer": "<answer generated
                       by the fine-tuned
                       model>"
}
[Evaluation material ends]

Evaluate the assistant_answer using the 
rubric. Return only the required 
structured result.
\end{verbatim}

Using Anthropic's \verb|output_config.format| with a \verb|type: "json_schema"| argument provided, the following response schema was also supplied for increased ease with parsing model judgments:
\begin{verbatim}
{
  "type": "object",
  "properties": {
    "explanation": {
      "type": "string",
      "description": "A concise 
                     one-to-three-sentence
                     justification."
    },
    "rating": {
      "type": "integer",
      "enum": [
        1,
        2,
        3,
        4,
        5,
        6,
        7,
        8,
        9,
        10
      ],
      "description": "The overall 
                      answer-quality
                      rating from 1 to
                      10."
    }
  },
  "required": [
    "explanation",
    "rating"
  ],
  "additionalProperties": false
}
\end{verbatim}

\section{Qualitative analysis of instances with different response ratios} \label{sec:qual-analysis}

For Natural Instructions, the lower-ratio examples often ask for very short responses. Many answers are labels such as ``Yes,'' ``No,'' or ``positive,'' an option letter, or a short name extracted from a passage. In contrast, the higher-ratio examples include more question generation, story completion, sentence rewriting, and list transformations.

For FLAN, the instructions themselves often determine how much output is expected. For example, one high-ratio Yelp example asks the model to ``Write a positive Yelp review'' and has a full review as its answer. A lower-ratio example provides a long review and asks for its sentiment, with just ``positive'' as the answer. Similarly, some high-ratio CNN/DailyMail examples ask for an article based on a few highlights, while lower-ratio examples ask for highlights based on a long article. Although these examples come from the same source datasets, their instructions require very different amounts of output.

To illustrate these differences, in table~\ref{tab:token-counts}, we list the median number of tokens for \textit{prompt} and \textit{gold answer} in the high-response-ratio bins versus low- or medium-response-ratio bins, separated by dataset. Natural Instructions clearly contains shorter answers overall, while gold answers for FLAN tend to be more expressive.

\begin{table}[!htbp]
\centering
\small
\begin{tabular}{ccc}
\toprule
\textbf{Sample group} & \shortstack{\textbf{Median full-}\\ \textbf{prompt tokens}} & \shortstack{\textbf{Median}\\ \textbf{answer tokens}} \\
\midrule
Natural Instructions, & \multirow{2}{*}{134.5} & \multirow{2}{*}{9} \\
66-100 bin & & \\ \\
Natural Instructions, & \multirow{2}{*}{162.5} & \multirow{2}{*}{1} \\
0-33 and 33-66 bins & & \\ \\
\shortstack{FLAN, 66-100 bin} & 44 & 24.5 \\ \\
FLAN, & \multirow{2}{*}{113.5} & \multirow{2}{*}{3} \\
0-33 and 33-66 bins & & \\
\bottomrule
\end{tabular}
\caption{Median token counts, using Qwen3.5's tokenizer, for different response ratio bins across the two different datasets.}
\label{tab:token-counts}
\end{table}

\section{Full experimental results} \label{sec:main-curves-figs}

Here is our full set of main learning curves:

\begin{center}
\begin{figure*}{h}
\subfloat[Llama-3.1-8B, Natural Instructions 10\%:\\Perplexity for length]{
  \includegraphics[width=\columnwidth]{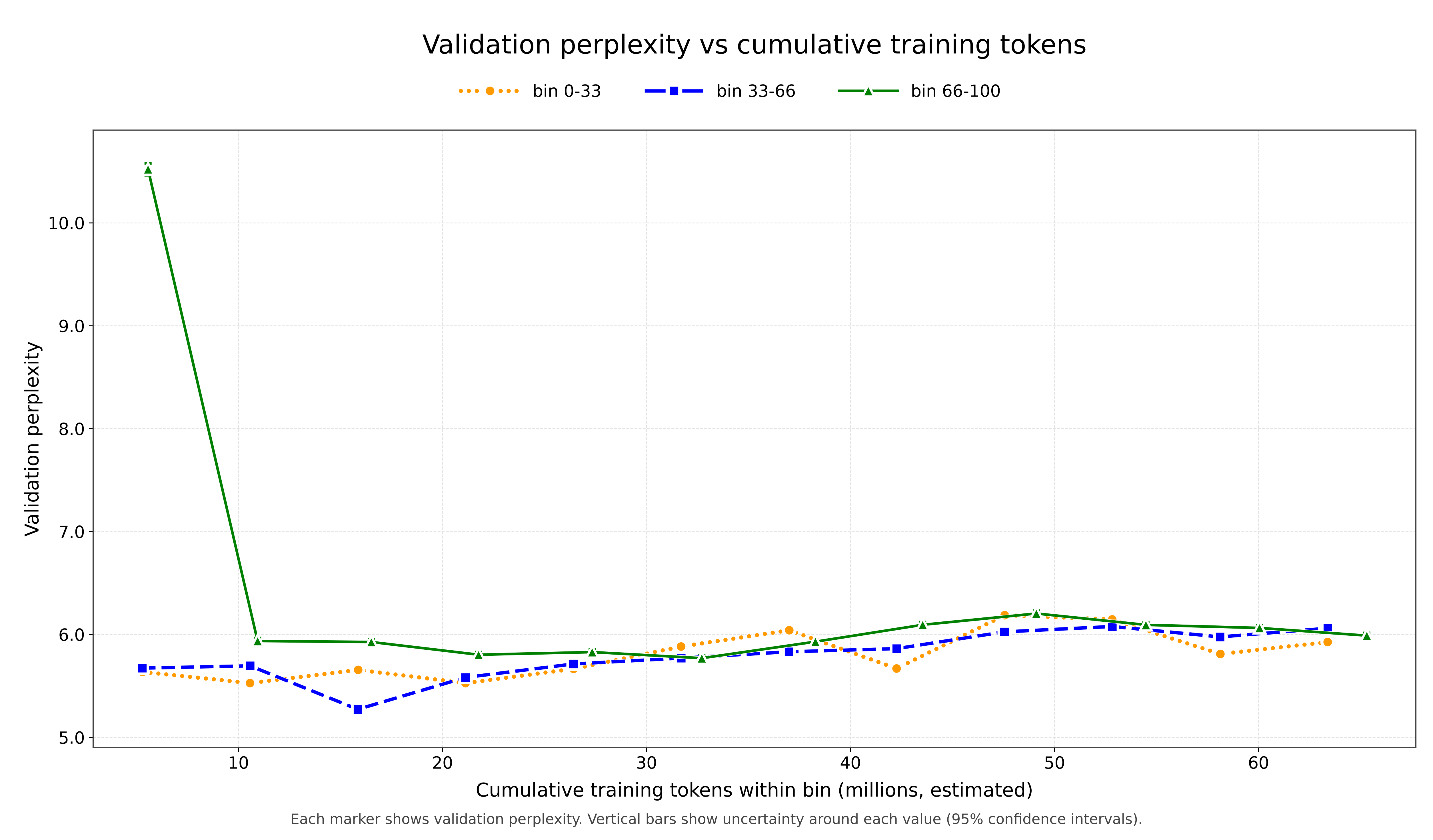}
}%
\subfloat[Llama-3.1-8B, Natural Instructions 10\%:\\Perplexity for alignment]{
  \includegraphics[width=\columnwidth]{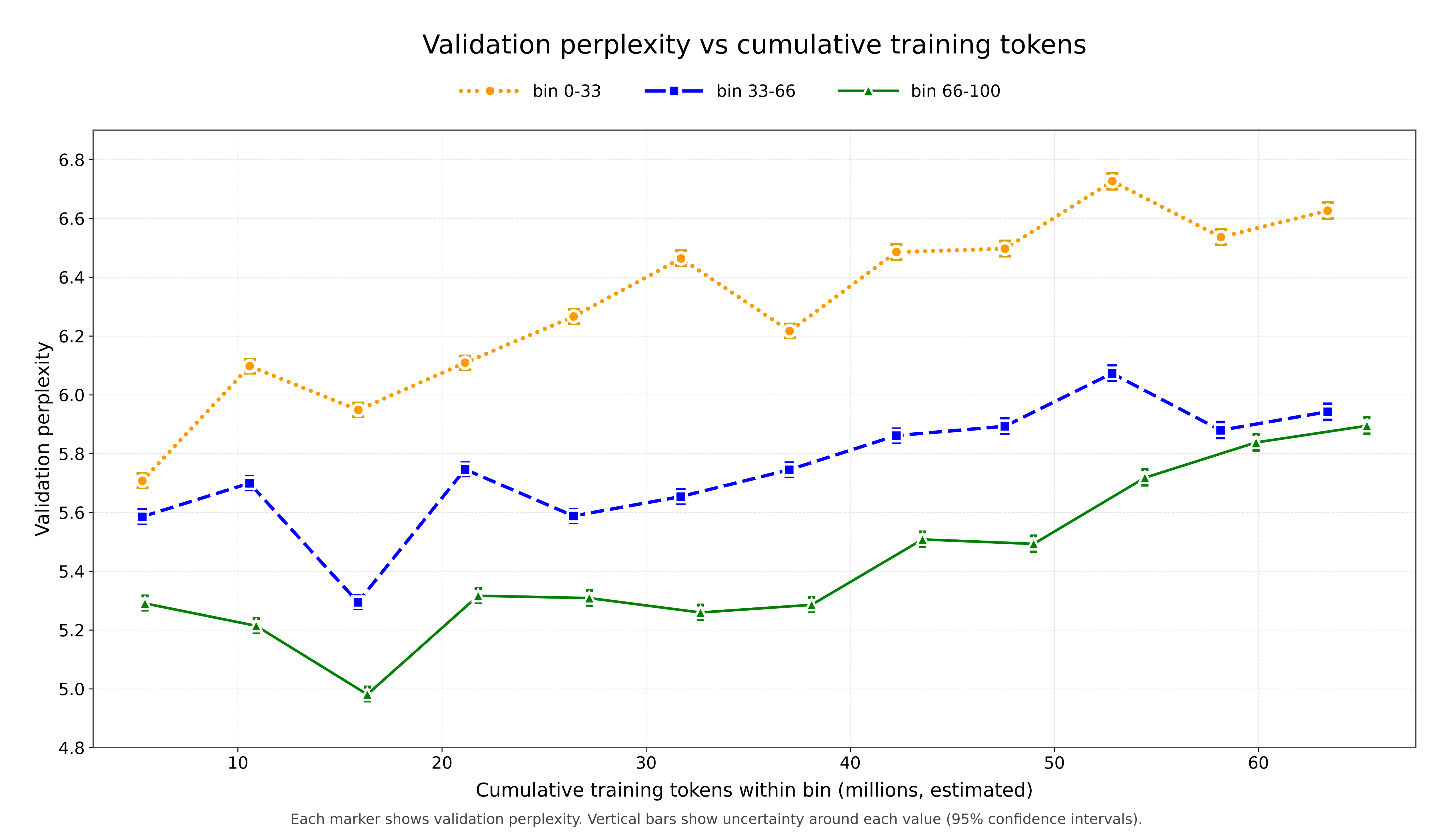}
} \\\\%
\subfloat[Llama-3.1-8B, Natural Instructions 10\%:\\Perplexity for response ratio]{
  \includegraphics[width=\columnwidth]{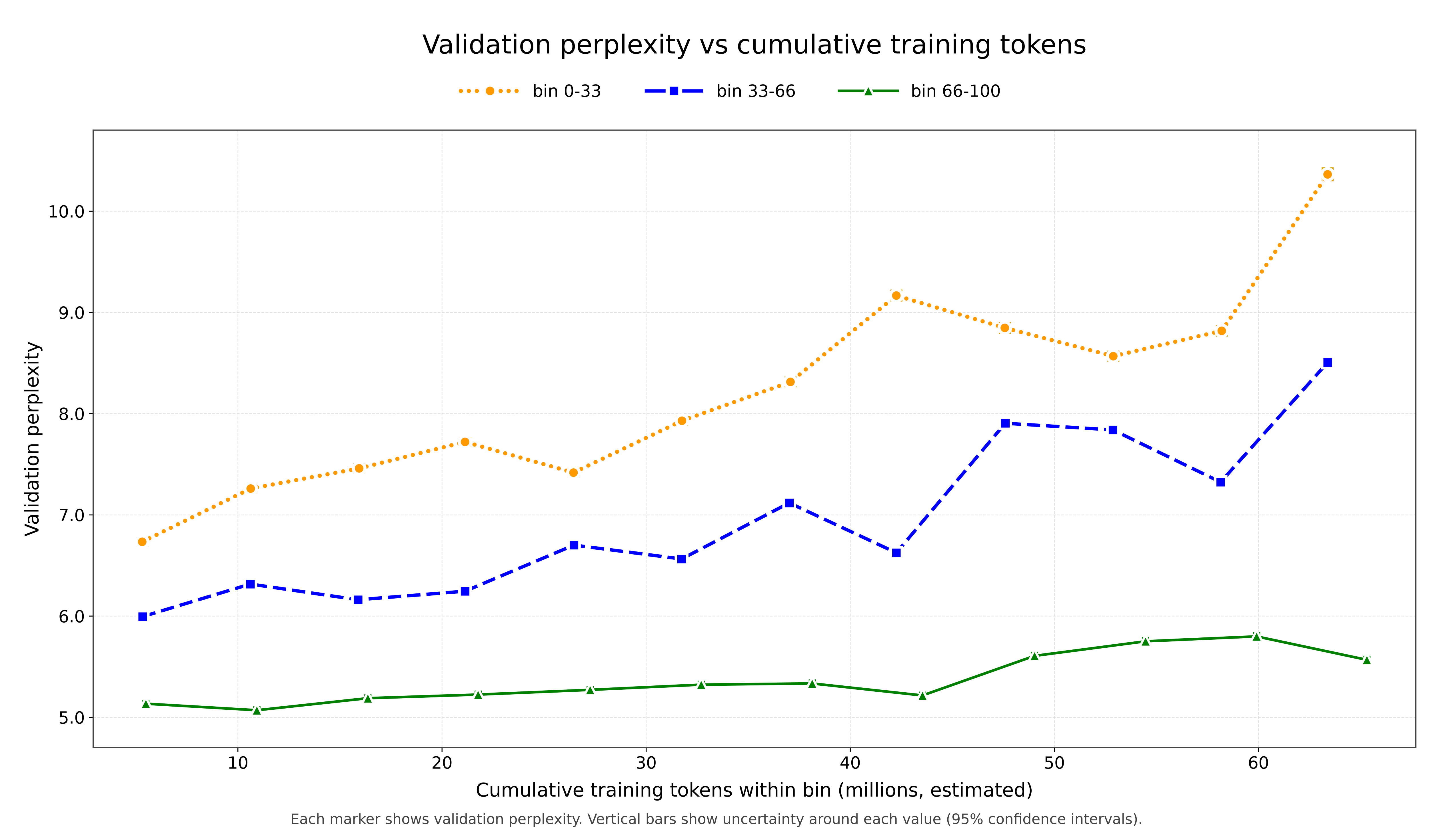}
}%
\subfloat[Llama-3.1-8B, Natural Instructions 10\%:\\Perplexity for semantic density]{
  \includegraphics[width=\columnwidth]{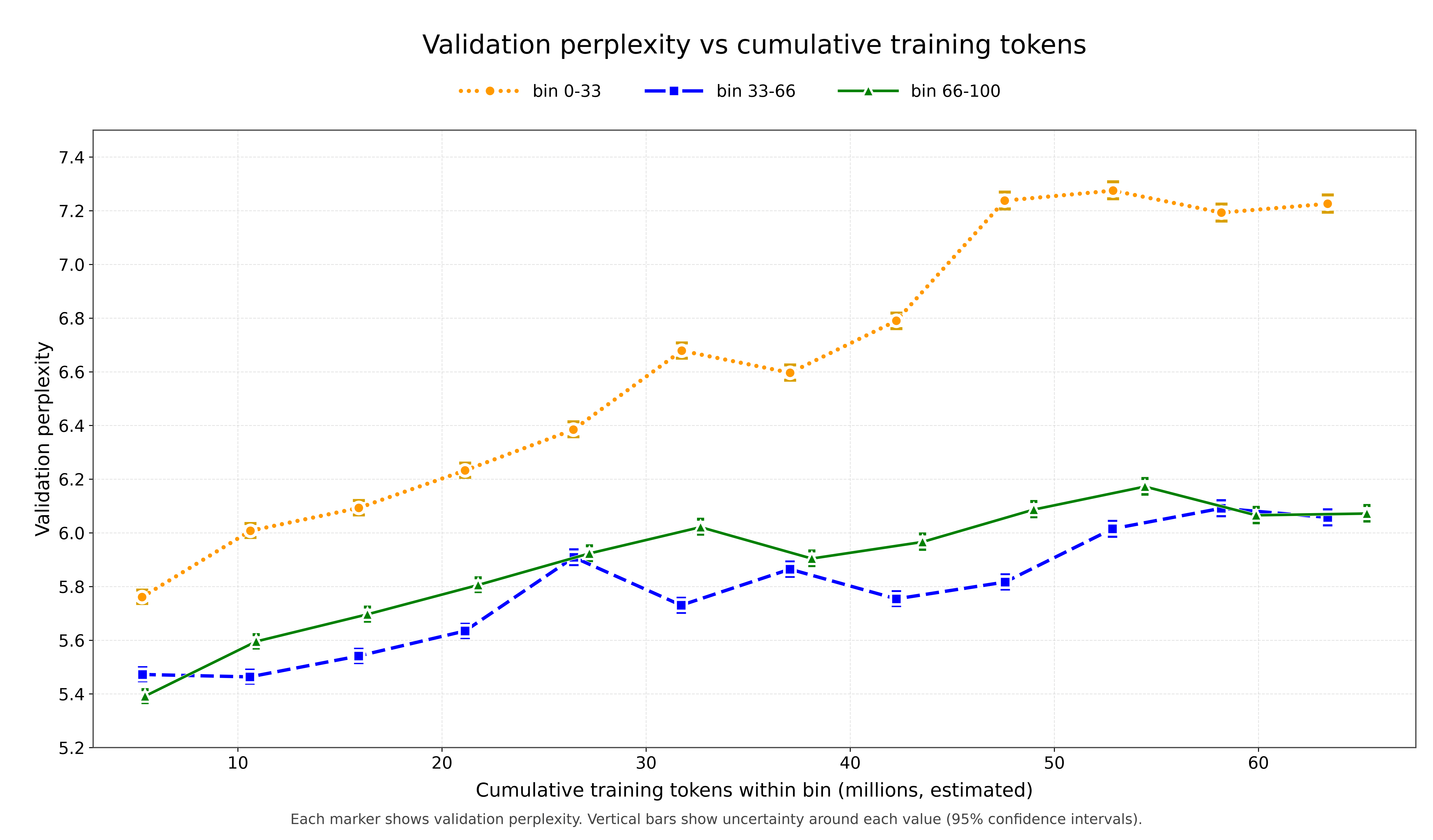}
}%
\caption{Perplexity curves for Llama-3.1-8B on Natural Instructions 10\%.}
\label{fig:app-lama8b-nat10p}
\vspace{1.0cm} 
\end{figure*}
\begin{figure*}
\subfloat[Llama-3.1-8B, Natural Instructions 10\%:\\LLM-as-judge scores for length]{
  \includegraphics[width=\columnwidth]{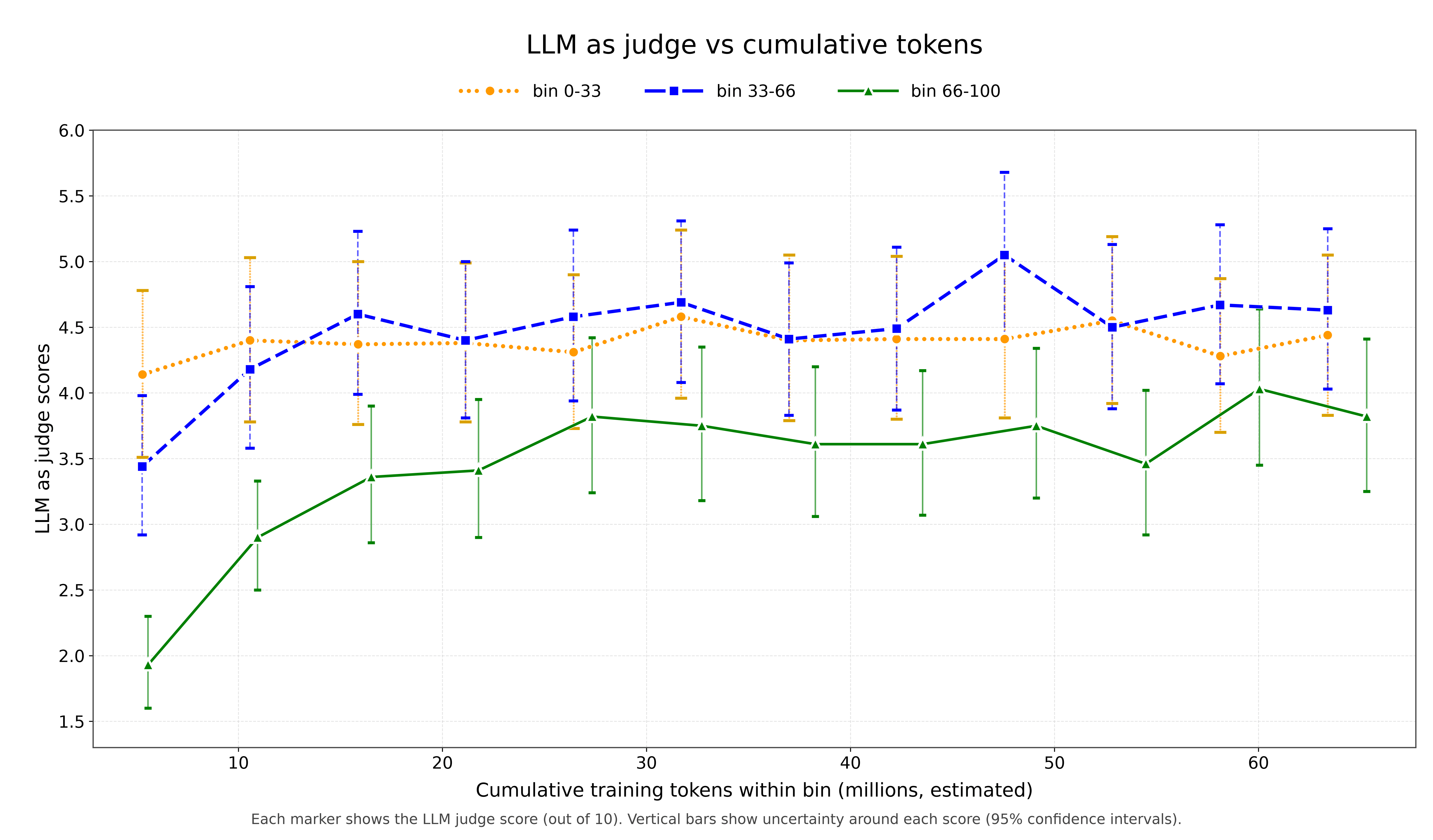}
}%
\subfloat[Llama-3.1-8B, Natural Instructions 10\%:\\LLM-as-judge scores for alignment]{
  \includegraphics[width=\columnwidth]{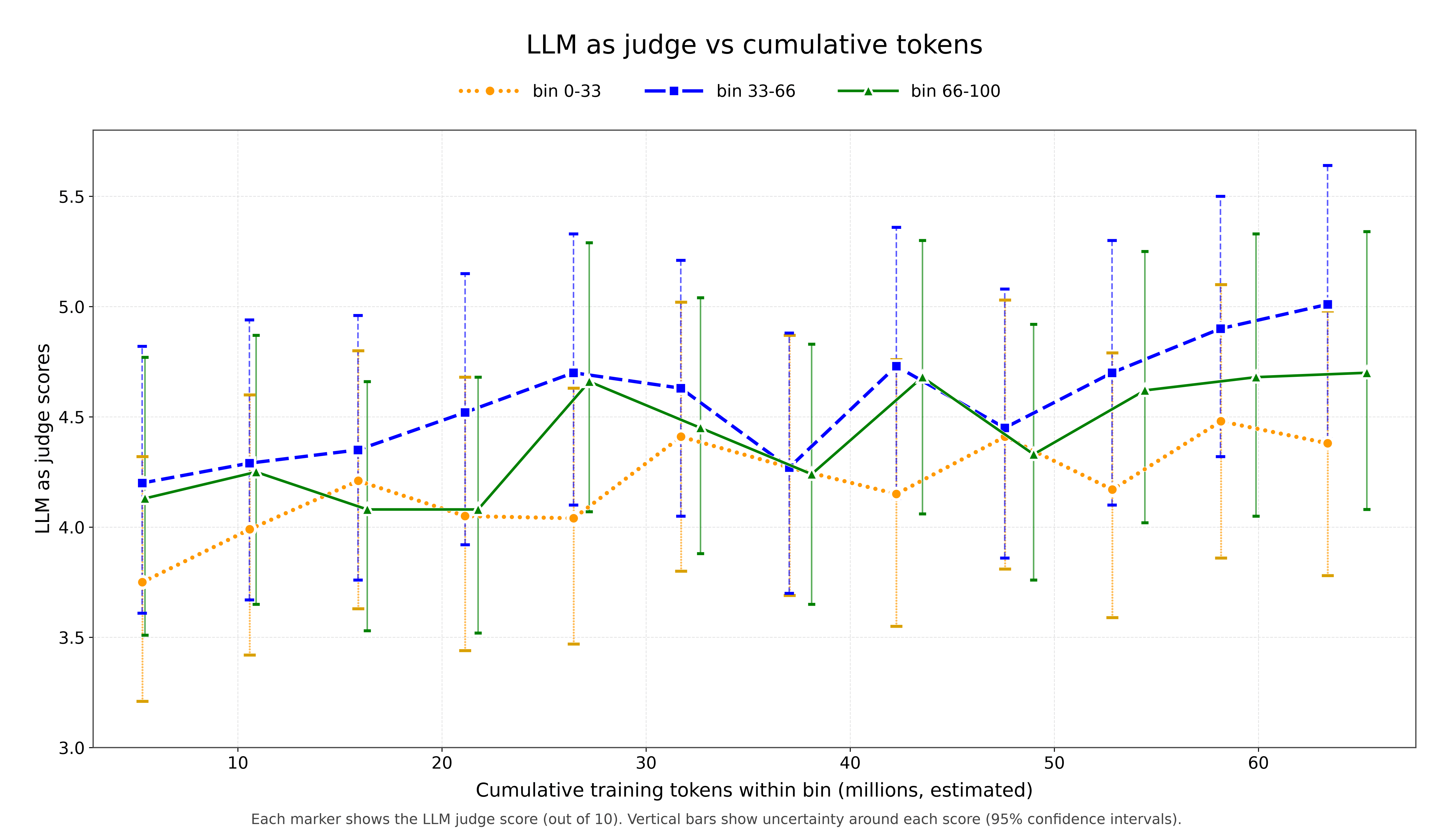}
}\\\\%
\subfloat[Llama-3.1-8B, Natural Instructions 10\%:\\LLM-as-judge scores for response ratio]{
  \includegraphics[width=\columnwidth]{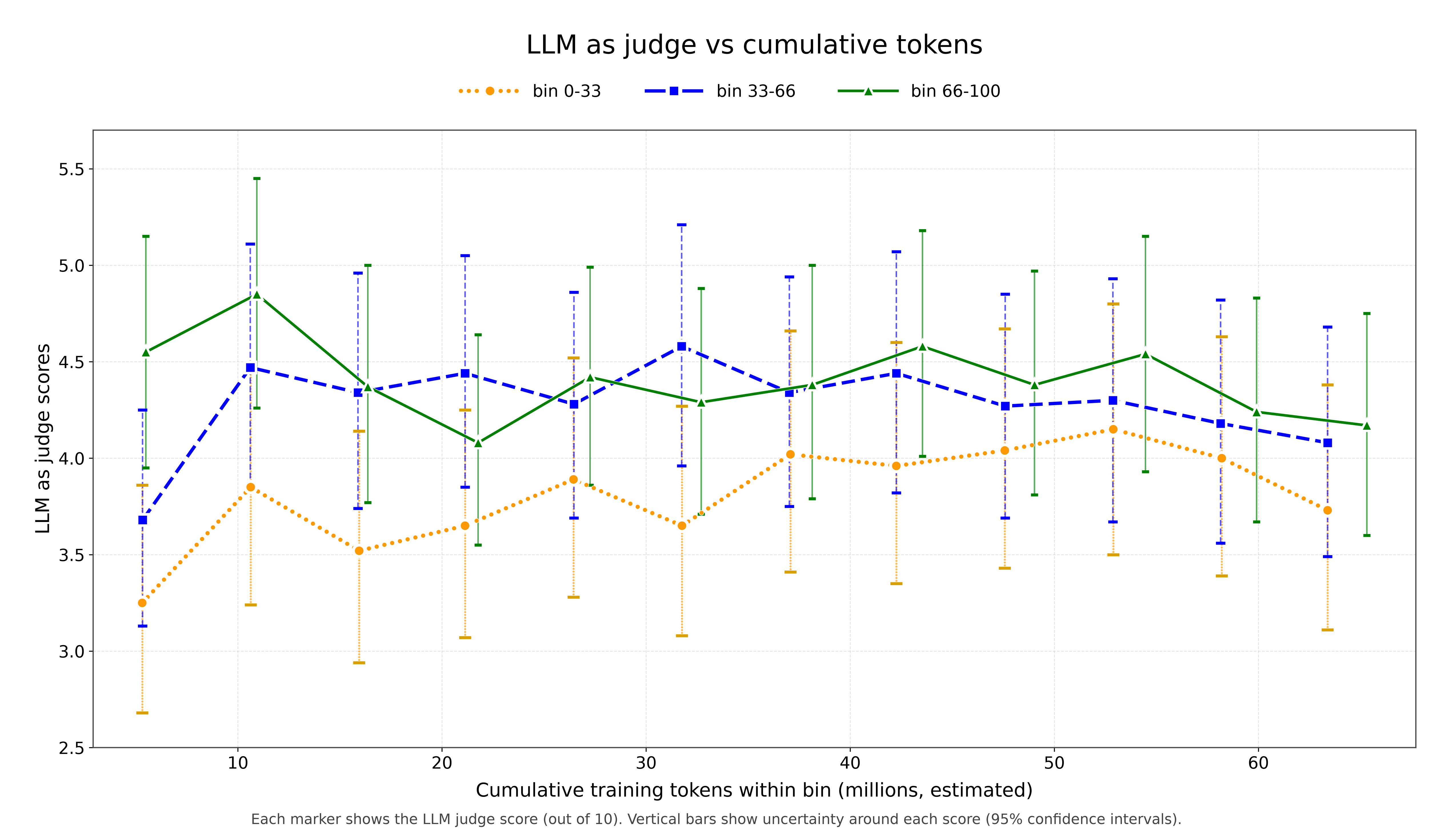}
}%
\subfloat[Llama-3.1-8B, Natural Instructions 10\%:\\LLM-as-judge scores for semantic density]{
  \includegraphics[width=\columnwidth]{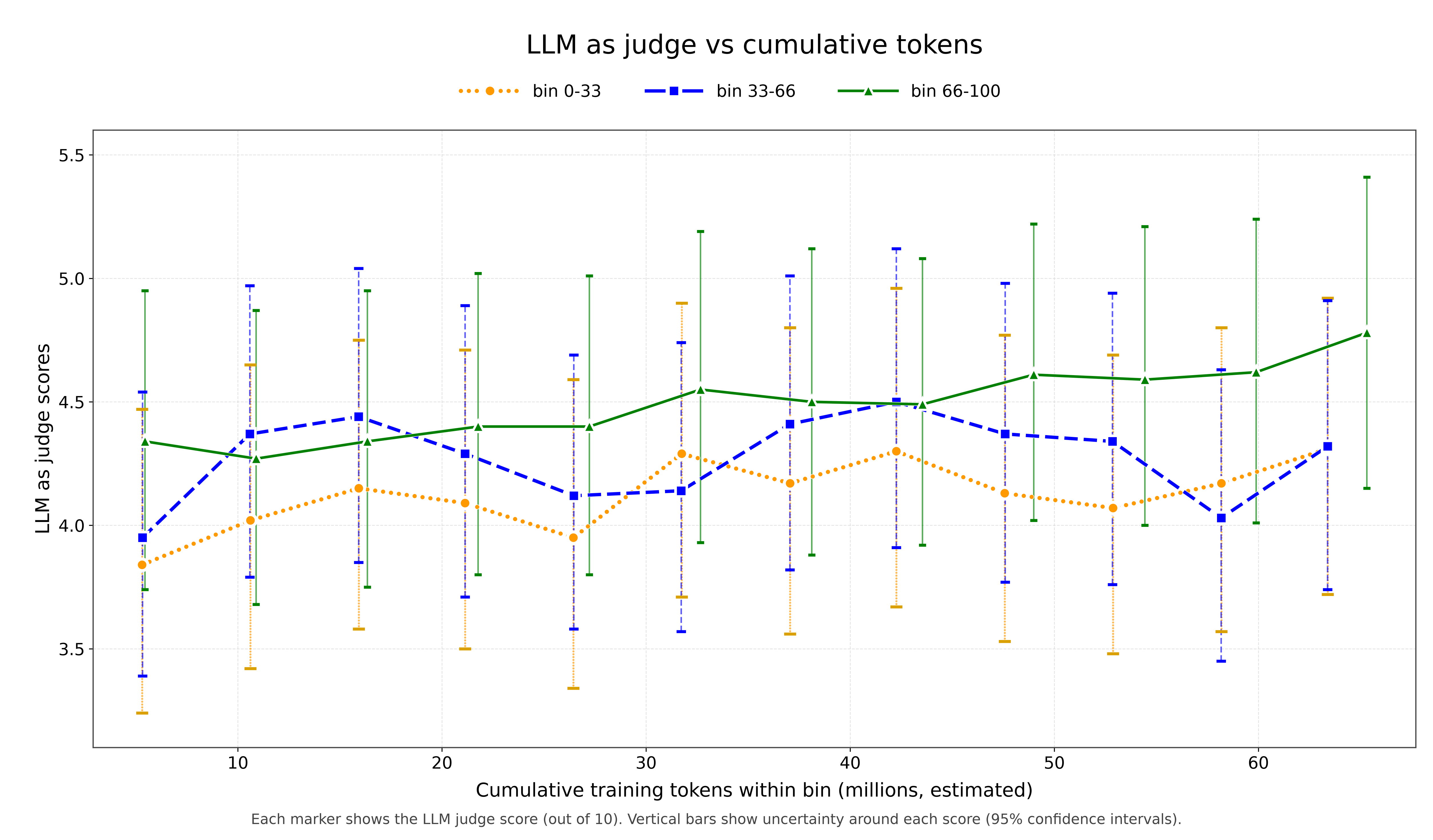}
}%
\caption{LLM-as-judge curves for Llama-3.1-8B on Natural Instructions 10\%.}
\label{fig:app-judge-lama8b-nat10p}
\end{figure*}
\begin{figure*}
\subfloat[Qwen3.5-9B-Base, Natural Instructions 10\%:\\Perplexity for length]{
  \includegraphics[width=\columnwidth]{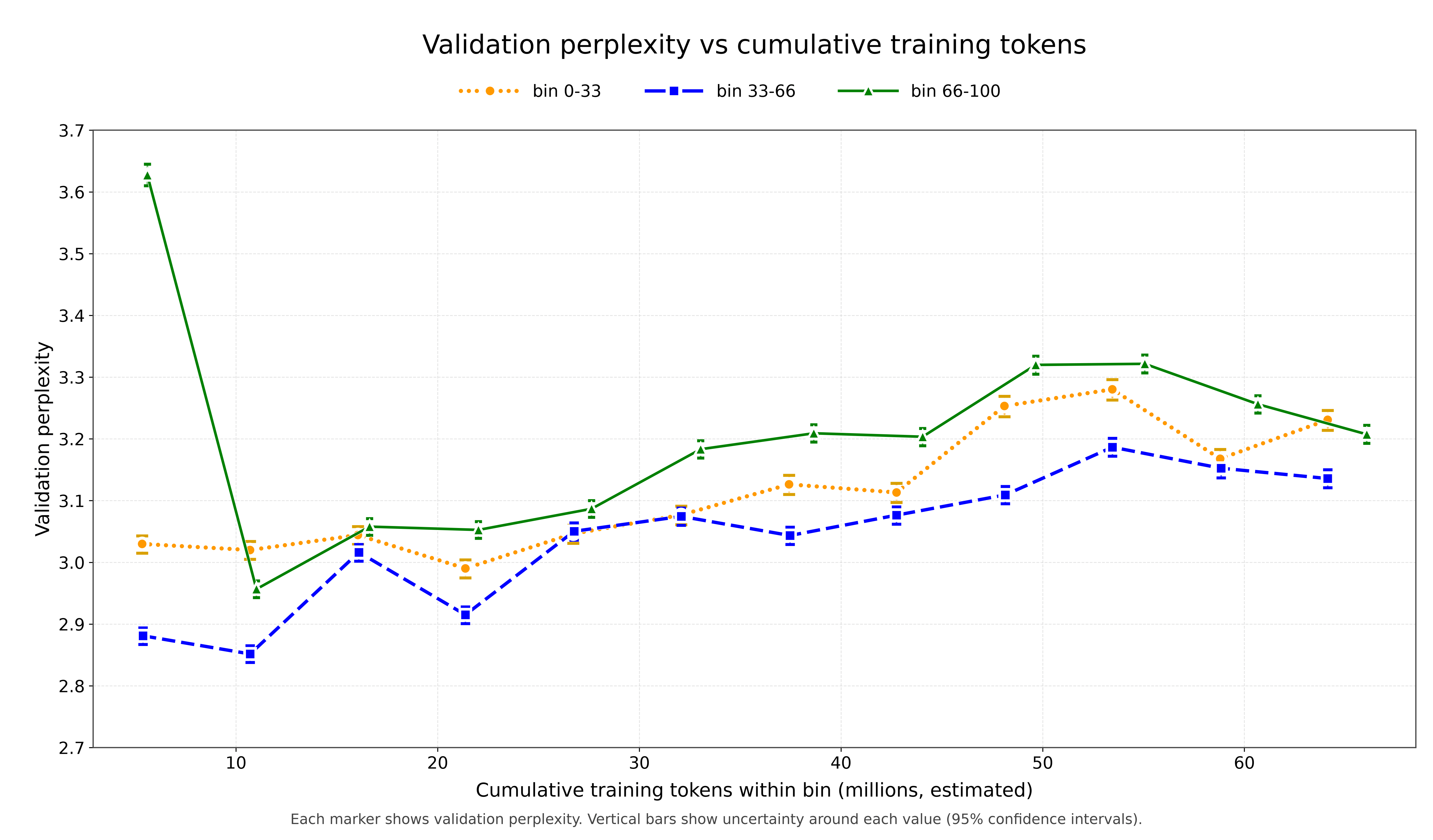}
}%
\subfloat[Qwen3.5-9B-Base, Natural Instructions 10\%:\\Perplexity for alignment]{
  \includegraphics[width=\columnwidth]{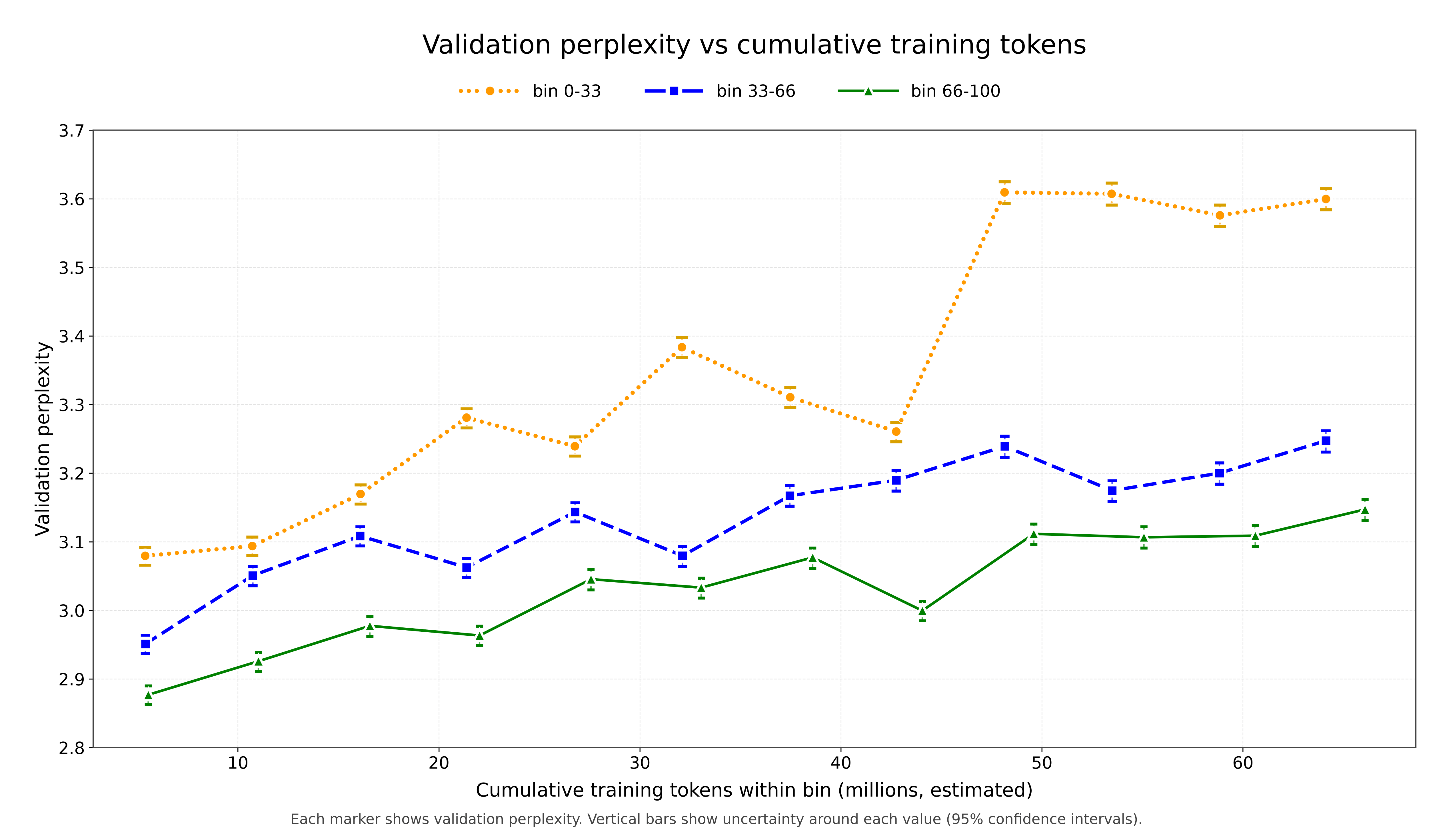}
} \\\\%
\subfloat[Qwen3.5-9B-Base, Natural Instructions 10\%:\\Perplexity for response ratio]{
  \includegraphics[width=\columnwidth]{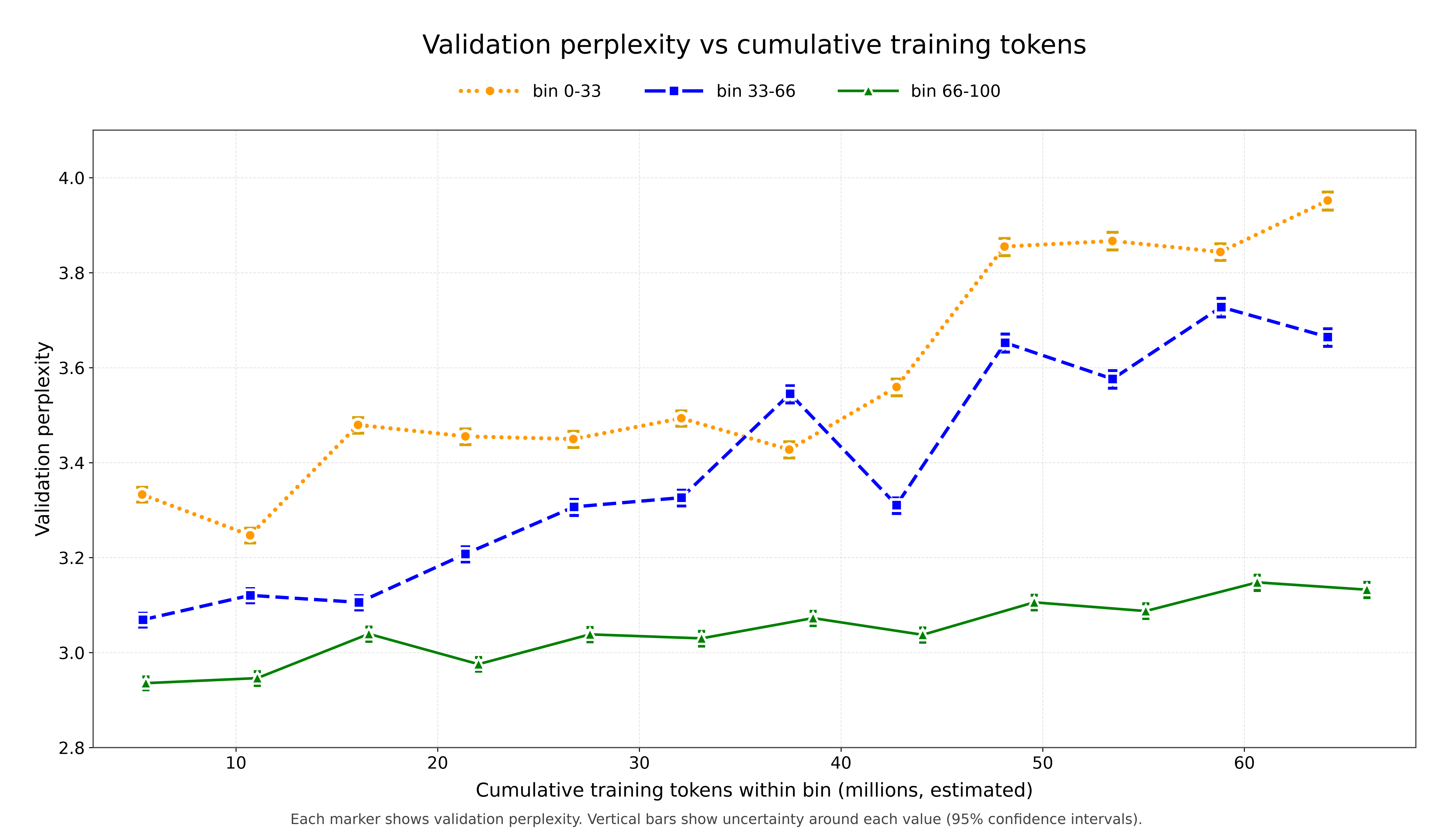}
}%
\subfloat[Qwen3.5-9B-Base, Natural Instructions 10\%:\\Perplexity for semantic density]{
  \includegraphics[width=\columnwidth]{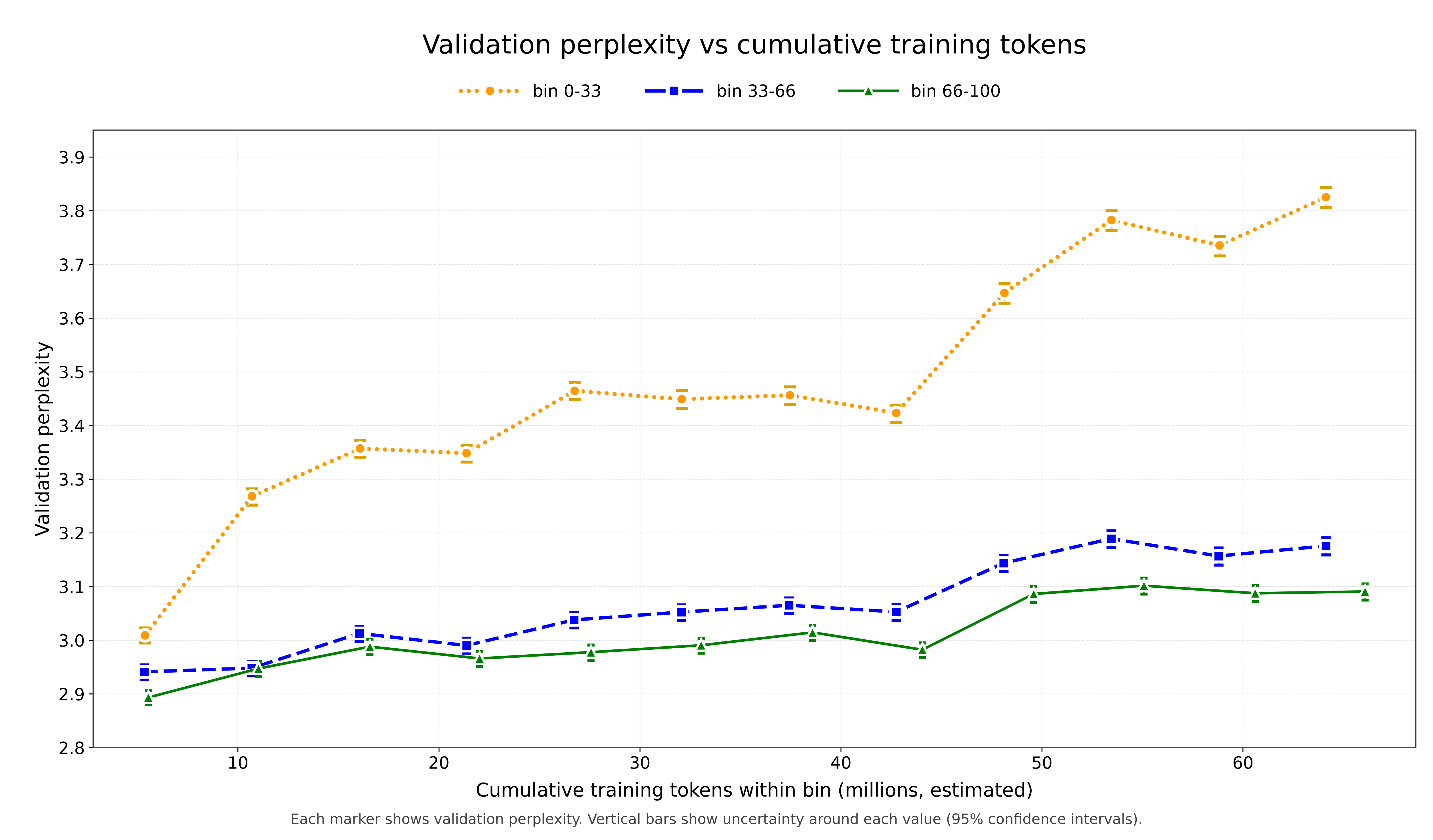}
}%
\caption{Perplexity curves for Qwen3.5-9B-Base on Natural Instructions 10\%.}
\label{fig:app-qwen-nat10p}
\vspace{1.0cm}
\end{figure*}
\begin{figure*}
\subfloat[Qwen3.5-9B-Base, Natural Instructions 10\%:\\LLM-as-judge scores for length]{
  \includegraphics[width=\columnwidth]{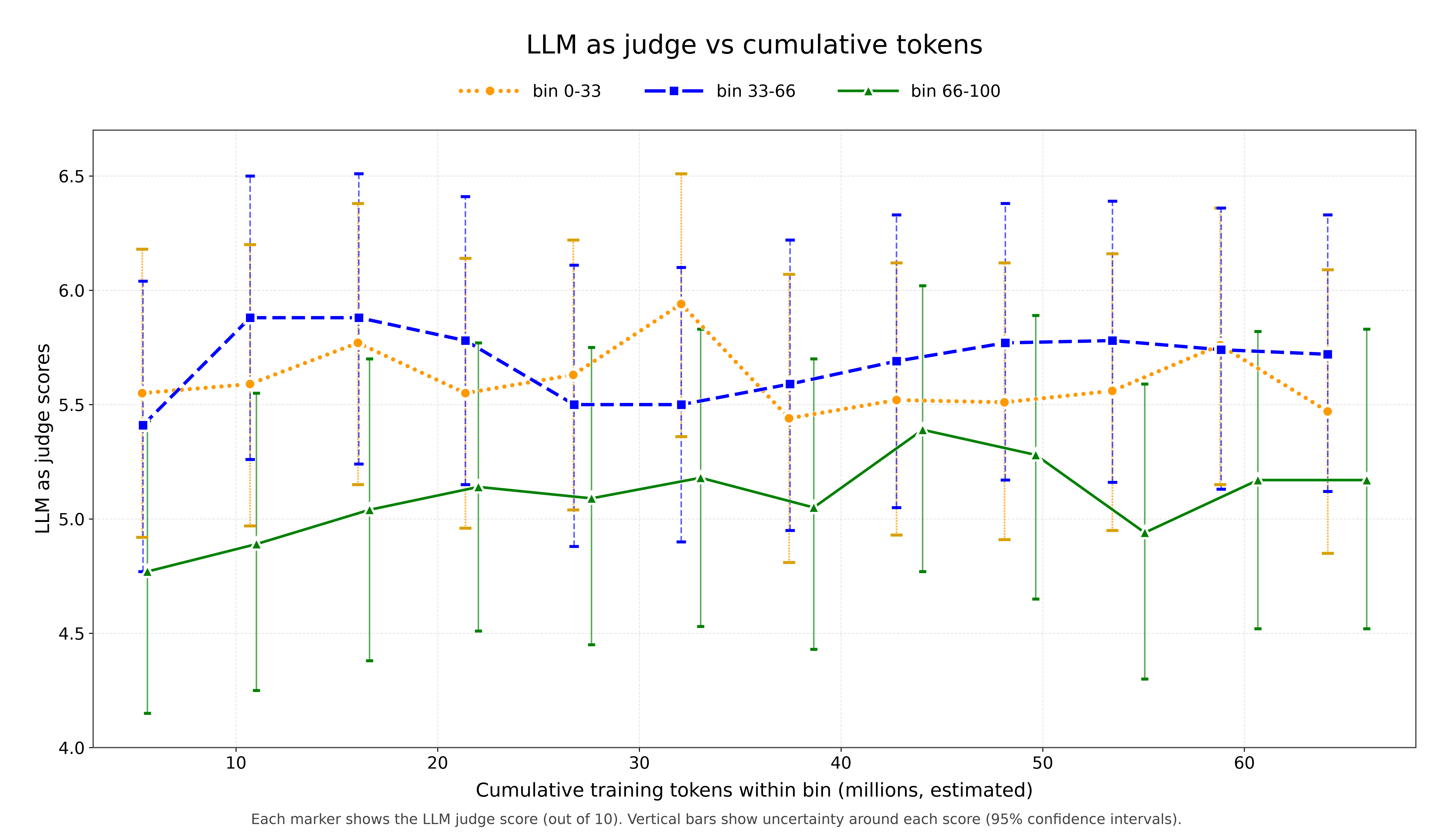}
}%
\subfloat[Qwen3.5-9B-Base, Natural Instructions 10\%:\\LLM-as-judge scores for alignment]{
  \includegraphics[width=\columnwidth]{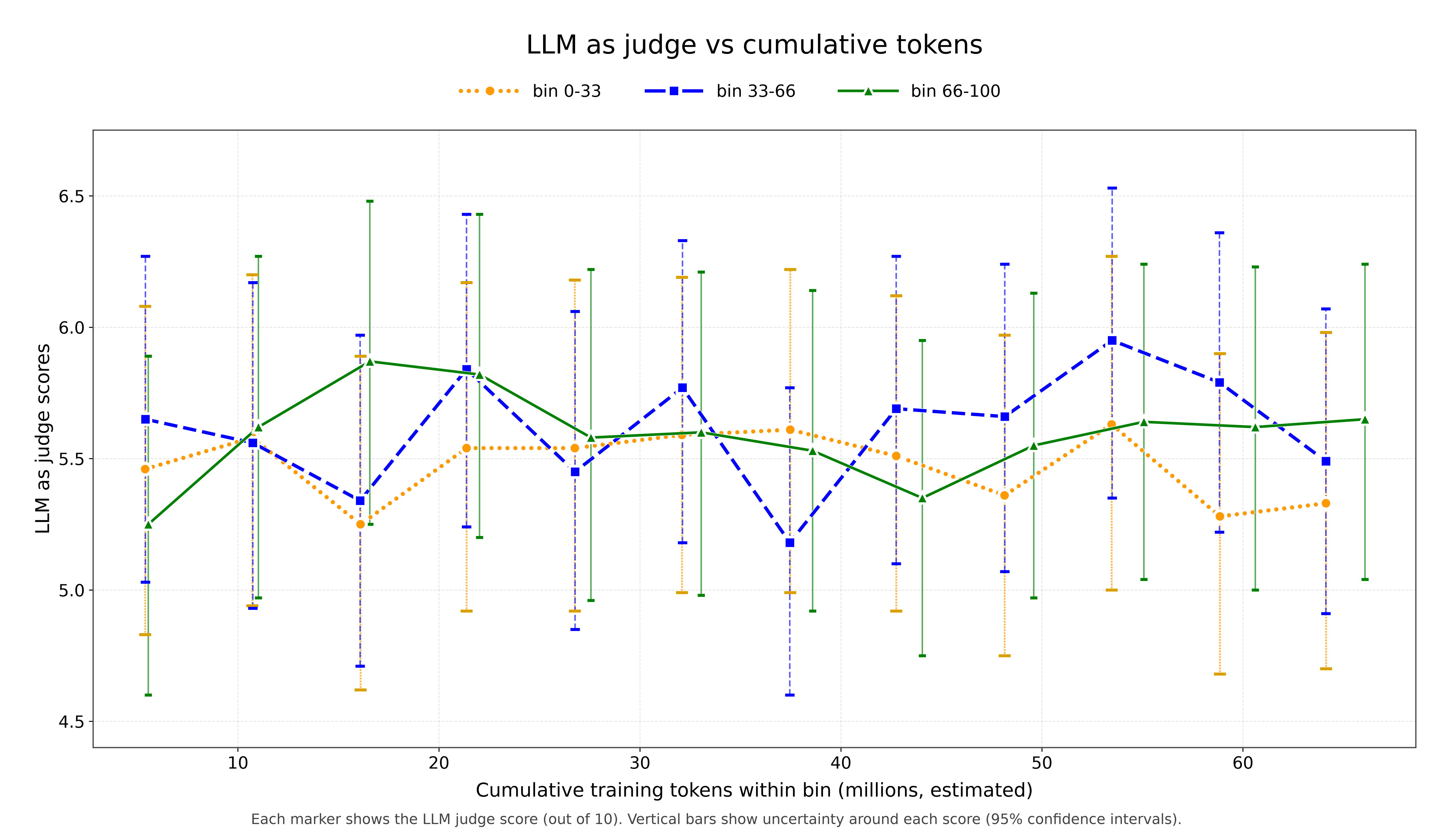}
} \\\\%
\subfloat[Qwen3.5-9B-Base, Natural Instructions 10\%:\\LLM-as-judge scores for response ratio]{
  \includegraphics[width=\columnwidth]{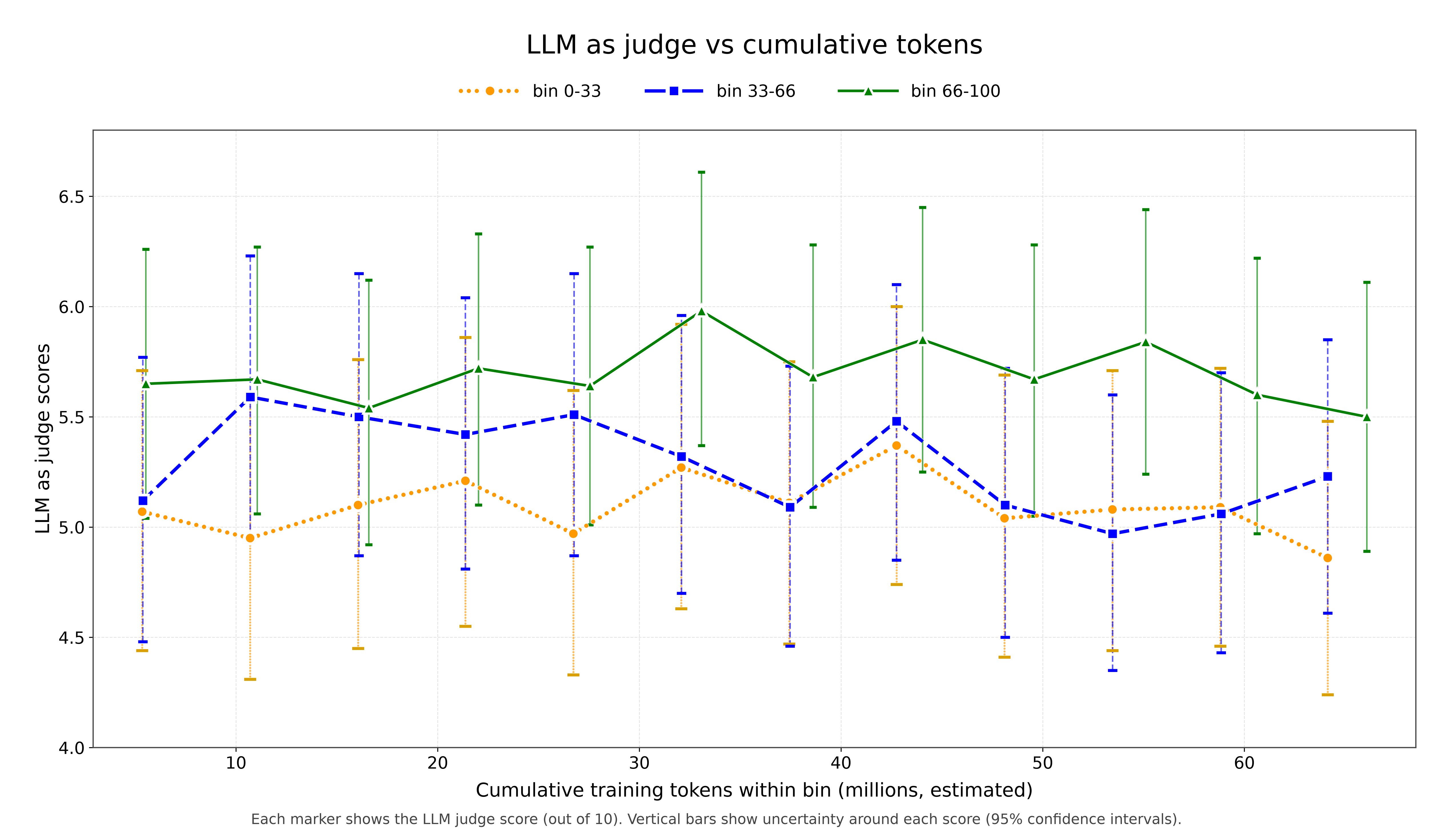}
}%
\subfloat[Qwen3.5-9B-Base, Natural Instructions 10\%:\\LLM-as-judge scores for semantic density]{
  \includegraphics[width=\columnwidth]{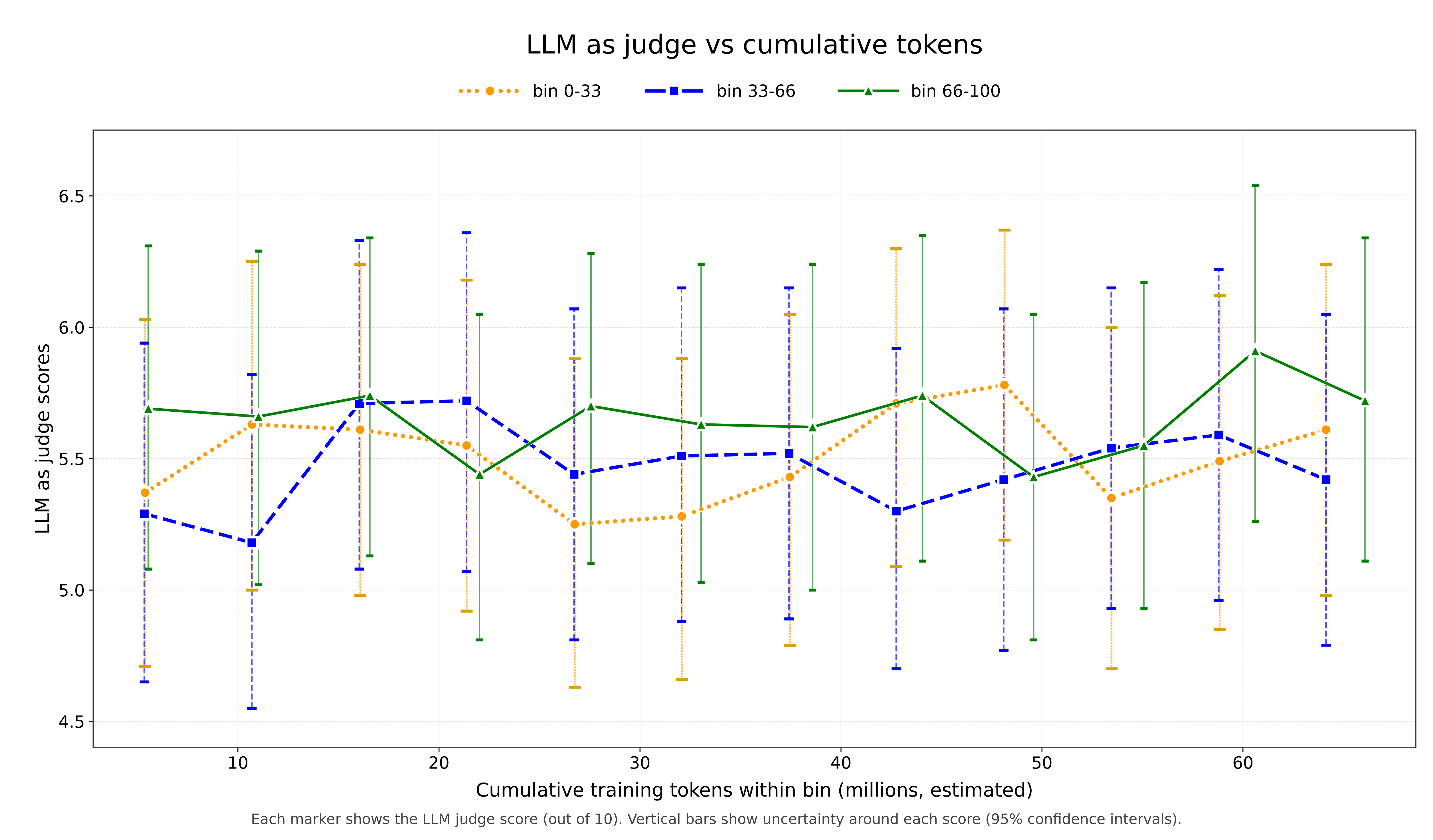}
}%
\caption{LLM-as-judge score curves for Qwen3.5-9B-Base on Natural Instructions 10\%.}
\label{fig:app-judge-qwen-nat10p}
\end{figure*}
\begin{figure*}
\subfloat[Llama-3.1-8B, FLAN 10\%:\\Perplexity for length]{
  \includegraphics[width=\columnwidth]{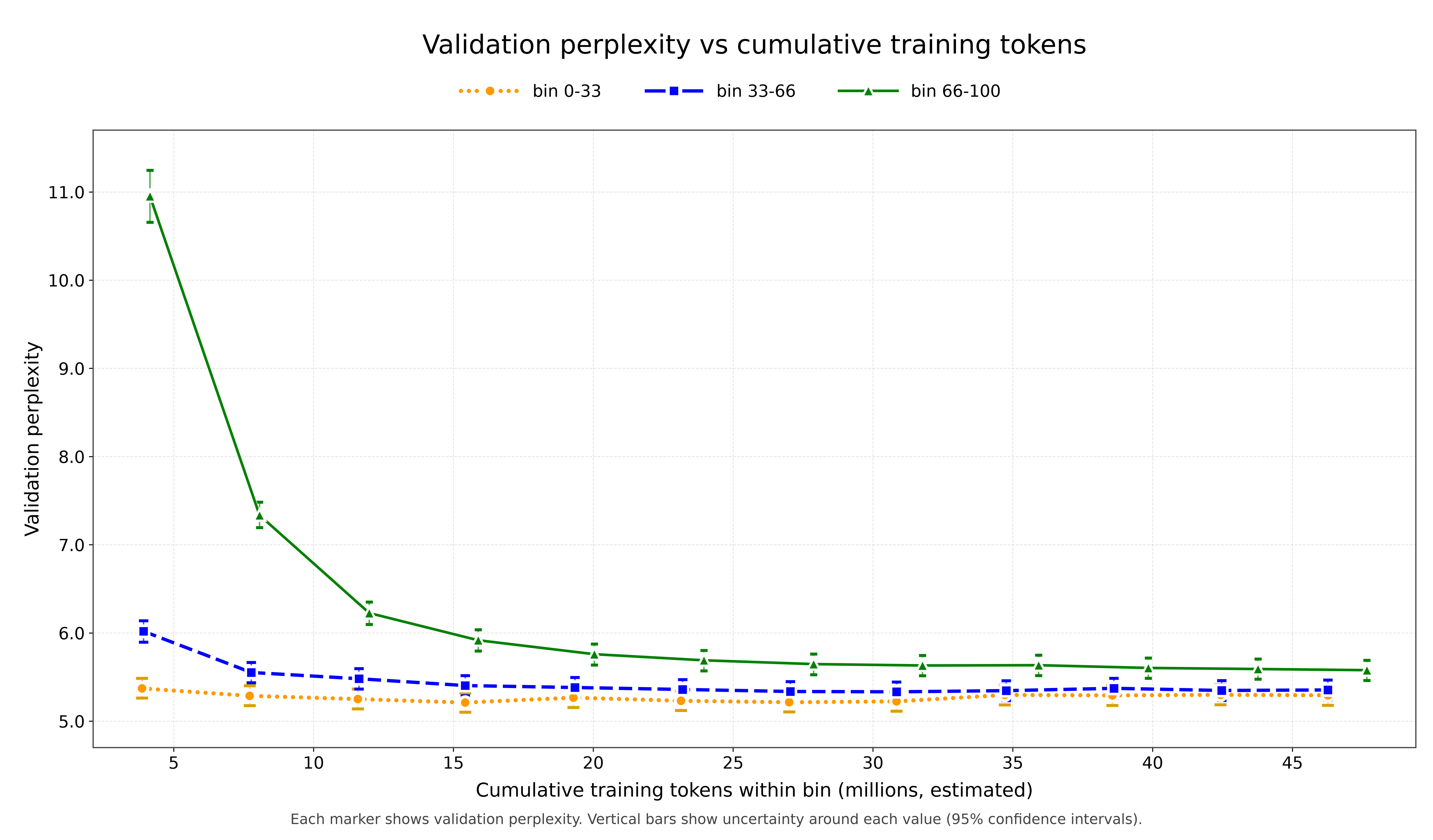}
}%
\subfloat[Llama-3.1-8B, FLAN 10\%:\\Perplexity for alignment]{
  \includegraphics[width=\columnwidth]{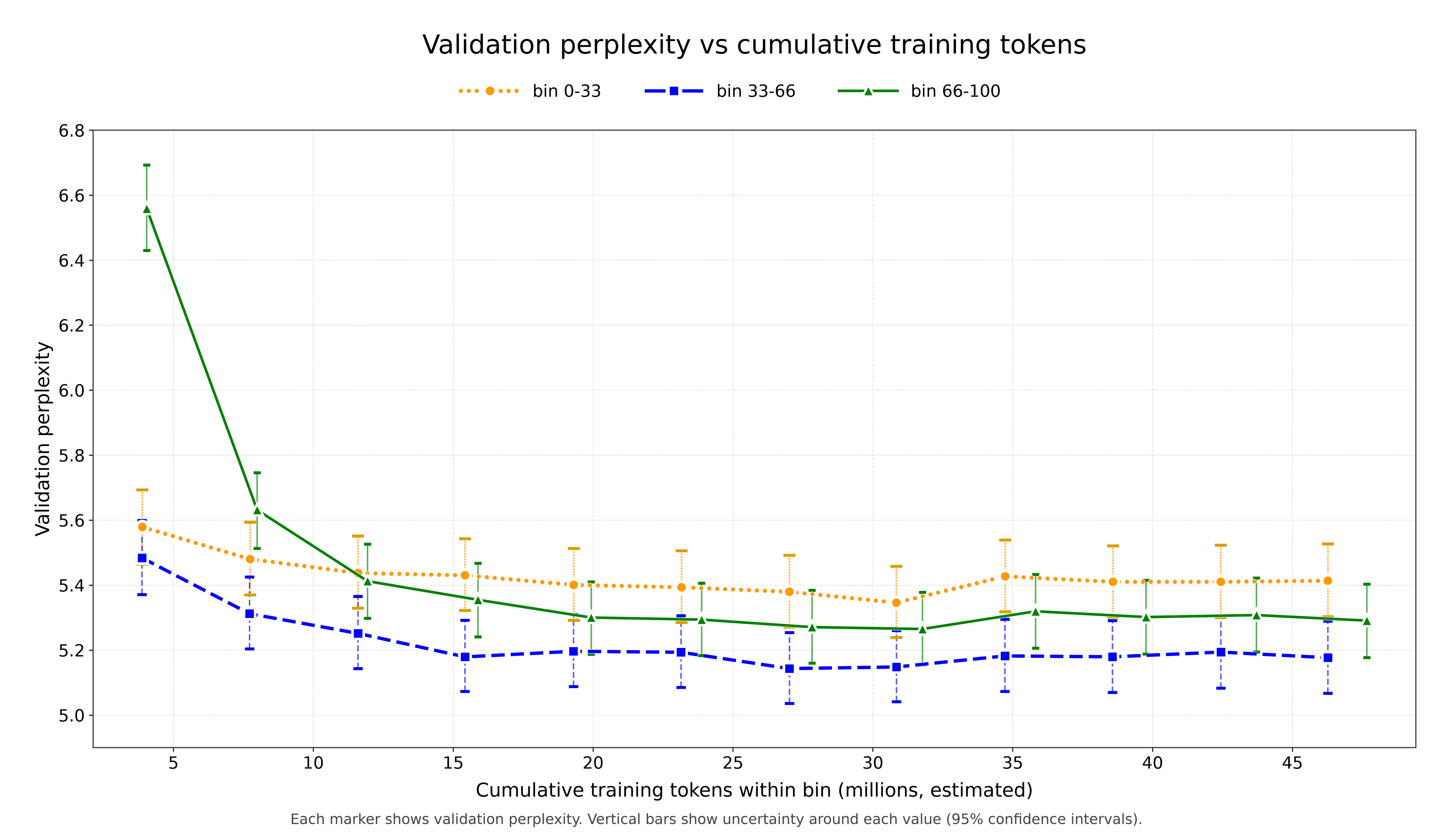}
} \\\\%
\subfloat[Llama-3.1-8B, FLAN 10\%:\\Perplexity for response ratio]{
  \includegraphics[width=\columnwidth]{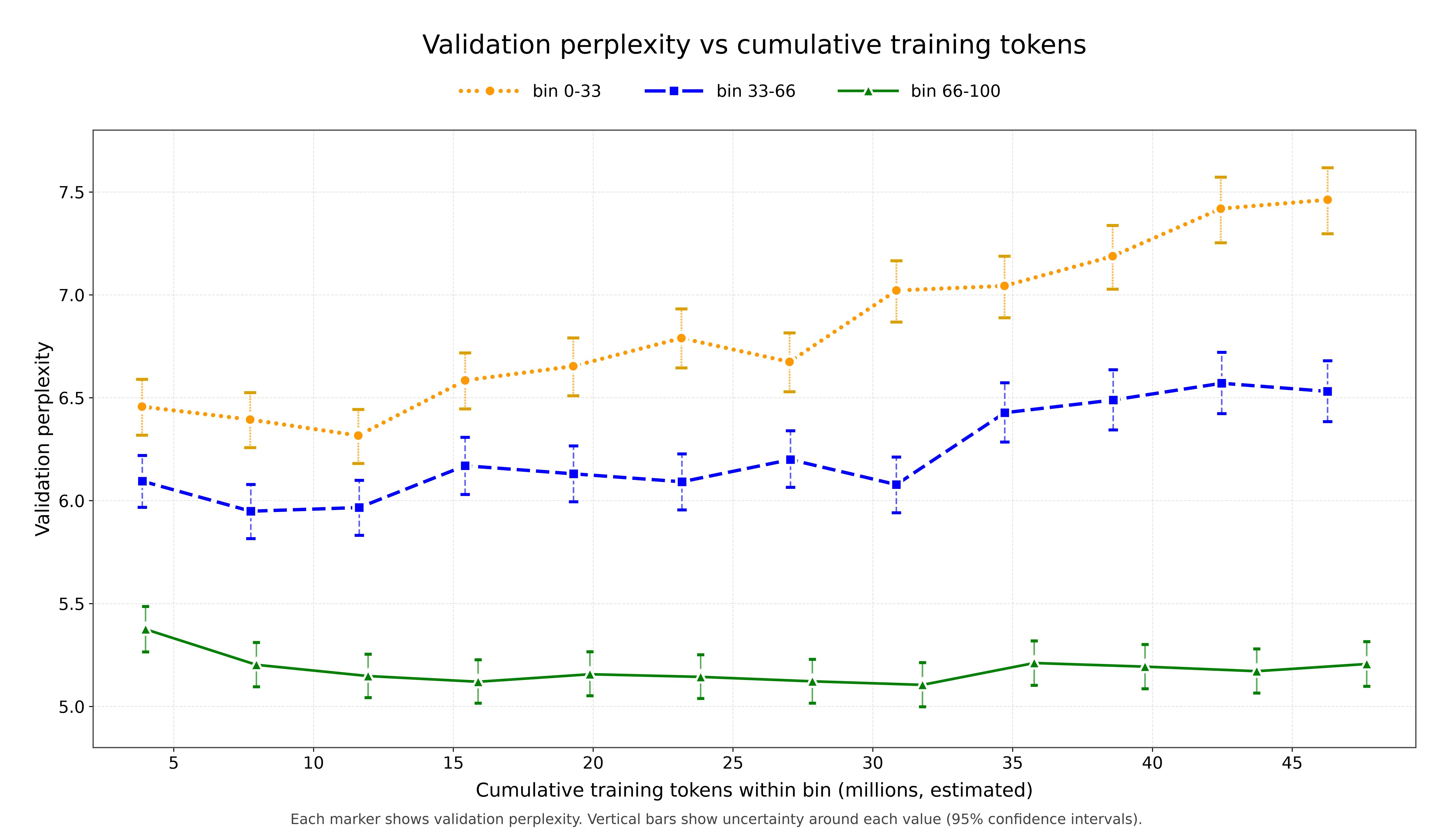}
}%
\subfloat[Llama-3.1-8B, FLAN 10\%:\\Perplexity for semantic density]{
  \includegraphics[width=\columnwidth]{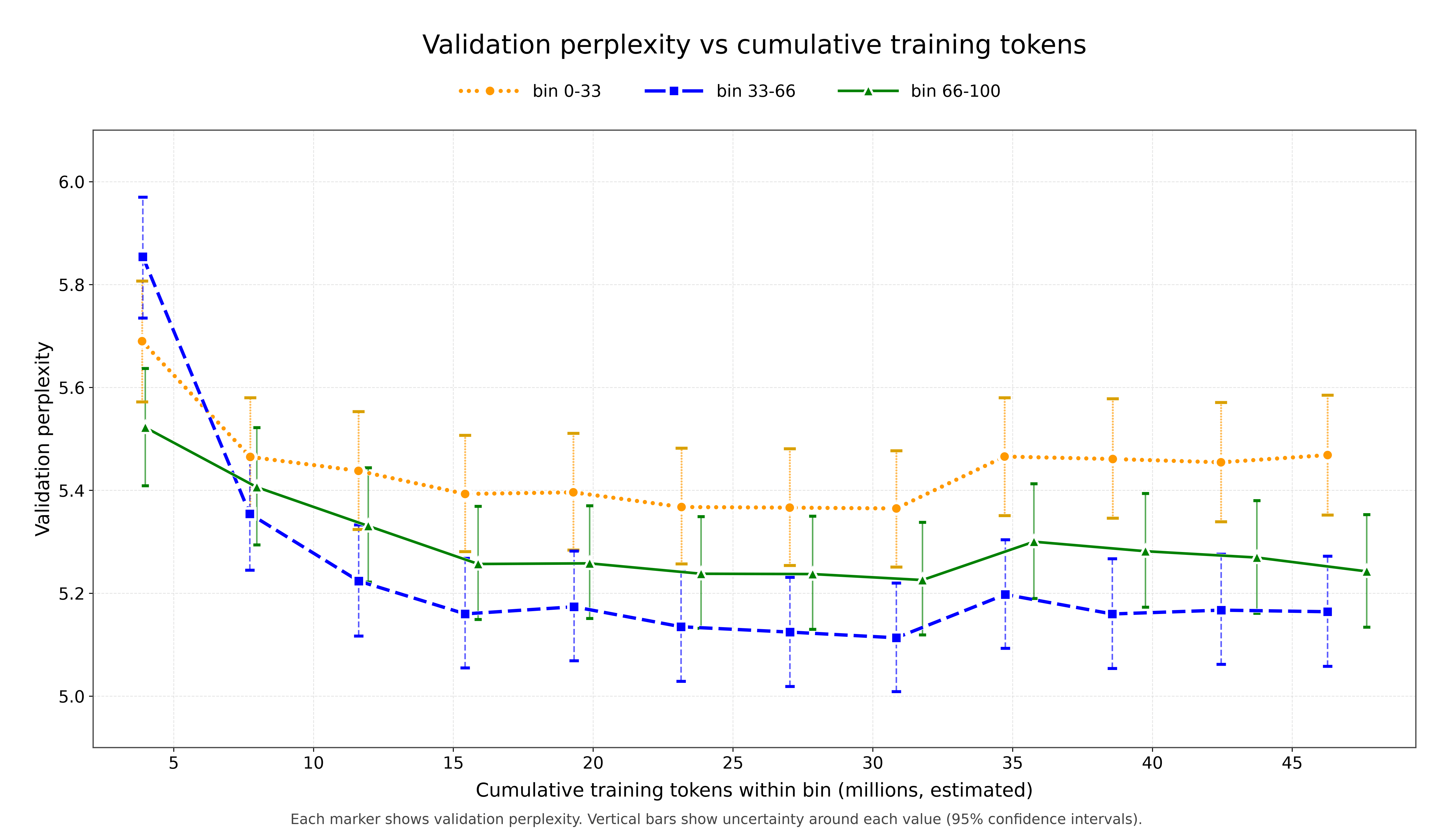}
}%
\caption{Perplexity curves for Llama-3.1-8B on FLAN 10\%.}
\label{fig:app-lama8b-flan10p}
\vspace{1.0cm}
\end{figure*}
\begin{figure*}
\subfloat[Llama-3.1-8B, FLAN 10\%:\\LLM-as-judge scores for length]{
  \includegraphics[width=\columnwidth]{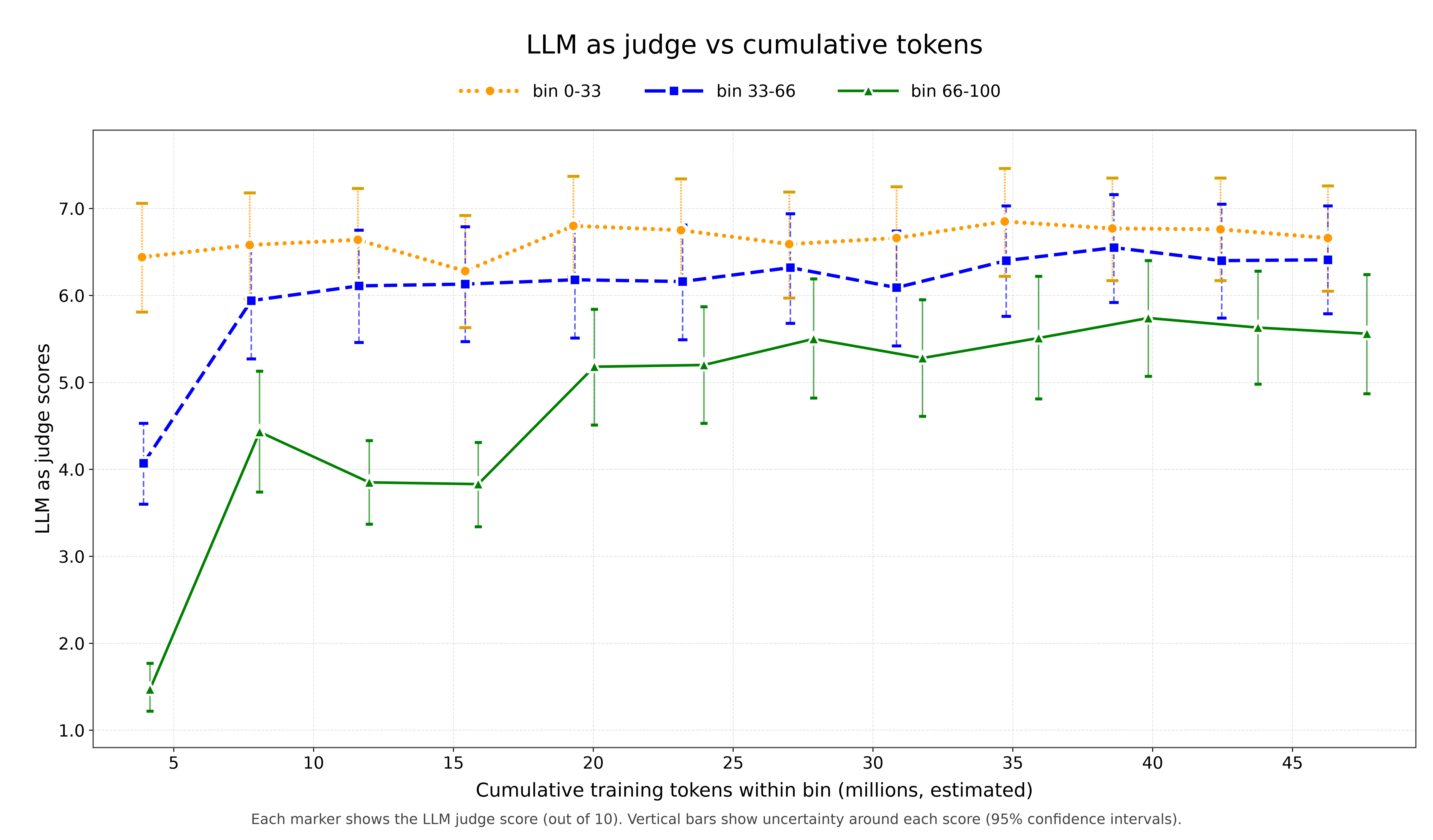}
}%
\subfloat[Llama-3.1-8B, FLAN 10\%:\\LLM-as-judge scores for alignment]{
  \includegraphics[width=\columnwidth]{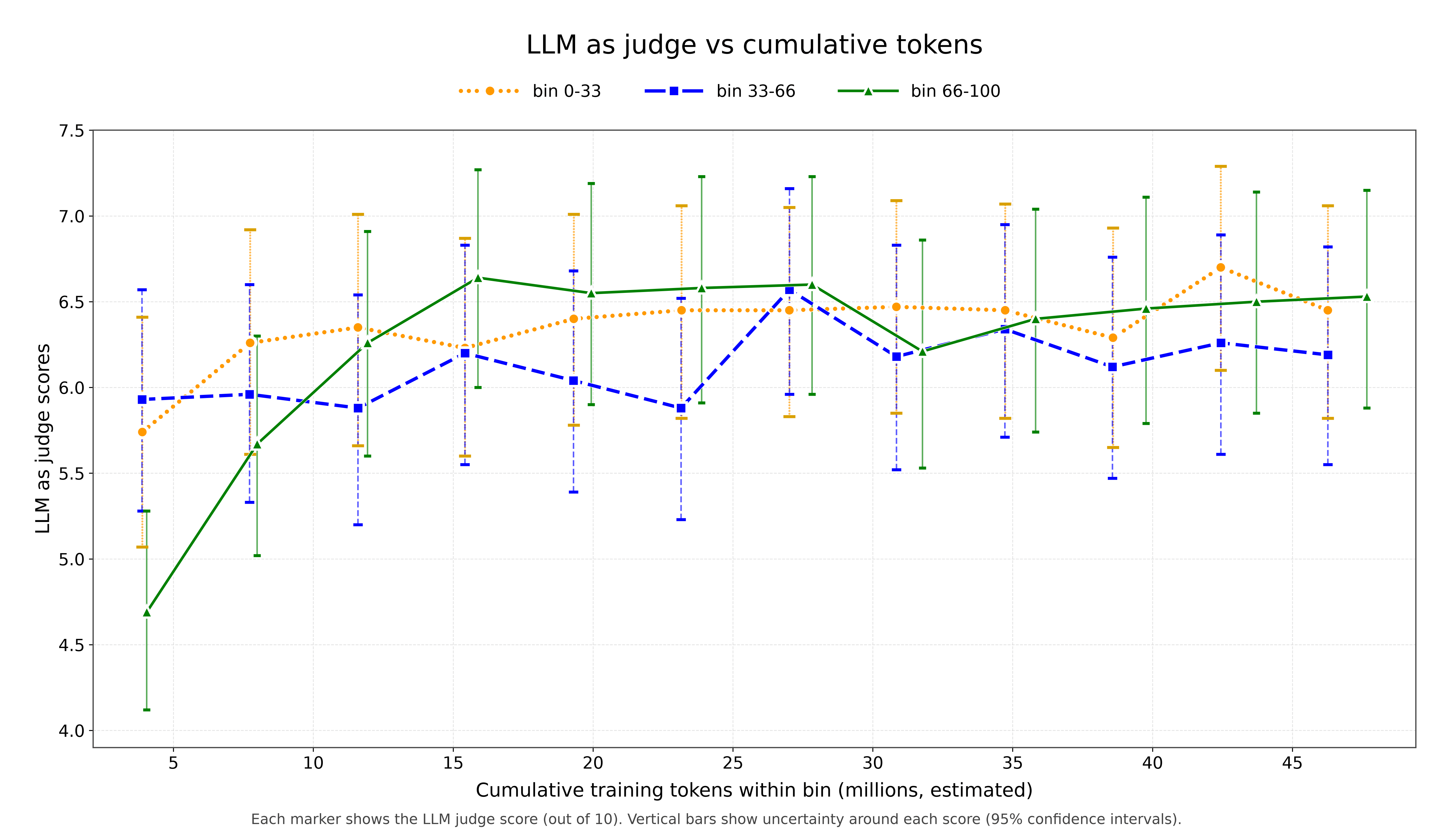}
} \\\\%
\subfloat[Llama-3.1-8B, FLAN 10\%:\\LLM-as-judge scores for response ratio]{
  \includegraphics[width=\columnwidth]{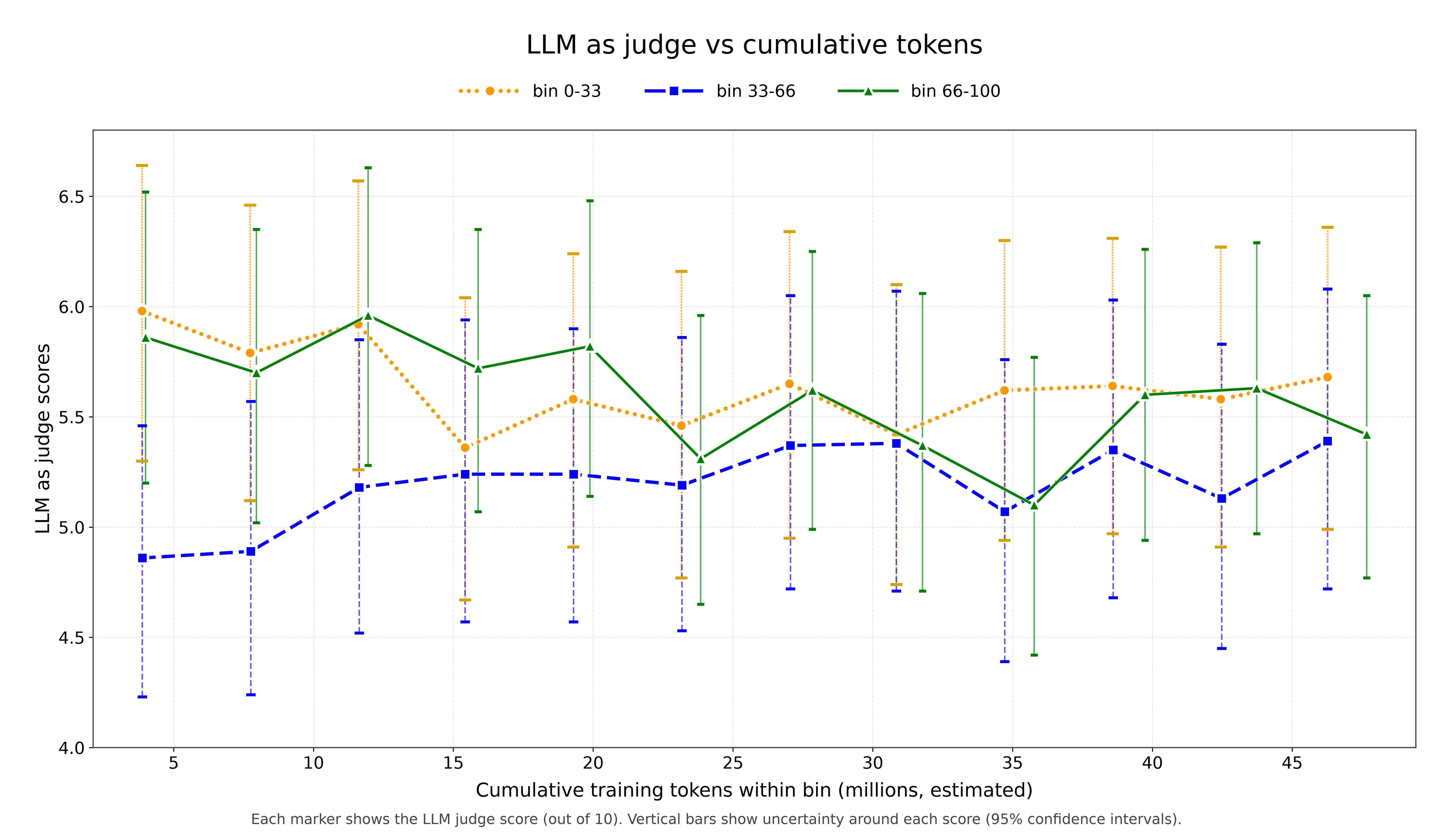}
}%
\subfloat[Llama-3.1-8B, FLAN 10\%:\\LLM-as-judge scores for semantic density]{
  \includegraphics[width=\columnwidth]{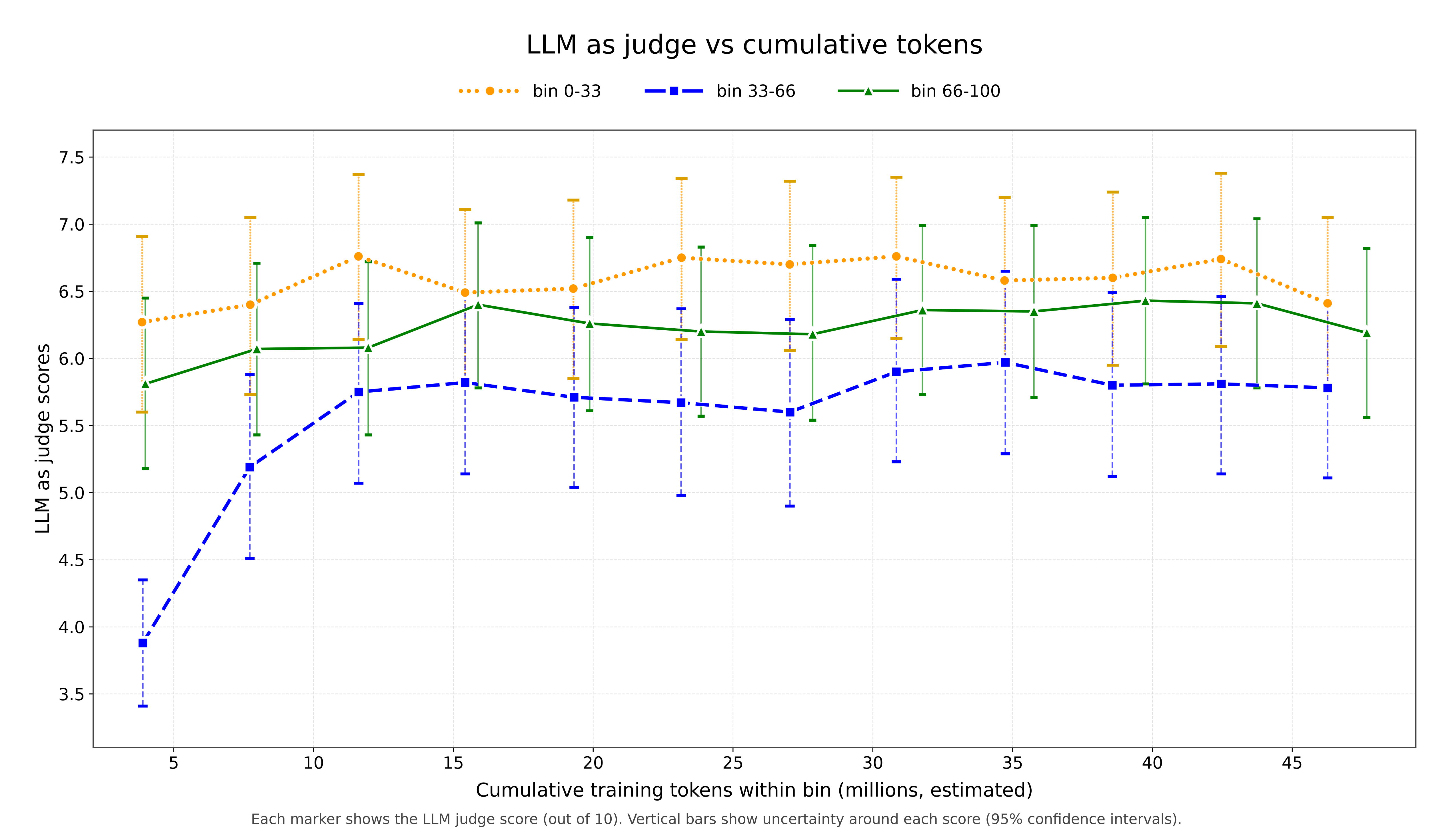}
}%
\caption{LLM-as-judge score curves for Llama-3.1-8B on FLAN 10\%.}
\label{fig:app-judge-lama8b-flan10p}
\end{figure*}
\begin{figure*}
\subfloat[Qwen3.5-9B-Base, FLAN 10\%:\\Perplexity for length]{
  \includegraphics[width=\columnwidth]{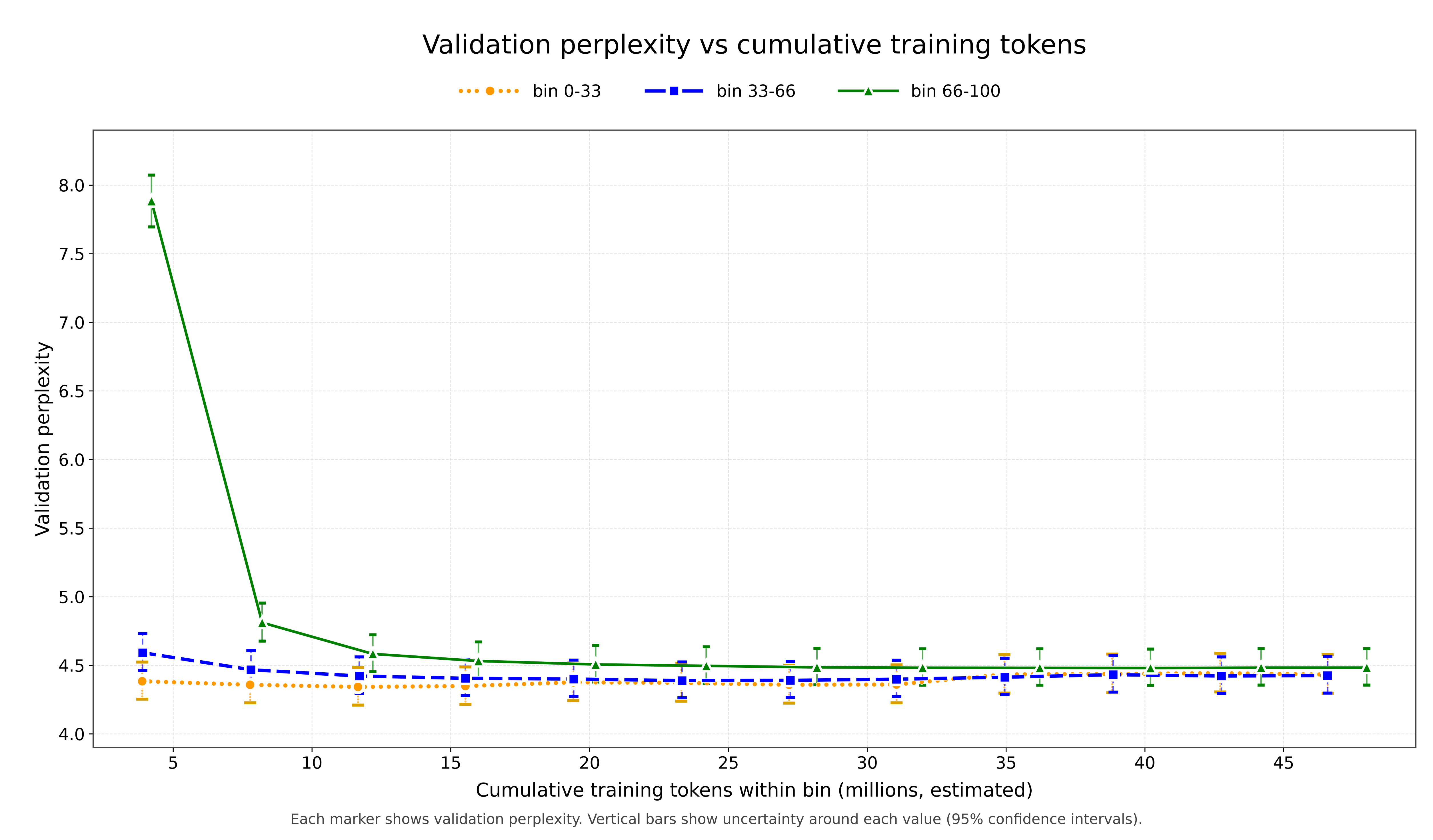}
}%
\subfloat[Qwen3.5-9B-Base, FLAN 10\%:\\Perplexity for alignment]{
  \includegraphics[width=\columnwidth]{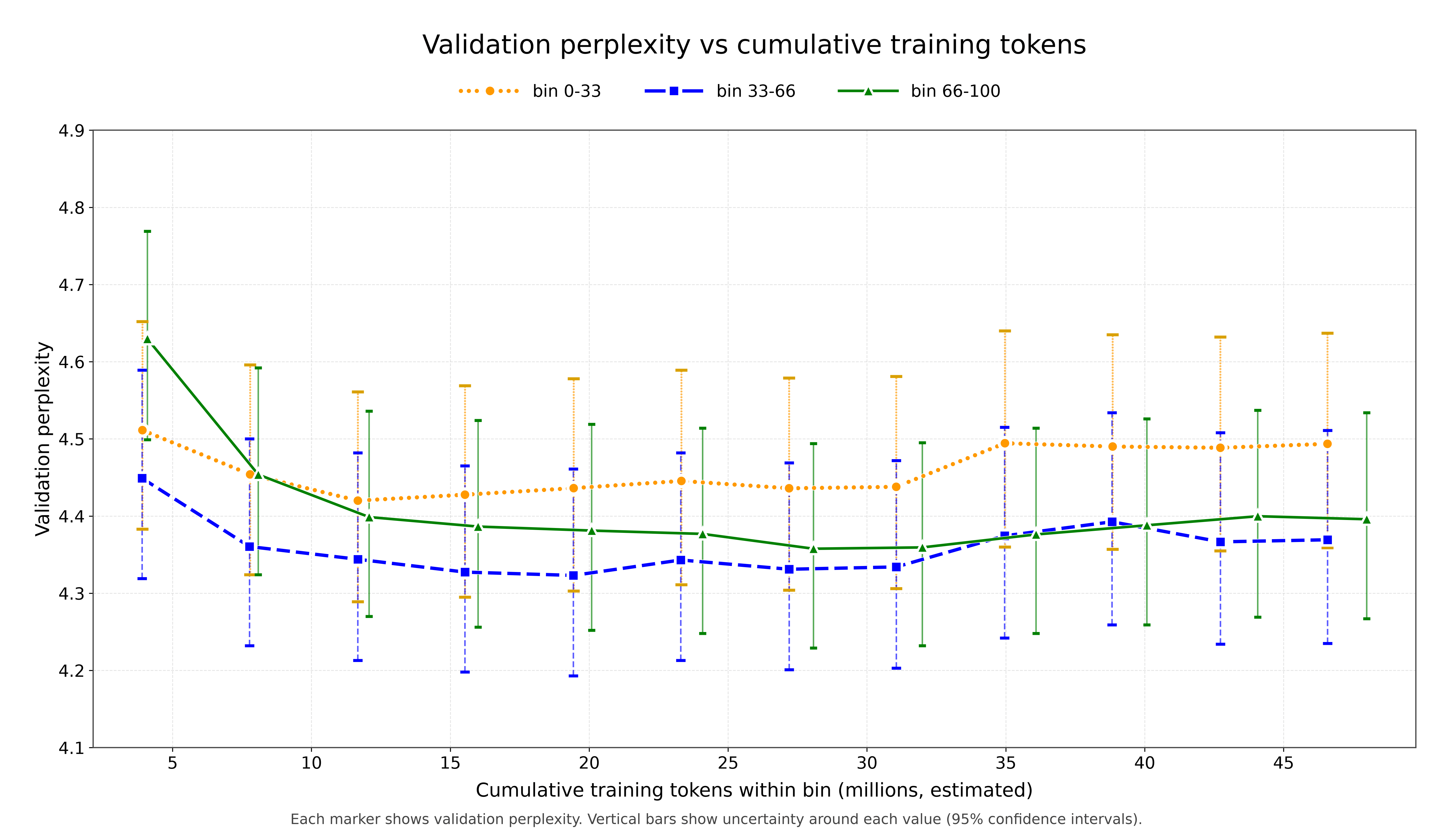}
}\\\\%
\subfloat[Qwen3.5-9B-Base, FLAN 10\%:\\Perplexity for response ratio]{
  \includegraphics[width=\columnwidth]{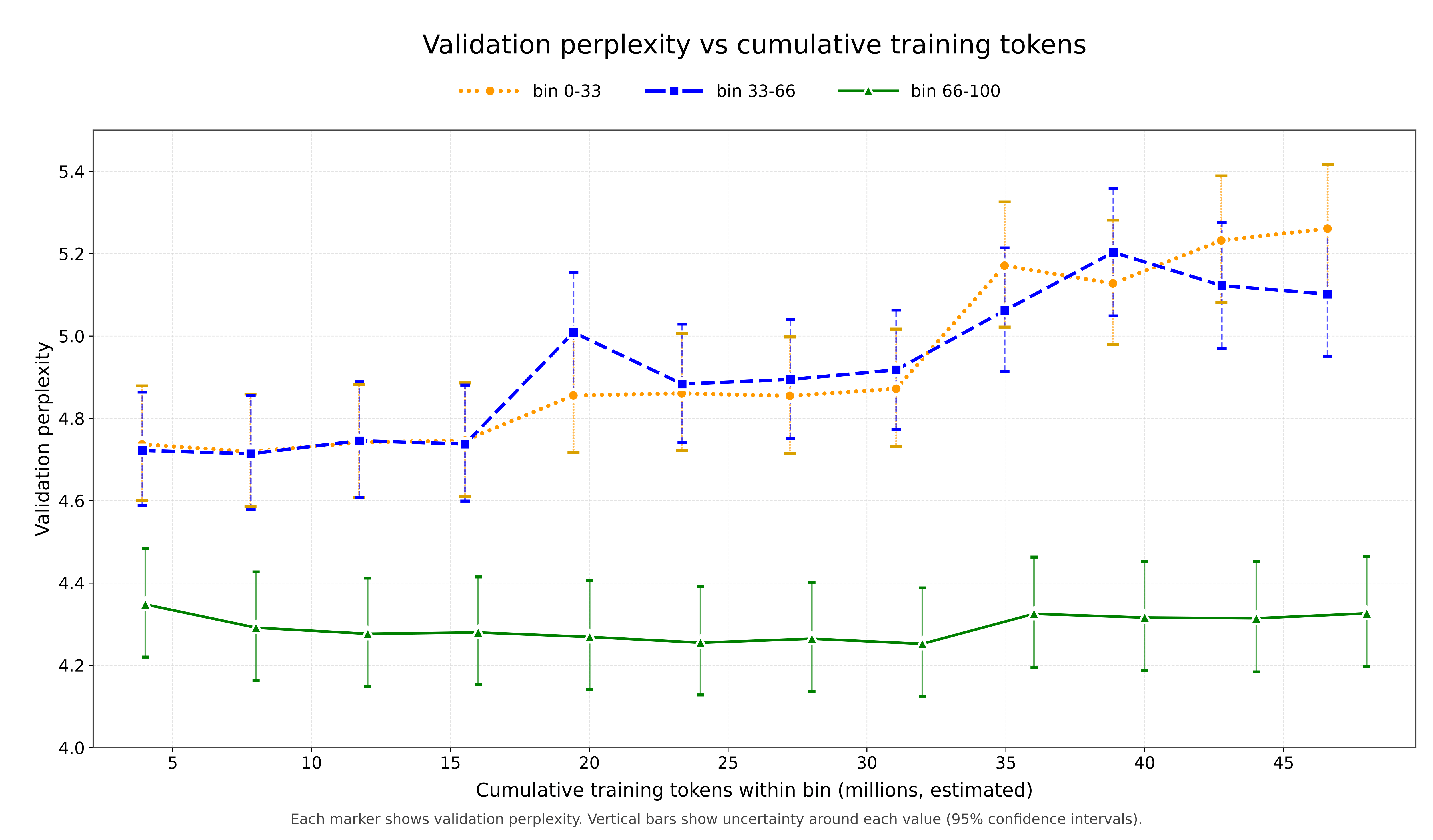}
}%
\subfloat[Qwen3.5-9B-Base, FLAN 10\%:\\Perplexity for semantic density]{
  \includegraphics[width=\columnwidth]{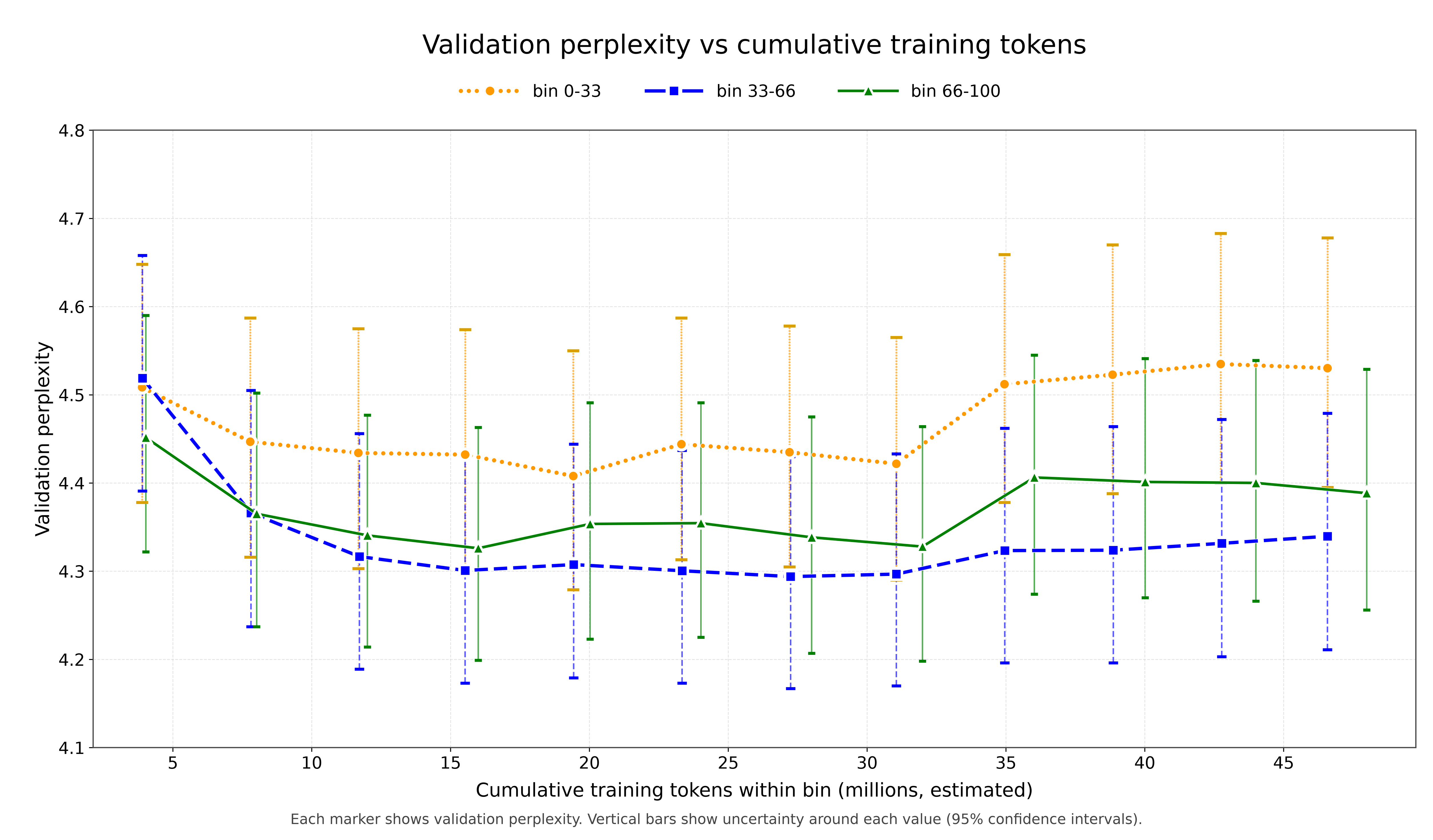}
}%
\caption{Perplexity curves for Qwen3.5-9B-Base on FLAN 10\%.}
\label{fig:app-qwen7b-flan10p}
\vspace{1.0cm}
\end{figure*}
\begin{figure*}
\subfloat[Qwen3.5-9B-Base, FLAN 10\%:\\LLM-as-judge scores for length]{
  \includegraphics[width=\columnwidth]{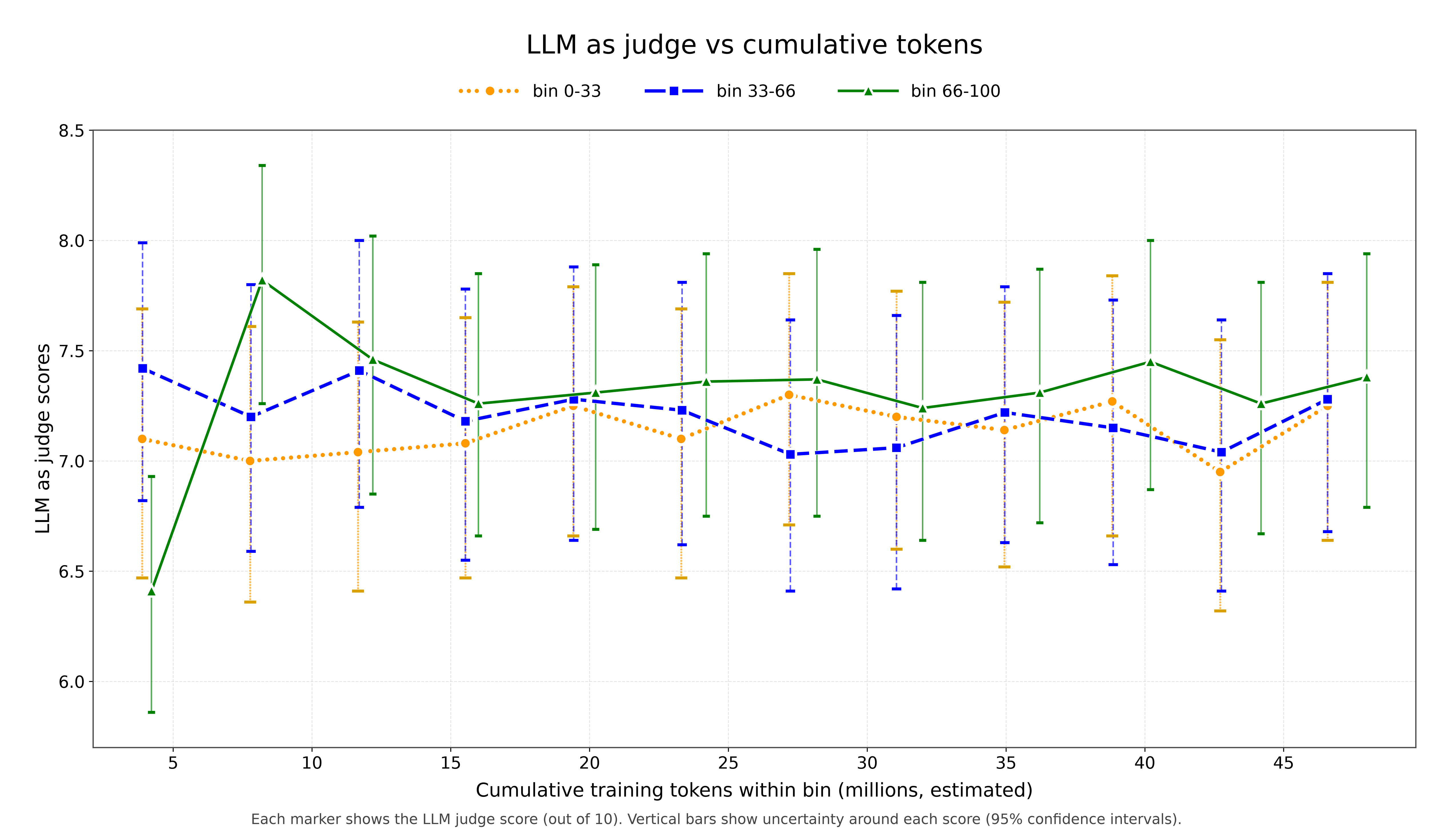}
}%
\subfloat[Qwen3.5-9B-Base, FLAN 10\%:\\LLM-as-judge scores for alignment]{
  \includegraphics[width=\columnwidth]{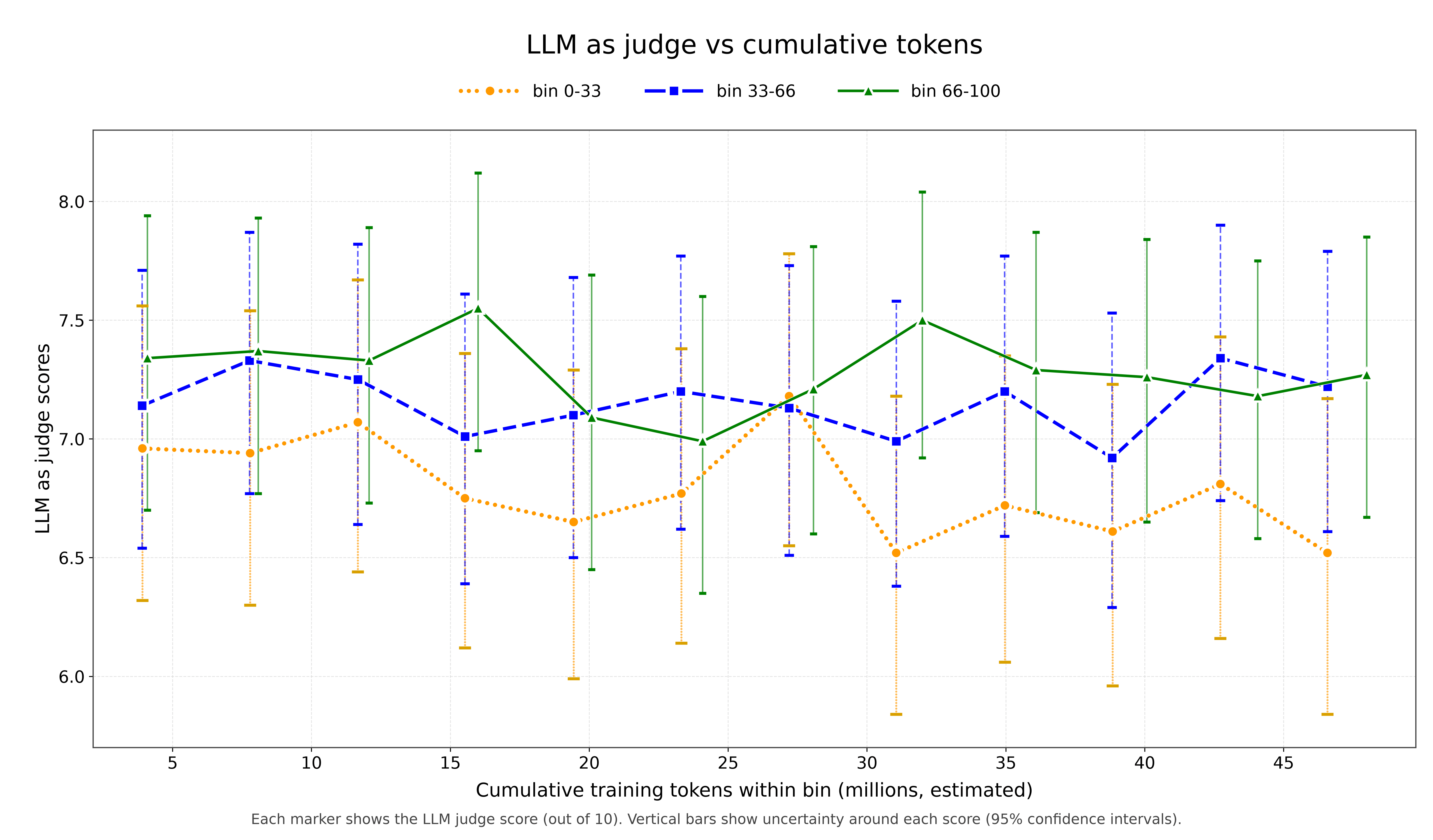}
}\\\\%
\subfloat[Qwen3.5-9B-Base, FLAN 10\%:\\LLM-as-judge scores for response ratio]{
  \includegraphics[width=\columnwidth]{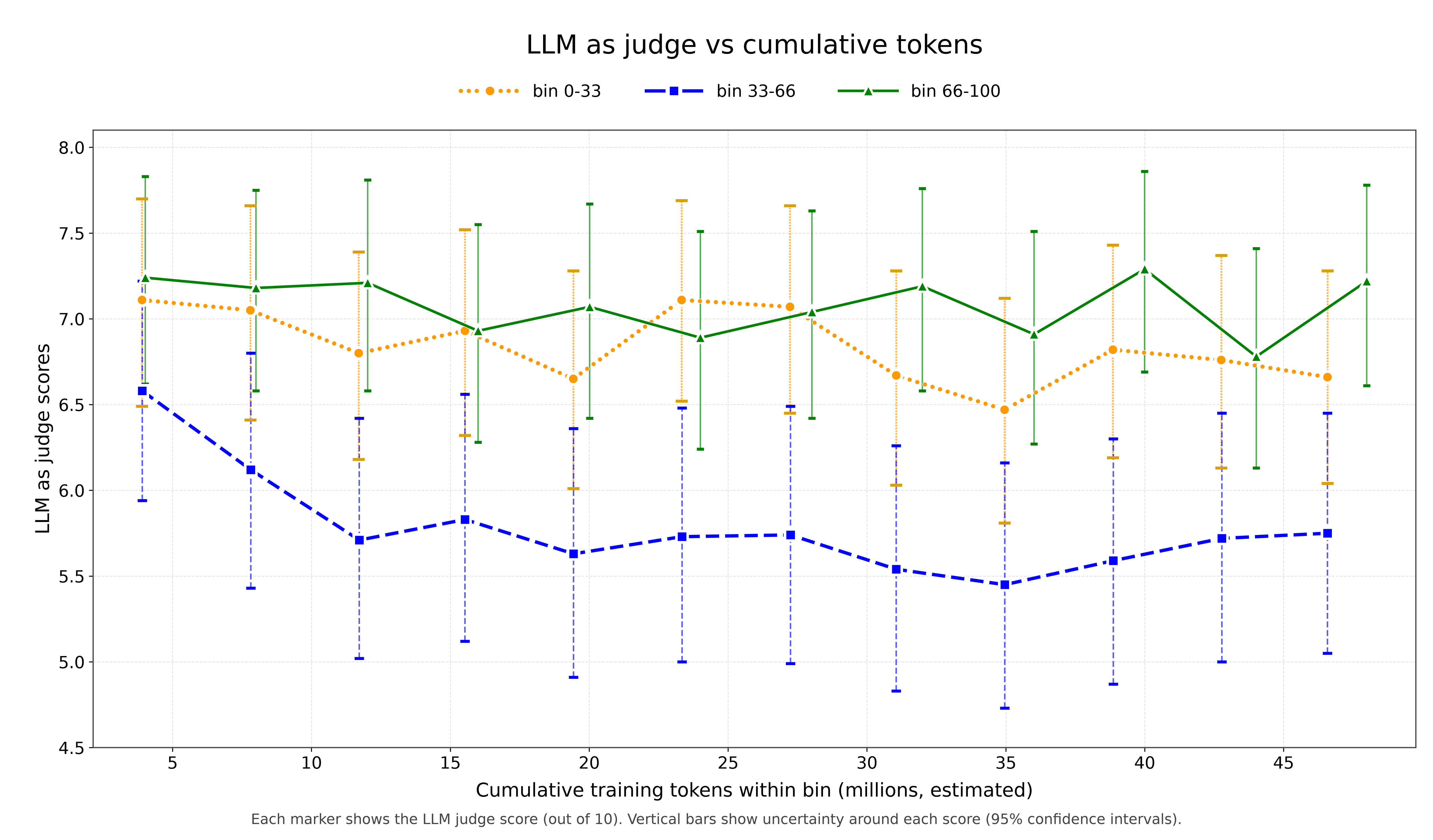}
}%
\subfloat[Qwen3.5-9B-Base, FLAN 10\%:\\LLM-as-judge scores for semantic density]{
  \includegraphics[width=\columnwidth]{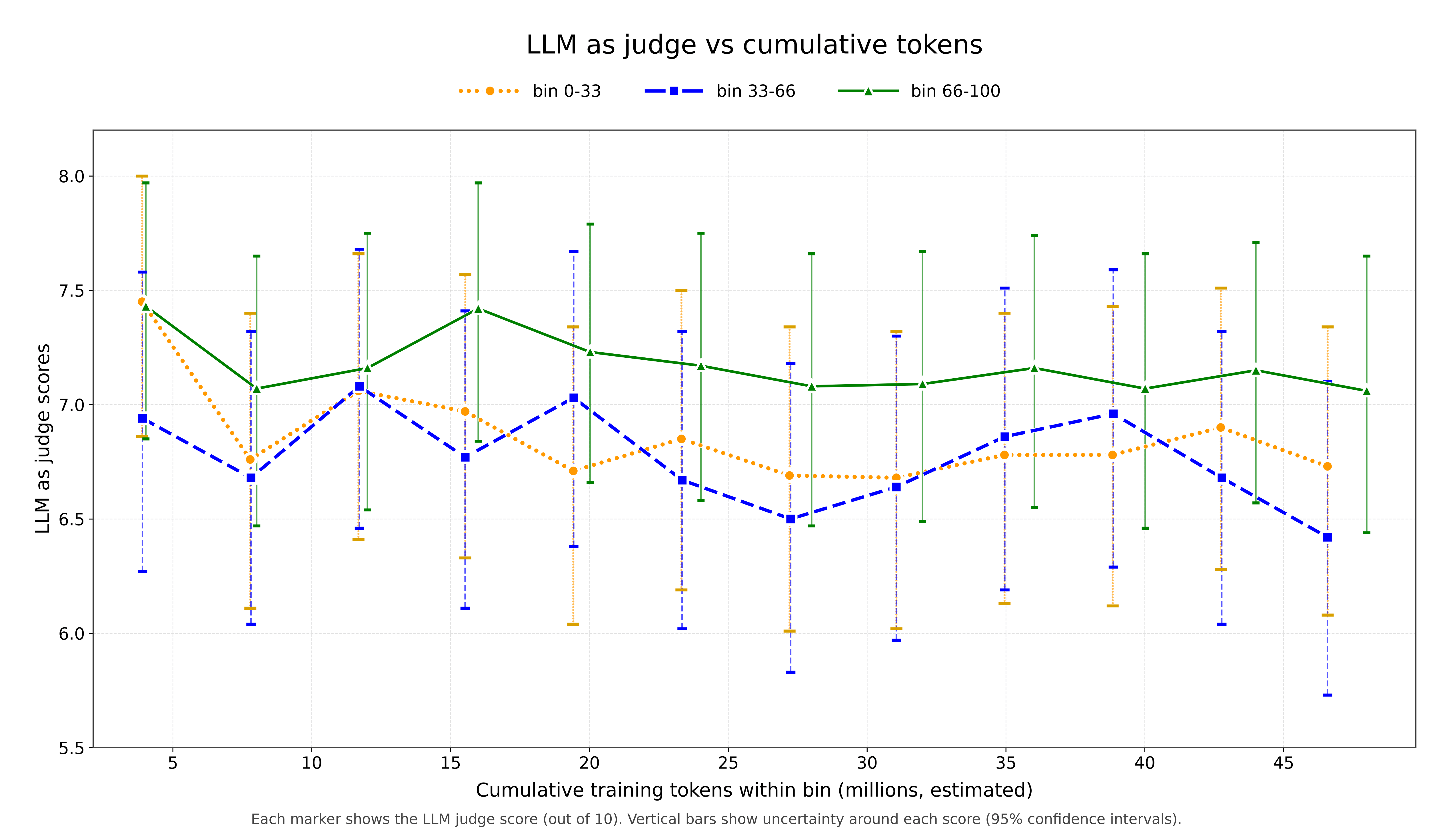}
}%
\caption{LLM-as-judge score curves for Qwen3.5-9B-Base on FLAN 10\%.}
\label{fig:app-judge-qwen7b-flan10p}
\end{figure*}
\begin{figure*}
\subfloat[Llama-3.1-8B, FLAN 10\%:\\Perplexity for alignment without high response ratio]{
  \includegraphics[width=\columnwidth]{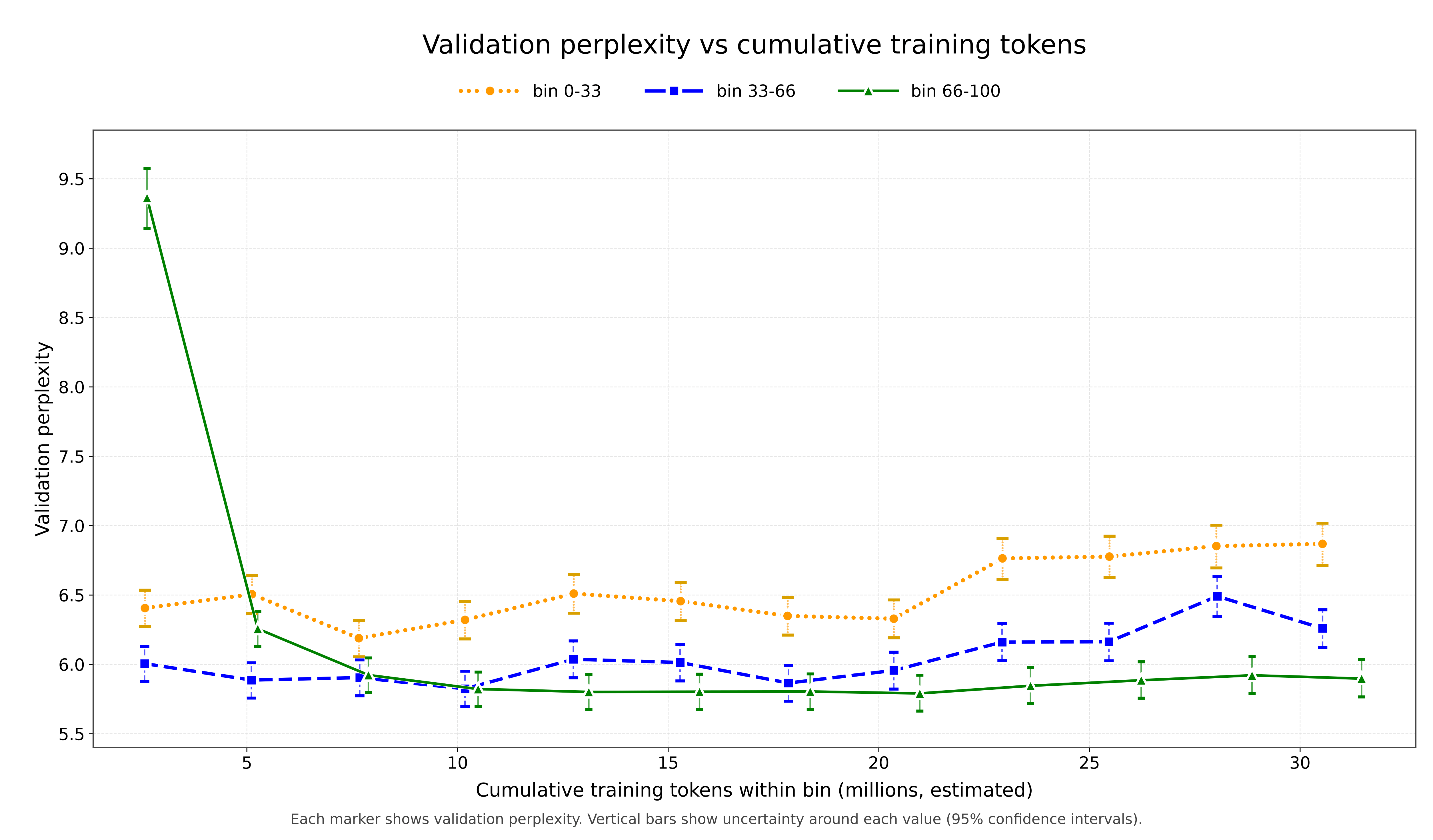}
}%
\subfloat[Llama-3.1-8B, FLAN 10\%:\\Perplexity for response ratio without high alignment]{
  \includegraphics[width=\columnwidth]{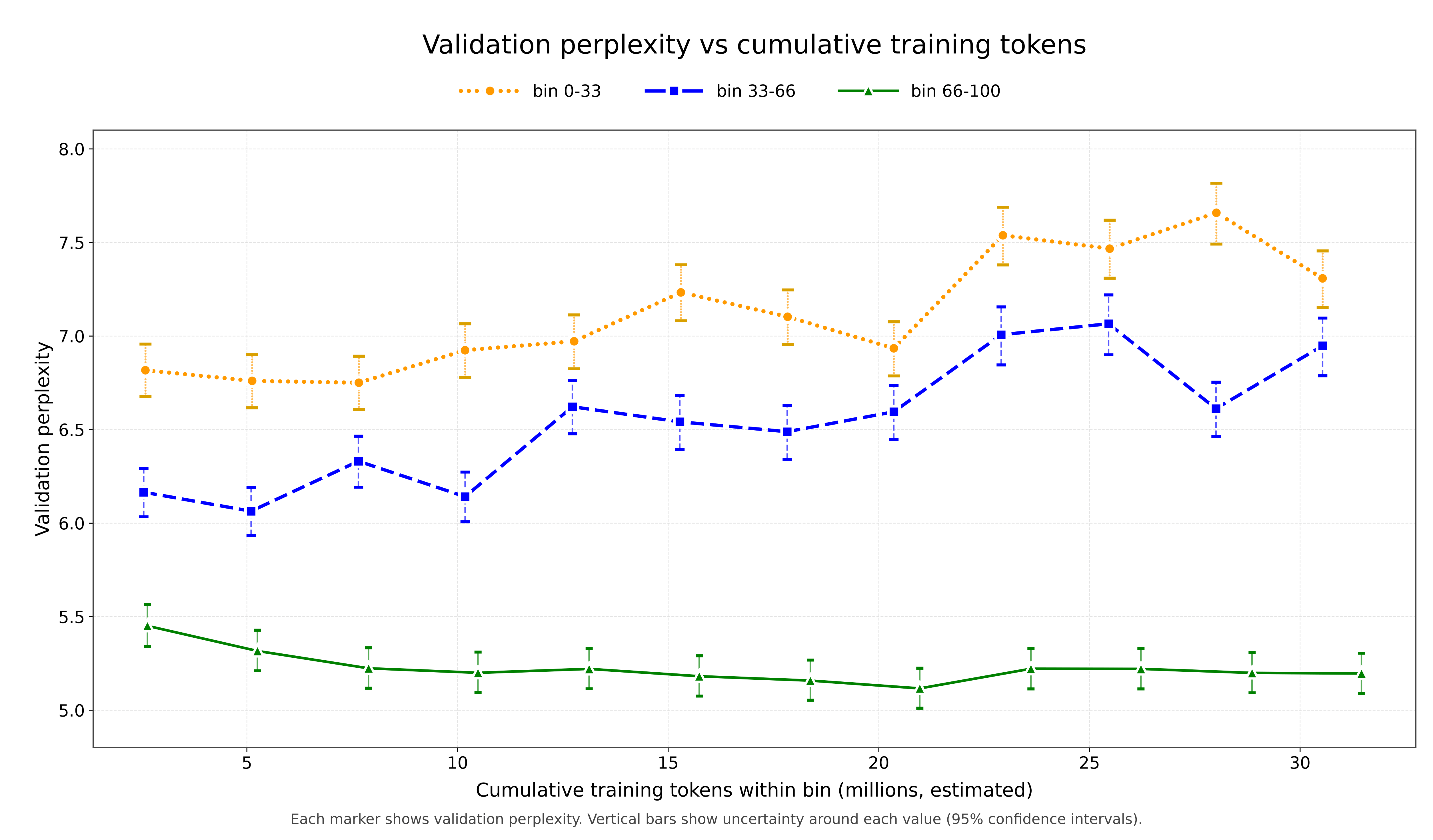}
}%
\caption{Perplexity curves for cross-ablation with Llama-3.1-8B on FLAN 10\%, comparing alignment after removing high-response-ratio examples with response ratio after removing high-alignment examples.}
\vspace{4cm}
\label{figs:standard-cross-ablation-llamaFLAN}
\end{figure*}
\begin{figure*}
\subfloat[Llama-3.1-8B, FLAN 10\%:\\LLM-as-judge scores for\\alignment without high response ratio]{
  \includegraphics[width=\columnwidth]{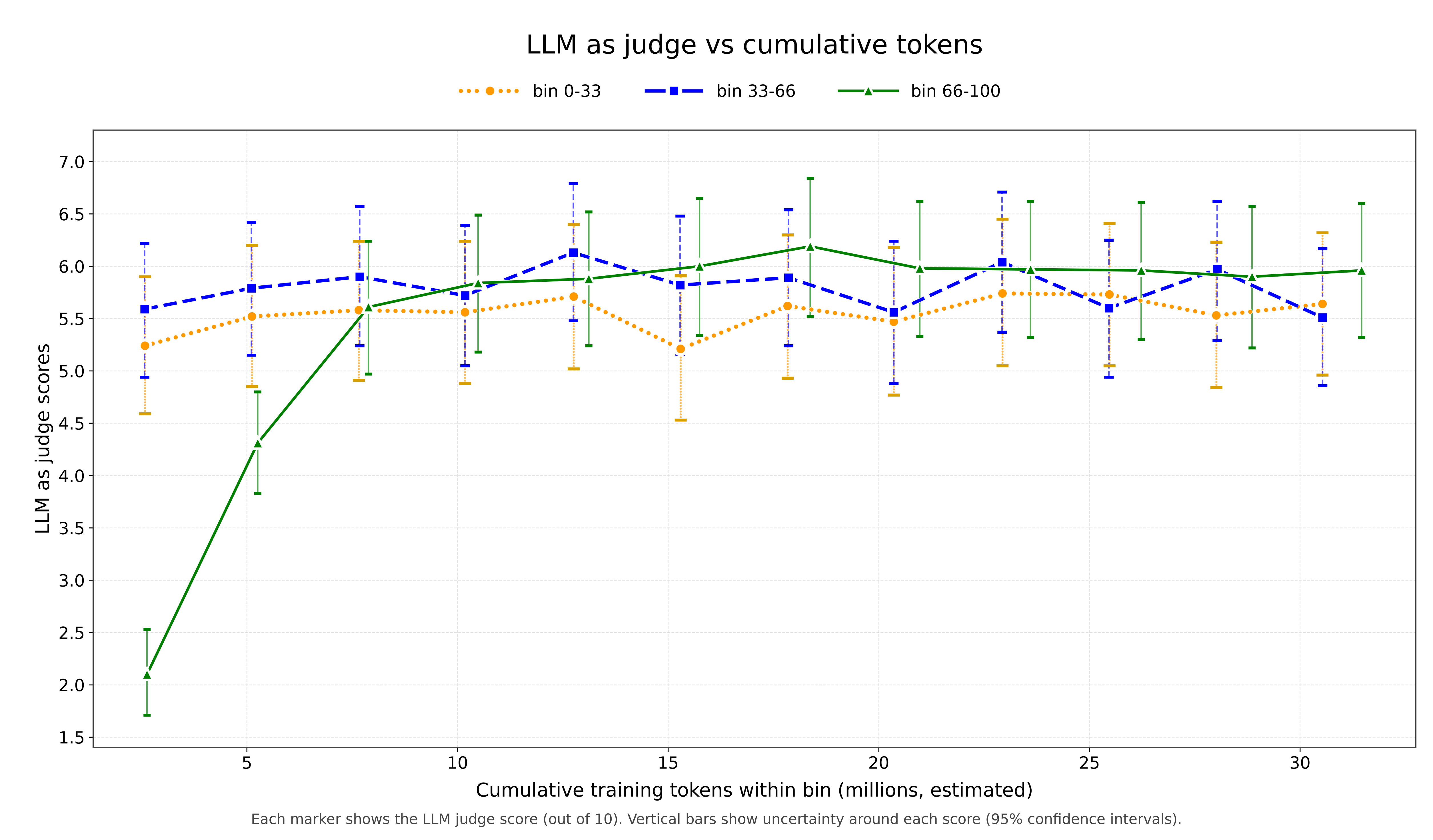}
}%
\subfloat[Llama-3.1-8B, FLAN 10\%:\\LLM-as-judge scores for\\response ratio without high alignment]{
  \includegraphics[width=\columnwidth]{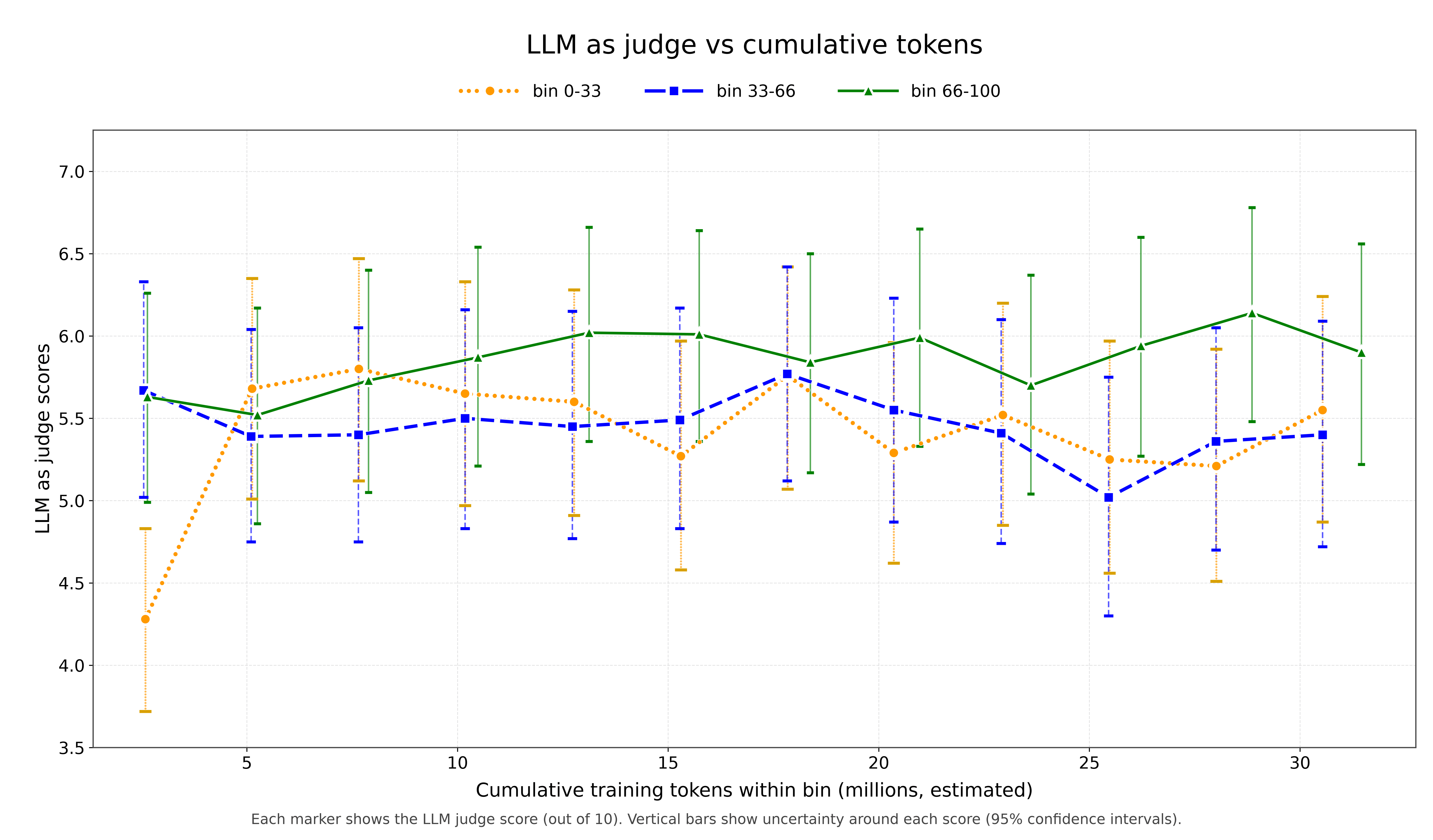}
}%
\caption{LLM-as-judge curves for cross-ablation with Llama-3.1-8B on FLAN 10\%, comparing alignment after removing high-response-ratio examples with response ratio after removing high-alignment examples.}
\vspace{4cm}
\label{figs:standard-cross-ablation-judge-llamaFLAN}
\end{figure*}
\begin{figure*}
\subfloat[Qwen3.5-9B-Base, FLAN 10\%:\\Perplexity for alignment without high response ratio]{
  \includegraphics[width=\columnwidth]{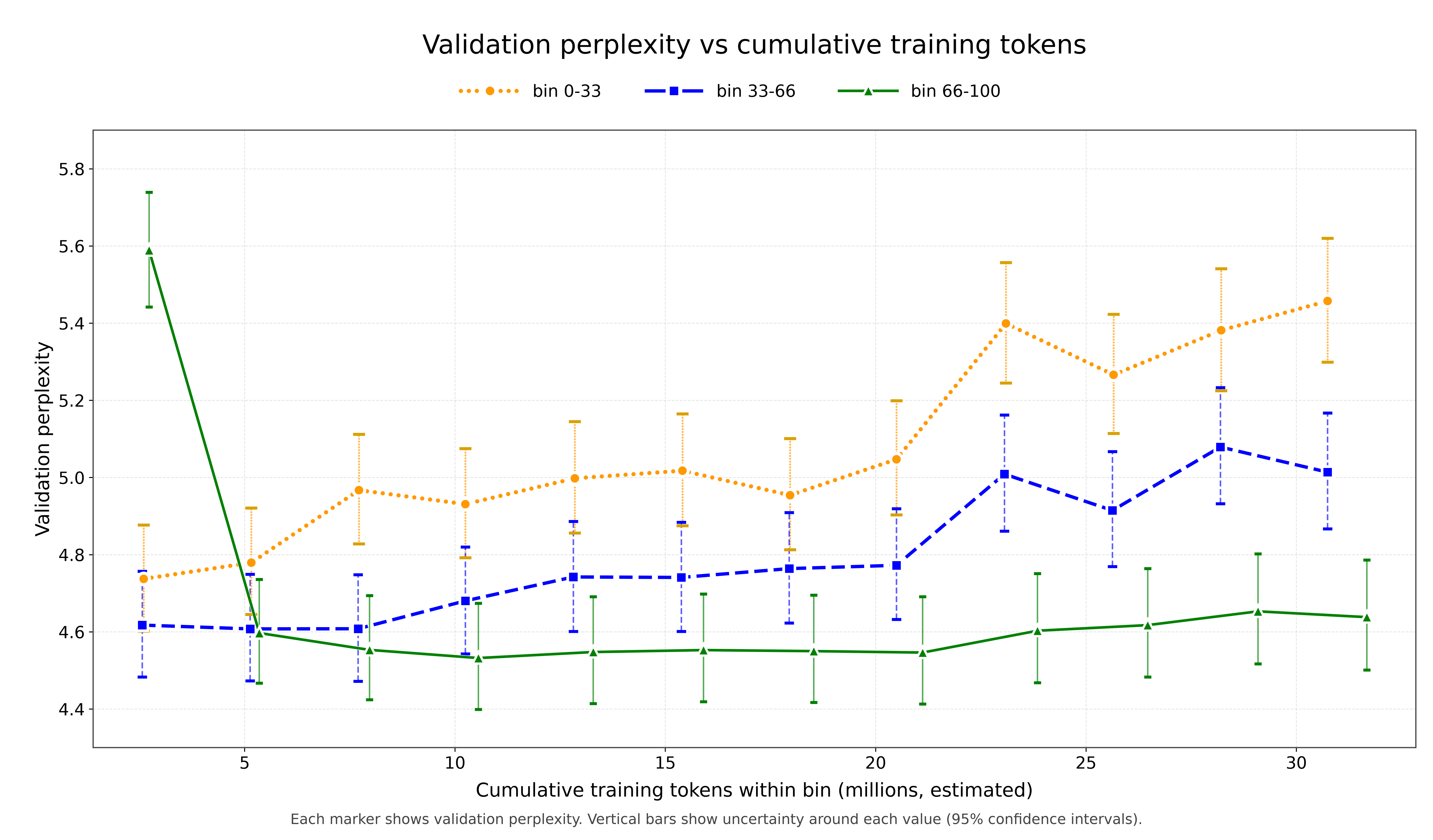}
}%
\subfloat[Qwen3.5-9B-Base, FLAN 10\%:\\Perplexity for response ratio without high alignment]{
  \includegraphics[width=\columnwidth]{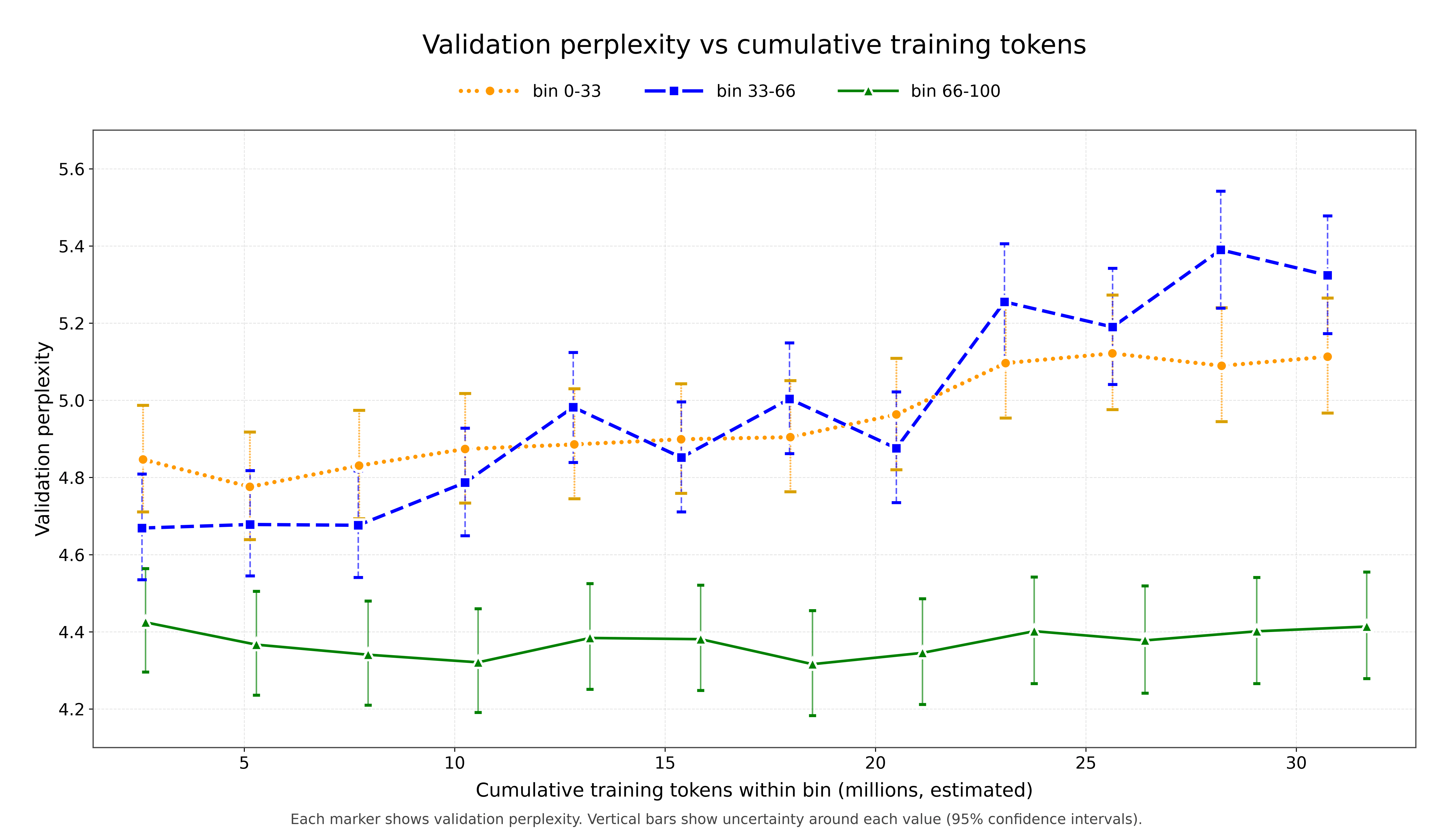}
}%
\caption{Perplexity curves for cross-ablation with Qwen-3.5-9B-Base on FLAN 10\%, comparing alignment after removing high-response-ratio examples with response ratio after removing high-alignment examples.}
\vspace{4cm}
\label{figs:standard-cross-ablation-qwenFLAN}
\end{figure*}
\begin{figure*}
\subfloat[Qwen3.5-9B-Base, FLAN 10\%:\\LLM-as-judge scores for\\alignment without high response ratio]{
  \includegraphics[width=\columnwidth]{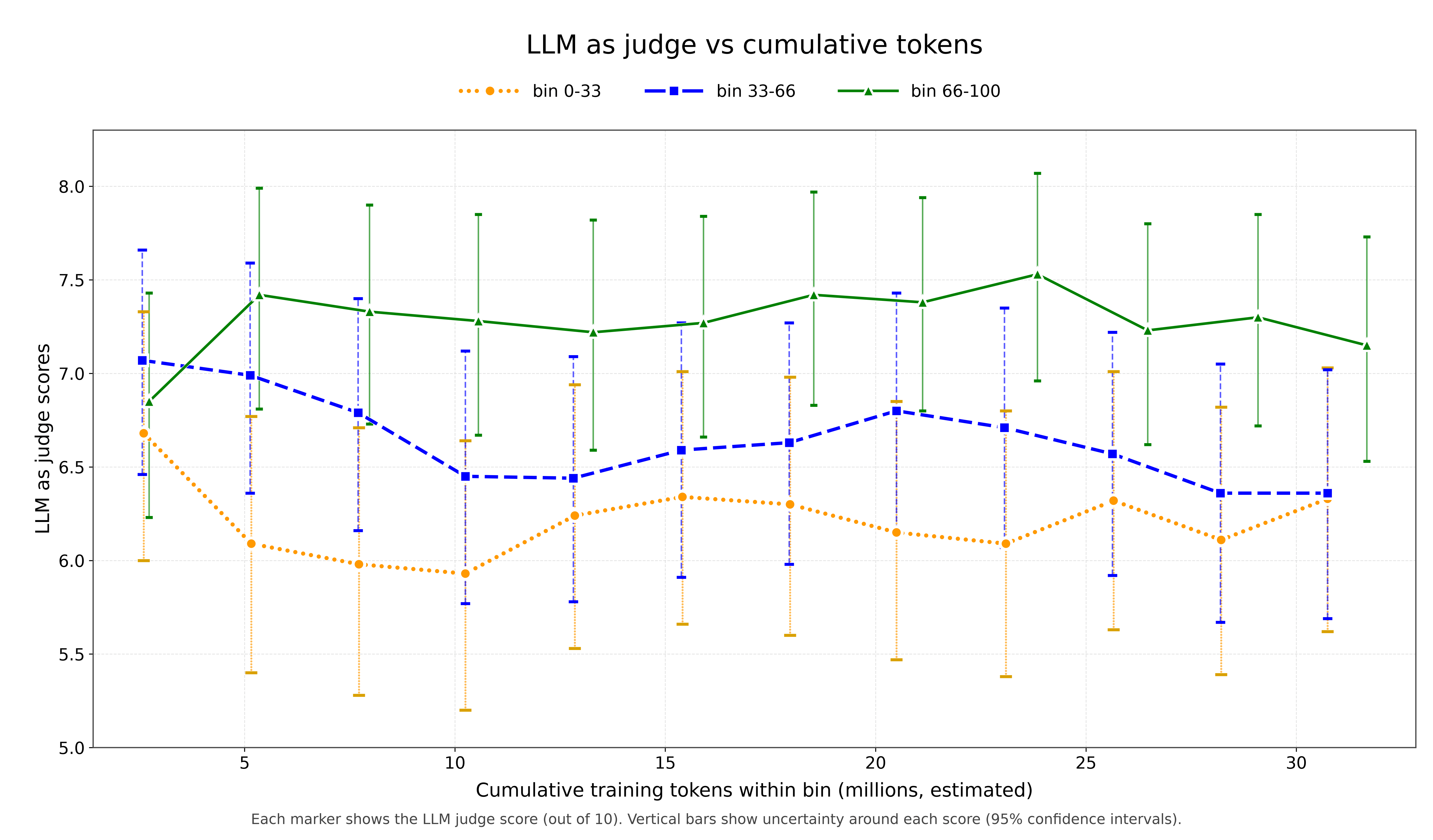}
}%
\subfloat[Qwen3.5-9B-Base, FLAN 10\%:\\LLM-as-judge scores for\\response ratio without high alignment]{
  \includegraphics[width=\columnwidth]{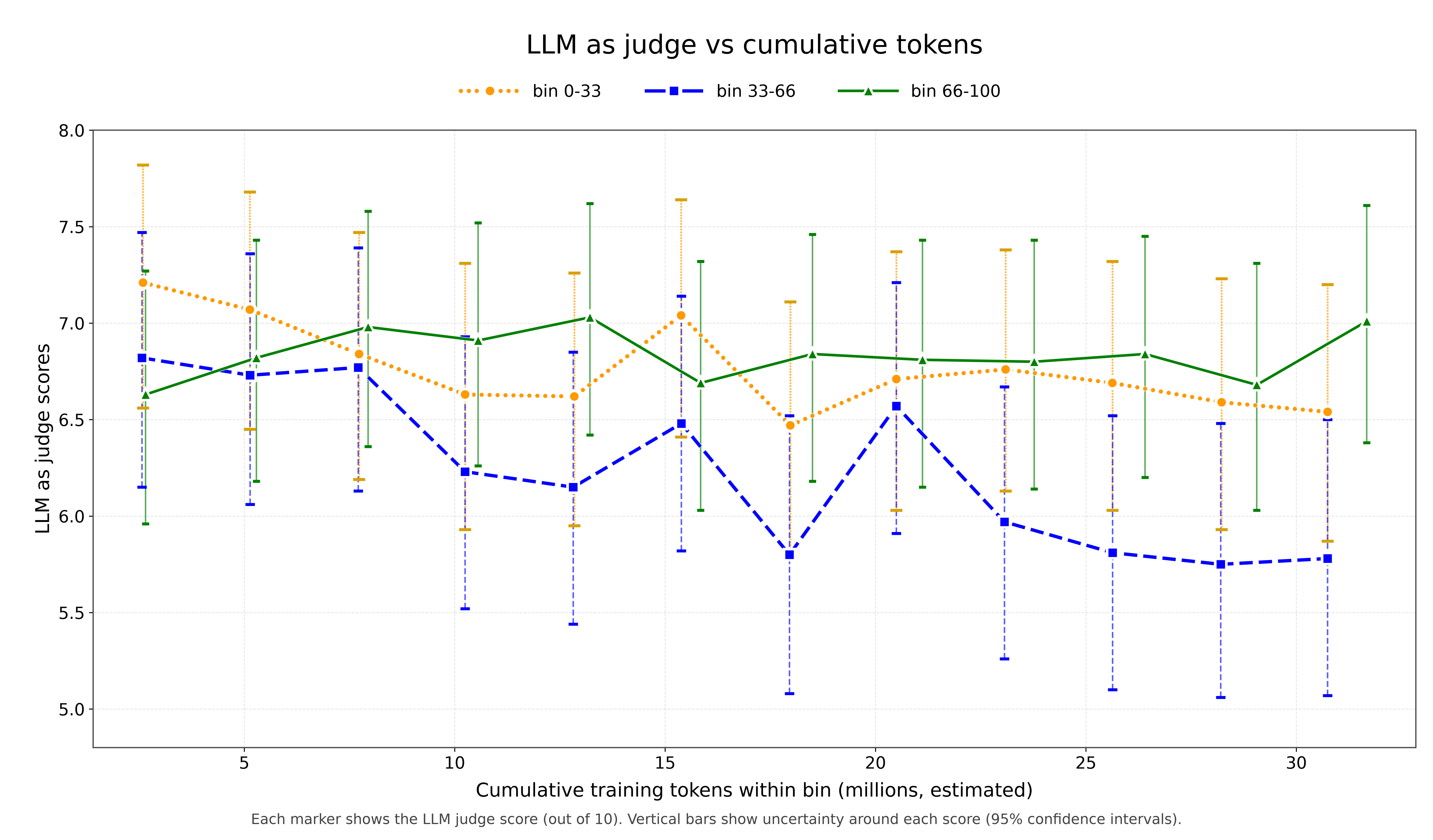}
}%
\caption{LLM-as-judge score curves for cross-ablation with Qwen-3.5-9B-Base on FLAN 10\%, comparing alignment after removing high-response-ratio examples with response ratio after removing high-alignment examples.}
\vspace{4cm}
\label{figs:standard-cross-ablation-judge-qwenFLAN}
\end{figure*}
\begin{figure*}
\subfloat[Llama-3.1-8B, Natural Instructions 10\%:\\Perplexity for alignment without high response ratio]{
  \includegraphics[width=\columnwidth]{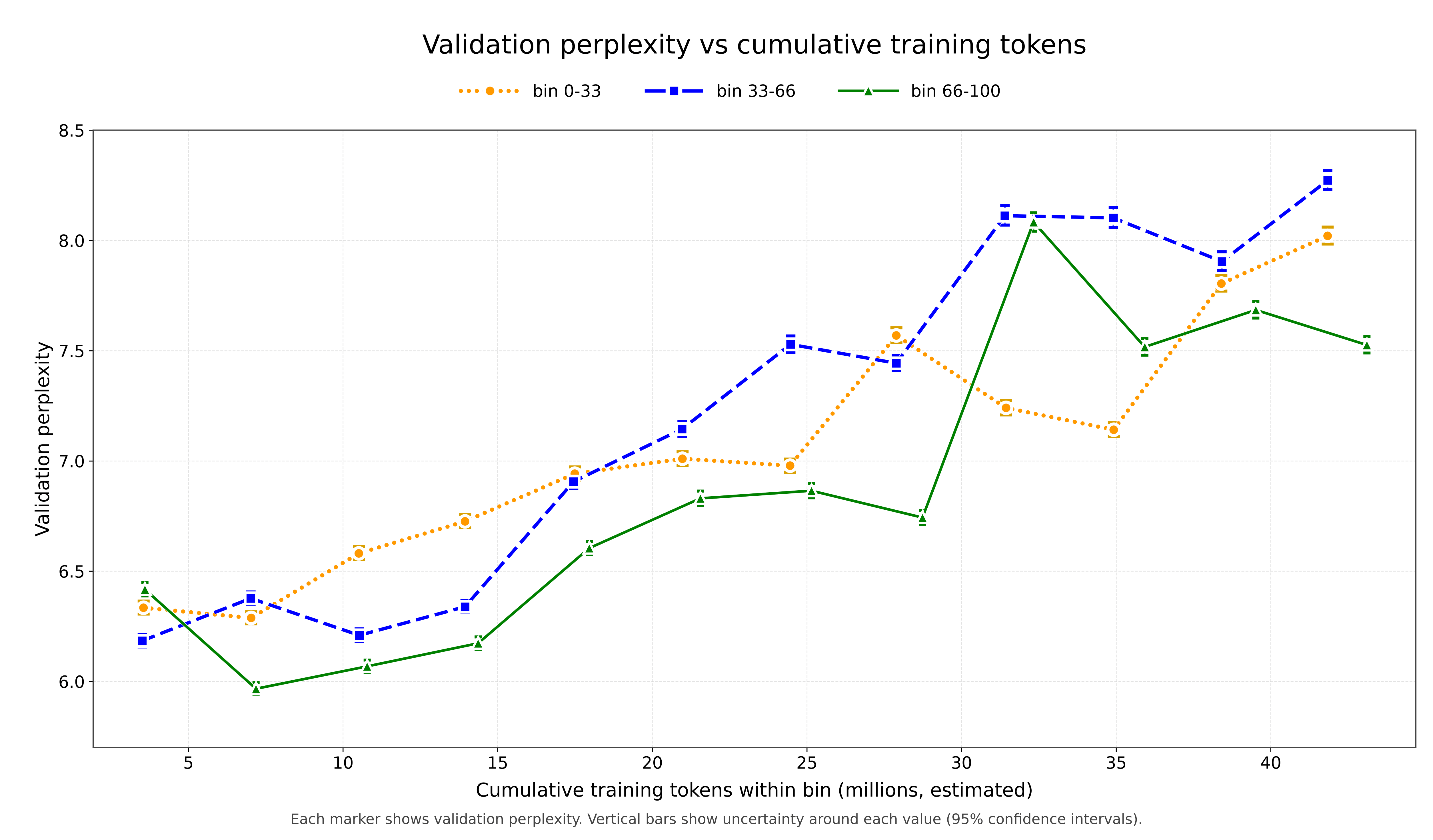}
}%
\subfloat[Llama-3.1-8B, Natural Instructions 10\%:\\Perplexity for response ratio without high alignment]{
  \includegraphics[width=\columnwidth]{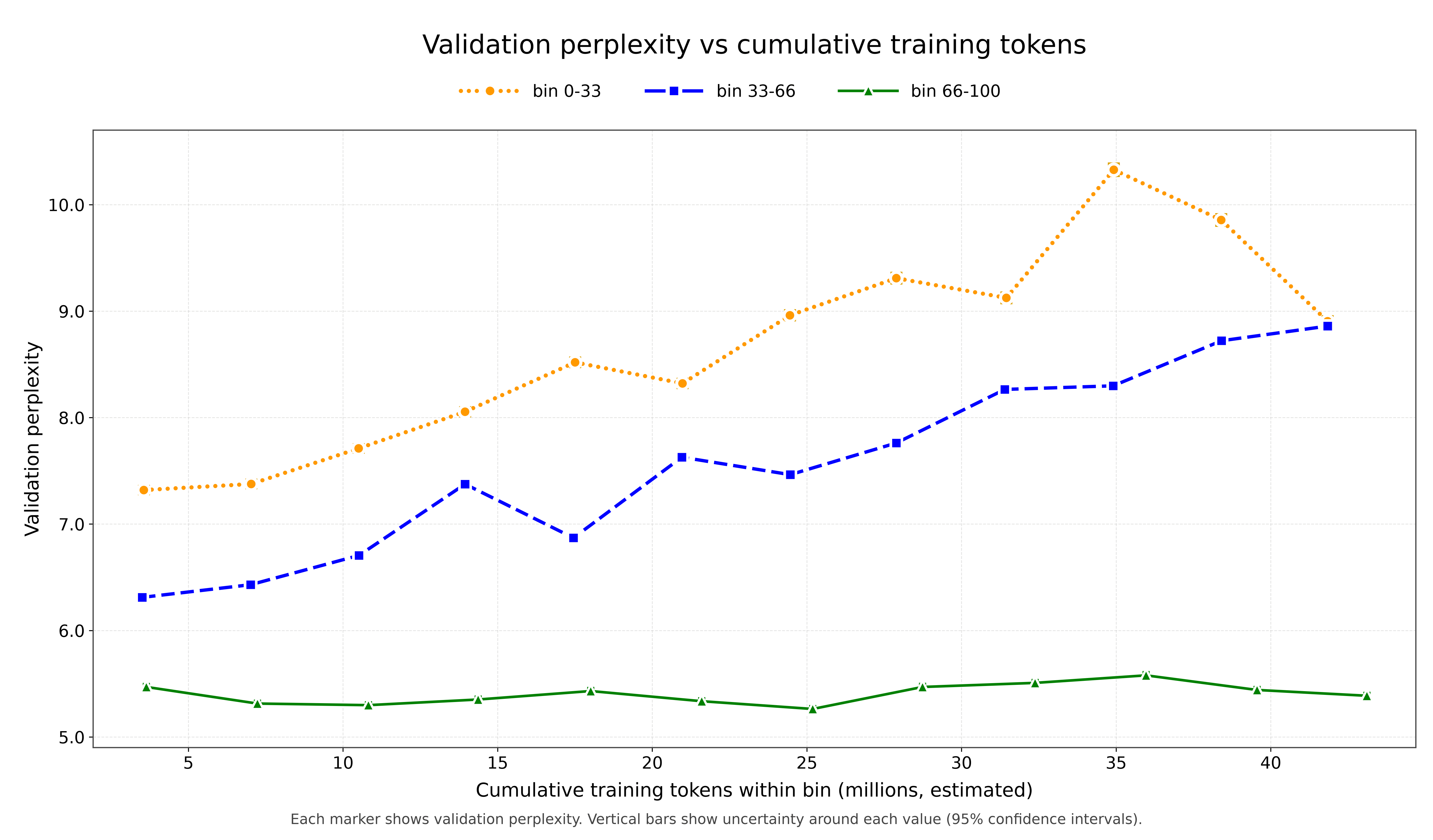}
}%
\caption{Perplexity curves for cross-ablation with Llama-3.1-8B on Natural Instructions 10\%, comparing alignment after removing high-response-ratio examples with response ratio after removing high-alignment examples.}
\vspace{4cm}
\label{figs:standard-cross-ablation-llamaNI}
\end{figure*}
\begin{figure*}
\subfloat[Llama-3.1-8B, Natural Instructions 10\%:\\LLM-as-judge scores for\\alignment without high response ratio]{
  \includegraphics[width=\columnwidth]{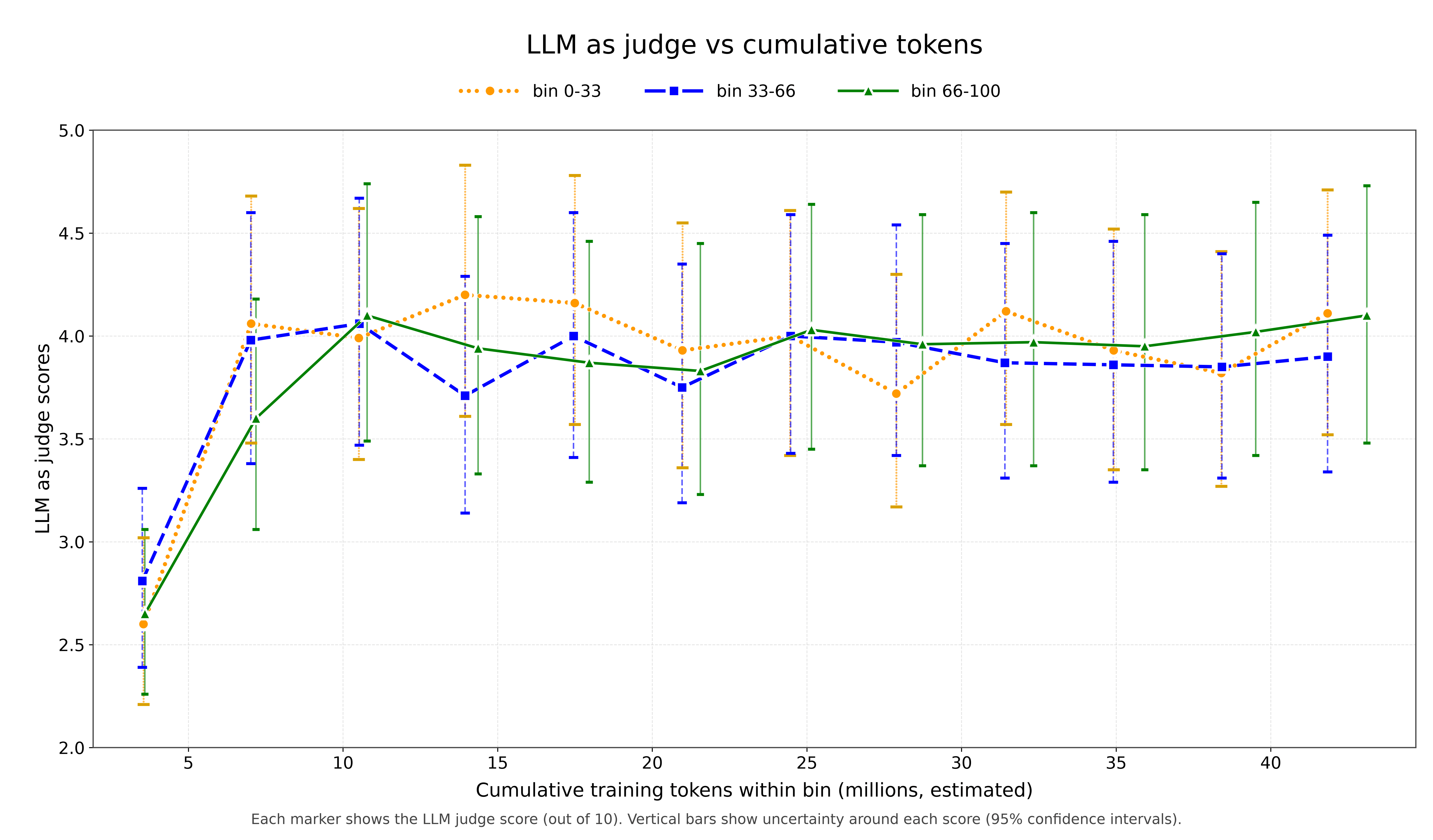}
}%
\subfloat[Llama-3.1-8B, Natural Instructions 10\%:\\LLM-as-judge scores for\\response ratio without high alignment]{
  \includegraphics[width=\columnwidth]{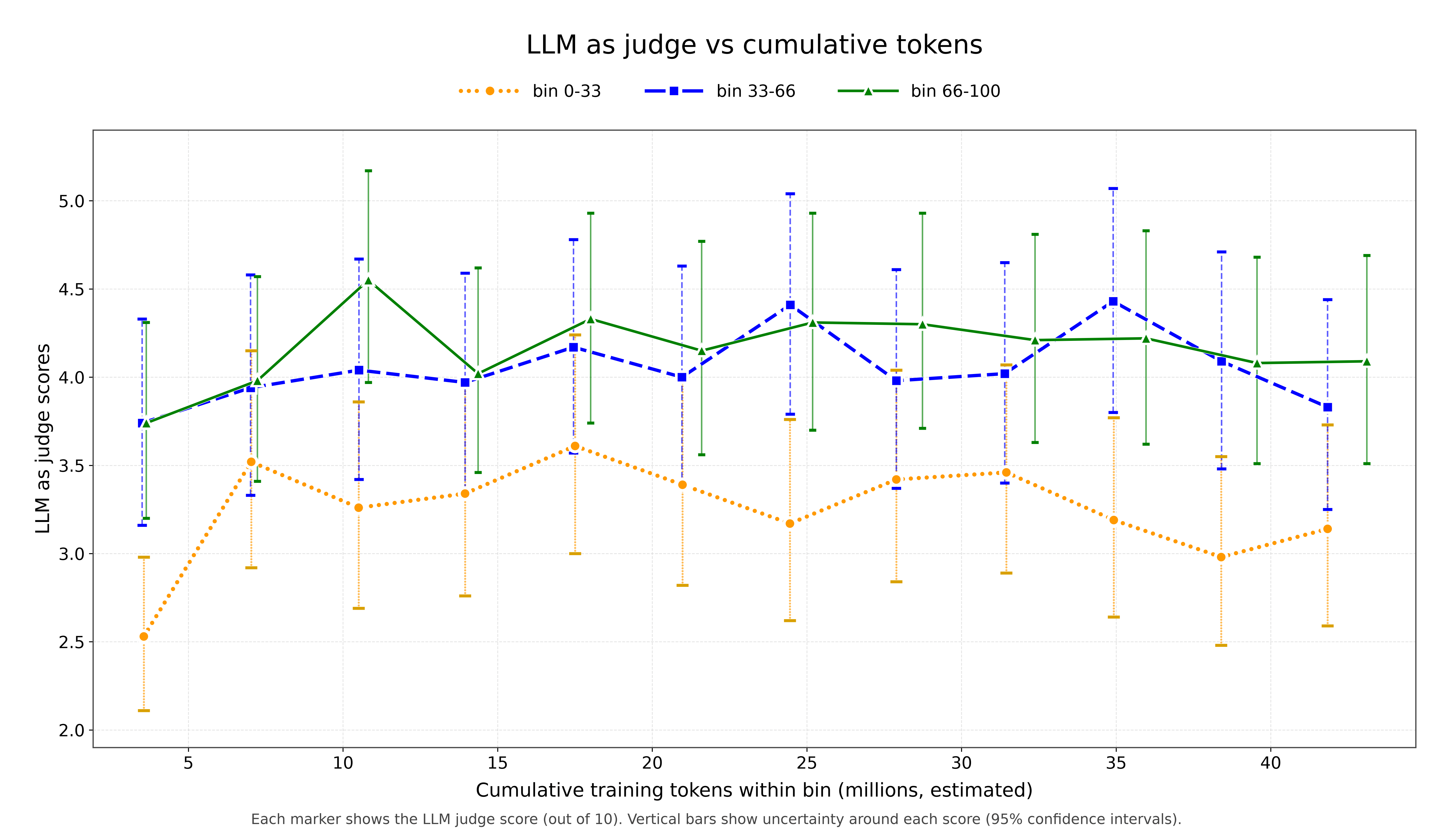}
}%
\caption{LLM-as-judge score curves for cross-ablation with Llama-3.1-8B on Natural Instructions 10\%, comparing alignment after removing high-response-ratio examples with response ratio after removing high-alignment examples.}
\vspace{4cm}
\label{figs:standard-cross-ablation-judge-llamaNI}
\end{figure*}
\begin{figure*}
\subfloat[Qwen3.5-9B-Base, Natural Instructions 10\%:\\Perplexity for alignment without high response ratio]{
  \includegraphics[width=\columnwidth]{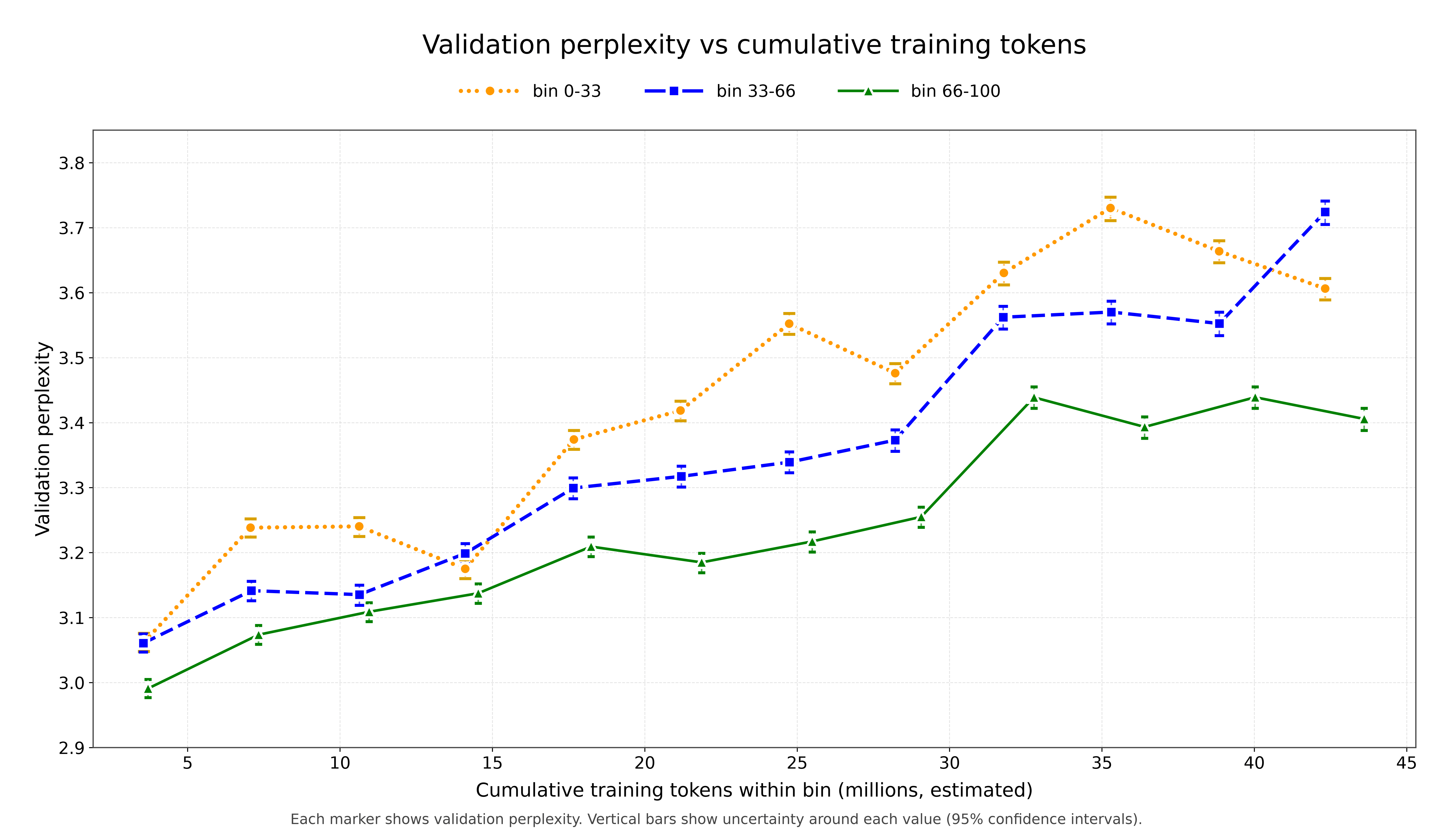}
}%
\subfloat[Qwen3.5-9B-Base, Natural Instructions 10\%:\\Perplexity for response ratio without high alignment]{
  \includegraphics[width=\columnwidth]{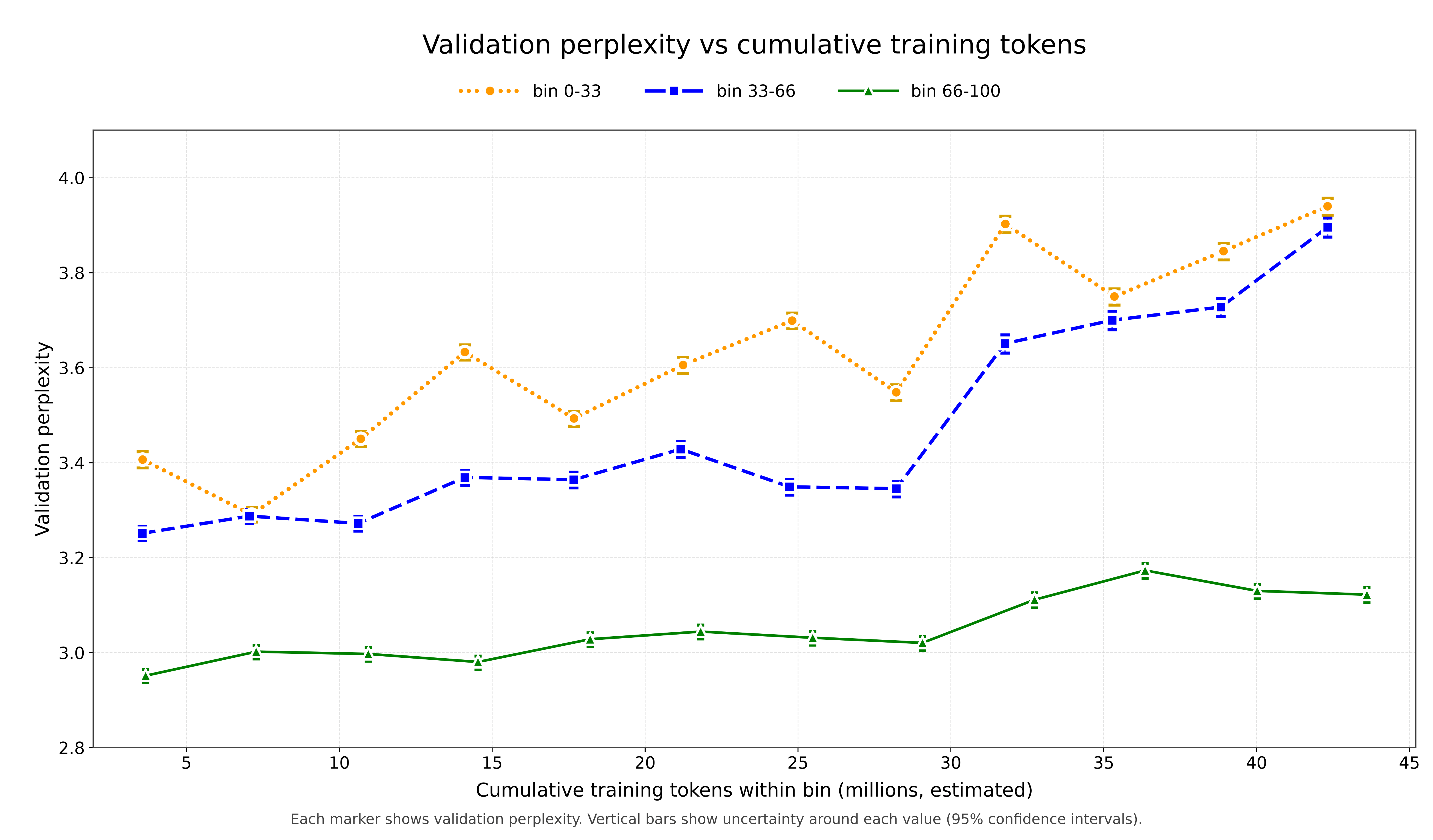}
}%
\caption{Perplexity curves for cross-ablation with Qwen3.5-9B-Base on Natural Instructions 10\%, comparing alignment after removing high-response-ratio examples with response ratio after removing high-alignment examples.}
\vspace{4cm}
\label{figs:standard-cross-ablation-qwenNI}
\end{figure*}
\begin{figure*}
\subfloat[Qwen3.5-9B-Base, Natural Instructions 10\%:\\LLM-as-judge scores for\\alignment without high response ratio]{
  \includegraphics[width=\columnwidth]{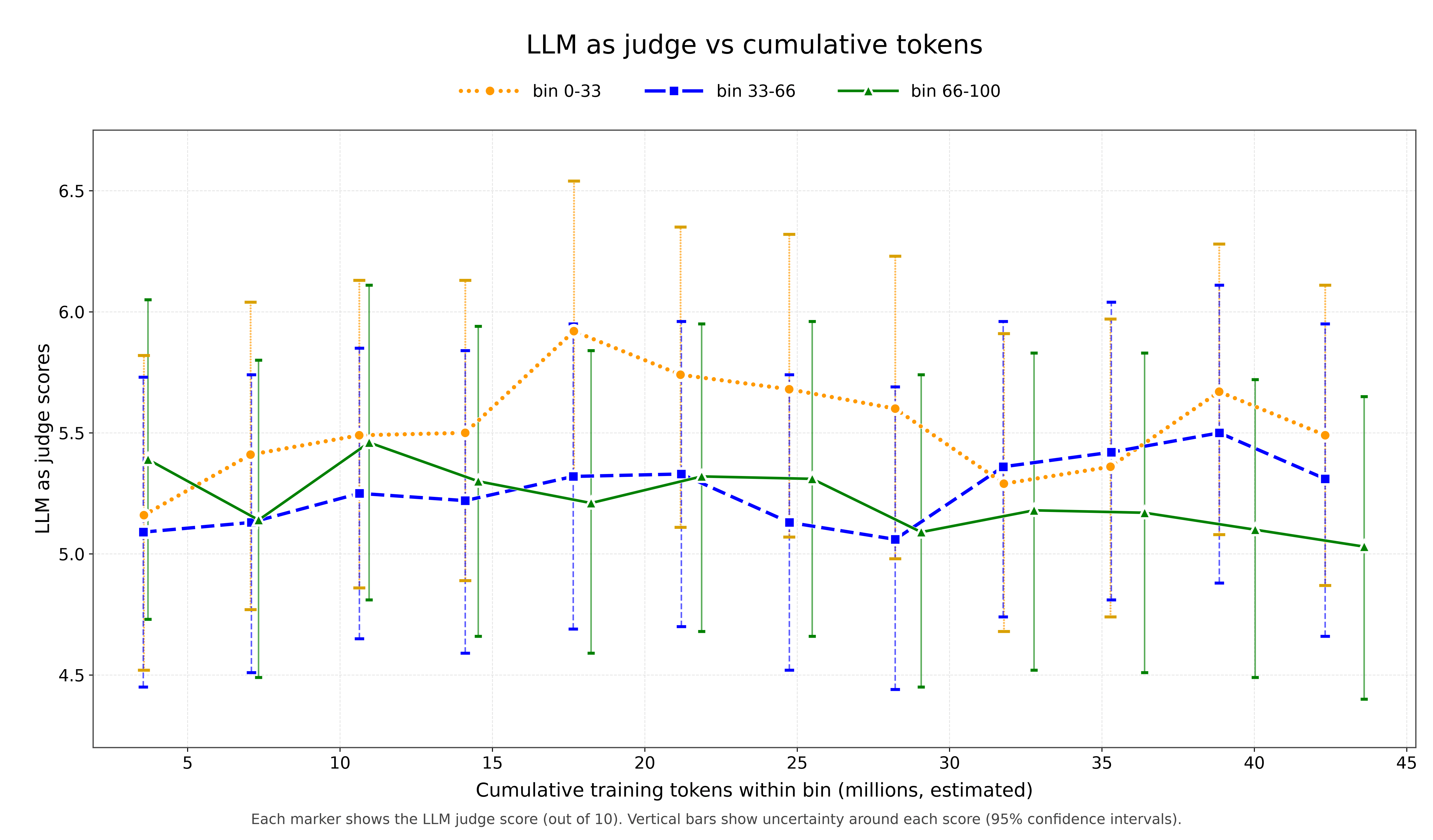}
}%
\subfloat[Qwen3.5-9B-Base, Natural Instructions 10\%:\\LLM-as-judge scores for\\response ratio without high alignment]{
  \includegraphics[width=\columnwidth]{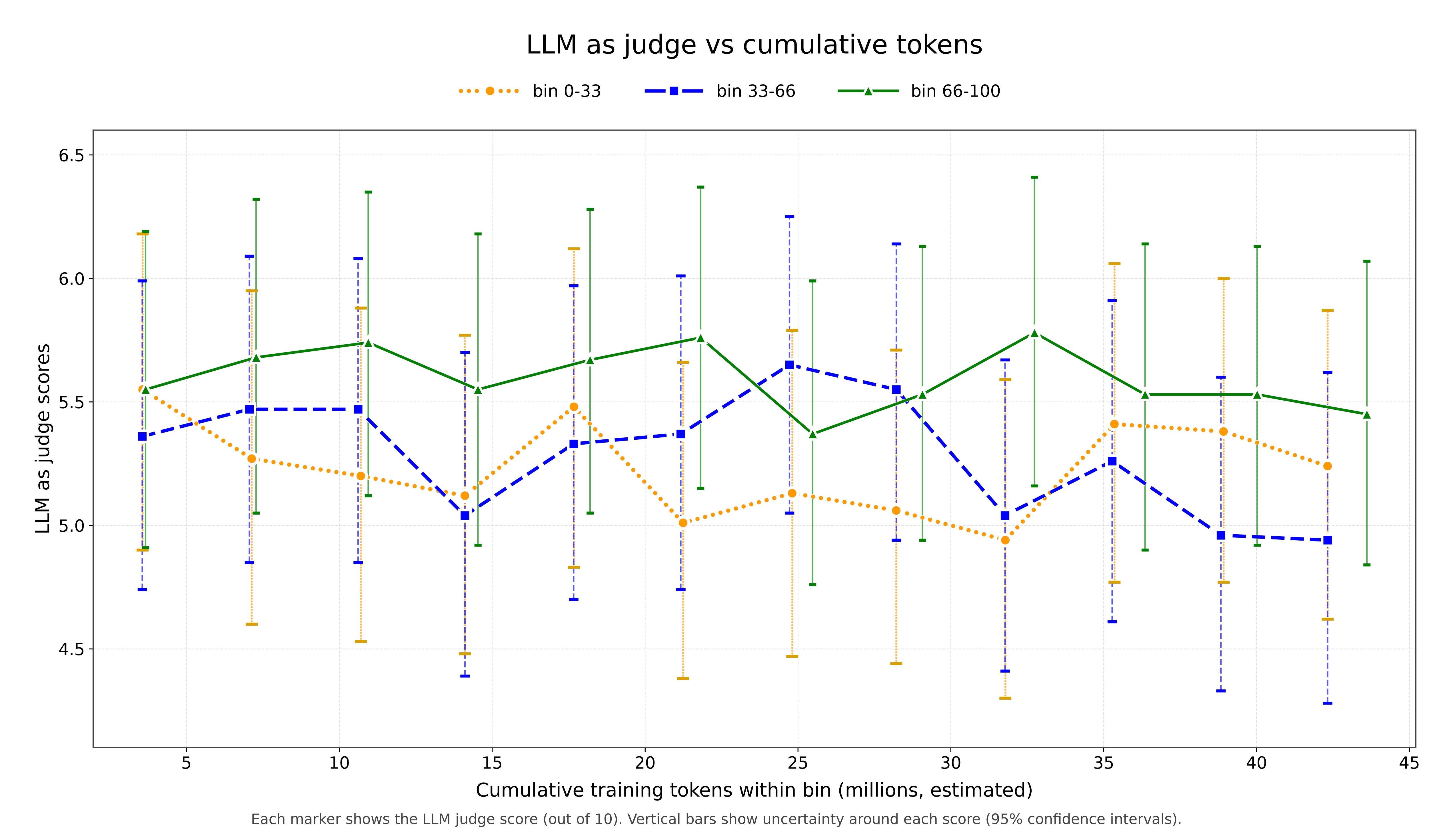}
}%
\caption{LLM-as-judge score curves for cross-ablation with Qwen3.5-9B-Base on Natural Instructions 10\%, comparing alignment after removing high-response-ratio examples with response ratio after removing high-alignment examples.}
\vspace{4cm}
\label{figs:standard-cross-ablation-judge-qwenNI}
\end{figure*}
\end{center}
\end{document}